\documentclass[lettersize,journal]{IEEEtran}
\usepackage{amsmath,amsfonts}
\usepackage{algorithmic}
\usepackage{array}
\usepackage[caption=false,font=normalsize,labelfont=sf,textfont=sf]{subfig}
\usepackage{textcomp}
\usepackage{stfloats}
\usepackage{url}
\usepackage{verbatim}
\usepackage{graphicx}
\def\BibTeX{{\rm B\kern-.05em{\sc i\kern-.025em b}\kern-.08em
    T\kern-.1667em\lower.7ex\hbox{E}\kern-.125emX}}
\usepackage{balance}

\usepackage{caption}  % Add the caption package for tables

\usepackage{bm}
\usepackage{overpic}
\usepackage{enumitem} %< control spacing in itemize/enumerate/...
\usepackage{overpic} %< add raw math symbols to figures
\usepackage{array}
\usepackage{pifont}% http://ctan.org/pkg/pifont

\usepackage{comment}
\usepackage{cite}

\usepackage{float} % for controlling figure placements
\usepackage{graphicx} % for including images
\usepackage{tabularx} % for flexible tables
\usepackage{booktabs}

\newcommand{\REMOVE}[1]{}
\newcommand{\methodname}{\textcolor{amethyst}{V}\textcolor{cadmiumorange}{I}\textcolor{limegreen}{R}\textcolor{royalazure}{Gi}}

\usepackage{xcolor}

\definecolor{amethyst}{rgb}{0.6, 0.4, 0.8}
\definecolor{darkpastelgreen}{rgb}{0.01, 0.75, 0.24}
\definecolor{amber}{rgb}{1.0, 0.75, 0.0}
\definecolor{cadmiumorange}{rgb}{0.93, 0.53, 0.18}
\definecolor{lawngreen}{rgb}{0.49, 0.99, 0.0}
\definecolor{limegreen}{rgb}{0.2, 0.8, 0.2}
\definecolor{neongreen}{rgb}{0.22, 0.88, 0.08}
\definecolor{amethyst}{rgb}{0.6, 0.4, 0.8}
\definecolor{darkpastelgreen}{rgb}{0.01, 0.75, 0.24}
\definecolor{greenbest}{RGB}{88,137,15}
\definecolor{redworst}{rgb}{0.83, 0.3, 0.30}
\definecolor{royalazure}{rgb}{0.25, 0.41, 0.88}

\definecolor{review}{rgb}{0.0, 0.0, 0.0}

\newcommand{\cmark}{\textcolor{limegreen}{\ding{51}}}
\newcommand{\xmark}{\textcolor{redworst}{\ding{55}}}%

\newcommand{\MLP}{{\textsc{MLP}}} % MLP
\newcommand{\MLPseg}{{\textsc{MLP}_{seg}}} % SEgmentation Module
\newcommand{\MLPc}{{\textsc{MLP}_c}} % Coloring Module
\newcommand{\MLPspec}{\textsc{MLP}_{\textrm{spec}}} % Specular Module
\newcommand{\MLPdiff}{\textsc{MLP}_{\textrm{diff}}} % Diffuse Module

\newcommand{\Cgs}{\textsc{C}} % Diffuse Module
\newcommand{\Cdiff}{\textsc{C}_{\textrm{diff}}} % Diffuse Module
\newcommand{\Cspec}{\textsc{C}_{\textrm{spec}}} % Diffuse Module
\newcommand{\hdiff}{h_{\textrm{diff}}} % Activations of the last layer of diffuse

\newcommand{\vanillamethod}{\textsc{Vanilla-}\methodname} % Baseline method (Vanilla VIRGi)

\newcommand{\viewd}{\theta}

\newcommand{\co}{c}
\newcommand{\rgbedit}{\textrm{{I}}^{\textrm{edit}}}

\newcommand{\feat}{\it{f}}

\begin{document}
\title{\methodname: View-dependent Instant Recoloring of 3D Gaussians Splats} % Elena v}

\author{Alessio Mazzucchelli, Ivan Ojeda-Martin, Fernando Rivas-Manzaneque, Elena Garces, Adrian Penate-Sanchez, Francesc Moreno-Noguer

\thanks{A. Mazzucchelli is with Arquimea Research Center, San Cristóbal de la Laguna, Santa Cruz de Tenerife, 38320 and Universidad Politécnica de Catalunya, Doctoral Degree in Automatic Control, Robotics and Vision, Carrer de Jordi Girona, 31, Les Corts, Barcelona, 08034, Spain.}
\thanks{I. Ojeda-Martin is with Arquimea Research Center, San Cristóbal de la Laguna, Santa Cruz de Tenerife, 38320.}
\thanks{F. Rivas-Manzaneque is with Volinga AI, San Cristóbal de la Laguna, Santa Cruz de Tenerife, 38320 and Universidad Politécnica de Madrid, Programa de Doctorado en Automática y Robótica, Calle de José Gutiérrez Abascal 2, Madrid, 28006, Spain.}
\thanks{E. Garces is with Universidad Rey Juan Carlos, C/Tulipán S/N, 28933 Móstoles (Madrid), Spain}
\thanks{A. Penate-Sanchez is with IUSANI, Universidad de Las Palmas de Gran Canaria, C/ Juan de Quesada 30, Las Palmas de Gran Canaria, 35001, Spain.}
\thanks{F. Moreno-Noguer is with Institut de Robòtica i Informàtica Industrial (IRI), CSIC-UPC, C/ Llorens i Artigas 4, Barcelona, 08028, Spain.}
}

\maketitle
\vspace{-6mm}

\begin{abstract}
\noindent3D Gaussian Splatting (3DGS) has recently transformed the fields of novel view synthesis and 3D reconstruction due to its ability to accurately model complex 3D scenes and its unprecedented rendering performance. However, a significant challenge persists: the absence of an efficient and photorealistic method for editing the appearance of the scene's content.
In this paper we introduce \methodname, a novel approach for rapidly editing the color of scenes modeled by 3DGS while preserving view-dependent effects such as specular highlights. 
Key to our method are a novel architecture that separates color into diffuse and view-dependent components, and a multi-view training strategy that integrates image patches from multiple viewpoints. Improving over the conventional single-view batch training, our 3DGS representation provides more accurate reconstruction and serves as a solid representation for the recoloring task.
For 3DGS recoloring, we then introduce a rapid scheme requiring only one manually edited image of the scene from the end-user. By fine-tuning the weights of a single MLP, alongside a module for single-shot segmentation of the editable area, the color edits are seamlessly propagated to the entire scene in just two seconds, facilitating real-time interaction and providing control over the strength of the view-dependent effects.
An exhaustive validation on diverse datasets demonstrates  significant quantitative and qualitative advancements over competitors based on Neural Radiance Fields representations.

%Our pipeline  comprises two key contributions. 
%Firstly, we present a mechanism for precisely learning view-dependent 3D Gaussian Splats. This involves enhancing the CUDA fast rasterizer of 3DGS to accommodate training batches that integrate image patches from multiple viewpoints, departing from the conventional single-view batch training. We demonstrate the significance of this multi-view strategy in accurately modeling specular and diffuse lighting components, each represented by separate MLPs.  Once this view-aware representation is acquired, we introduce a rapid recoloring scheme requiring only one manually edited image of the scene from the end-user. By fine-tuning the weights of the diffuse MLP alongside  a module for single-shot segmentation of the editable area, the color edits are seamlessly propagated to the entire scene in just two seconds, facilitating real-time interaction and providing control over the strength of the view-dependent effects.
%
%An exhaustive validation on diverse datasets demonstrates  significant quantitative and qualitative advancements over competitors based on Neural Radiance Fields representations.
\end{abstract}


\section{Introduction}  \label{sec:intro}
\noindent3D Gaussian Splatting (3DGS)~\cite{kerbl3Dgaussians} has recently established itself as a dominant technique in the fields of 3D reconstruction and image-based rendering. By employing a large set of anisotropic Gaussians to represent scenes, 3DGS facilitates rapid reconstruction and rendering processes, effectively preserving fine details and complex lighting effects. Compared to previous neural rendering methodologies such as NeRFs~\cite{mildenhall2020nerf}, 3DGS demonstrates markedly superior efficiency in both reconstruction and rendering.
%
%\begin{comment}
%    3DGS has applications in various downstream tasks, including text-to-3D generation~\cite{chen2024text,liang2024luciddreamer,FangGaussianEditor}, object addition/deletion~\cite{chen2023gaussianeditor}, semantic segmentation~\cite{zhou2023feature}, and object manipulation into separable parts~\cite{yu2023cogs}. However, unlike NeRFs, which has been extensively studied for its recoloring capabilities~\cite{gong2023recolornerf,kuang2023palettenerf,Lee_2023_ICCV,ireneCVPR2023,wang2023proteusnerf}, the recoloring problem has received little attention in the context of 3DGS.
%\end{comment}
%
%3D scene/object manipulation, such as instruction-based editing, drag-based editing, multi-object interaction, style transfer
\noindent 3DGS has application in various downstream tasks, including 3D scene/object manipulation~\cite{wang2024gscream, guedon2024sugar, zhou2023feature}, instruction-based editing~\cite{FangGaussianEditor, gaussctrl2024}, drag-based editing~\cite{shen2024draggaussian}, multi-object interaction~\cite{xie2023physgaussian} and style transfer~\cite{liu2024stylegaussian, zhang2024stylizedgs}. However, unlike NeRFs, which has been extensively studied for its recoloring capabilities~\cite{gong2023recolornerf,kuang2023palettenerf,Lee_2023_ICCV,ireneCVPR2023,wang2023proteusnerf}, the recoloring problem has received little attention in the context of 3DGS.

\begin{figure*}[t]
\centering
    \begin{center}
    \begin{tabularx}{\textwidth}{XXXXXX} % Define your column layout here
    \centering Original image & 
    \centering Recolor & 
    \centering Multi-recolor & 
    \centering Novel view & 
    \centering Diffuse & 
    \centering Specular
    \end{tabularx}
    \end{center}
\vspace{-2mm}
    \centering
    \includegraphics[trim={8.7cm 0 4.7cm 0}, clip, width=2.9cm]{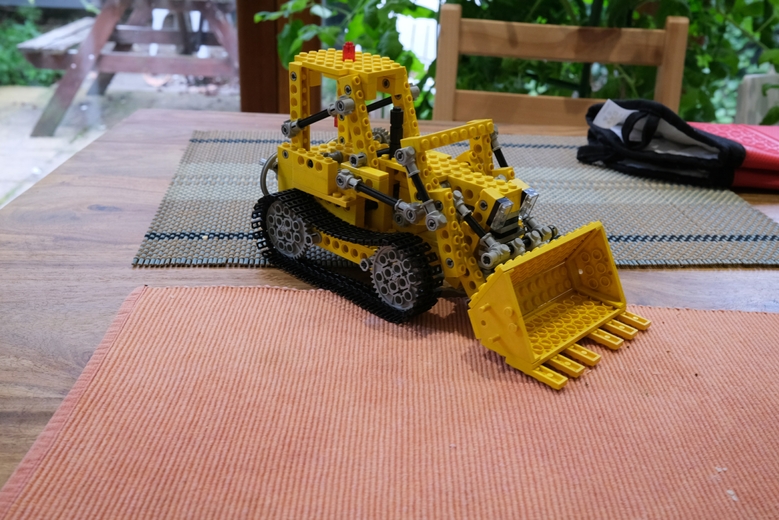}
    \includegraphics[trim={8.7cm 0 4.7cm 0}, clip, width=2.9cm]{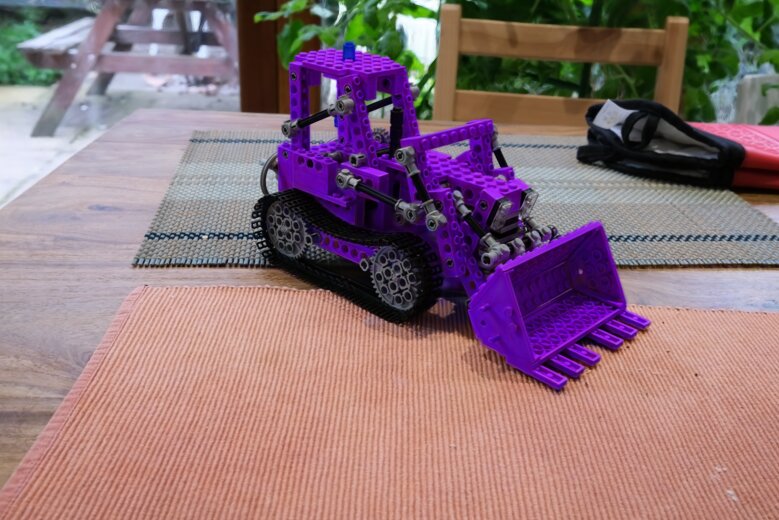}
    \includegraphics[trim={8.7cm 0 4.7cm 0}, clip, width=2.9cm]{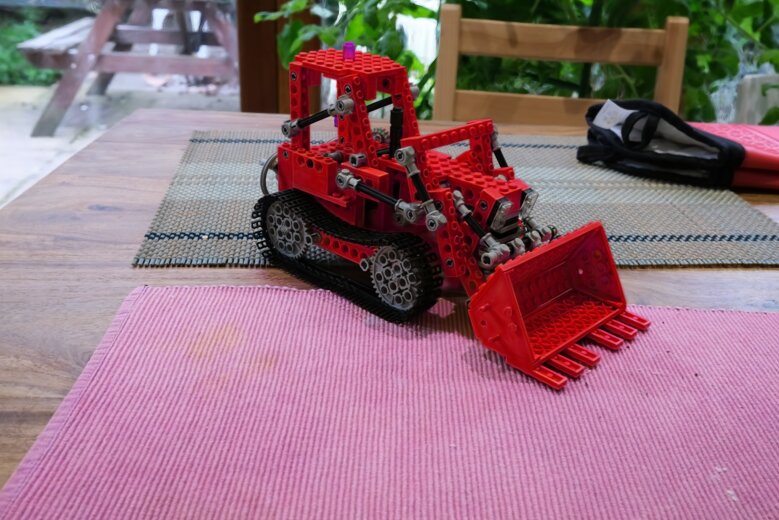}
    \includegraphics[trim={8.7cm 0 4.7cm 0}, clip, width=2.9cm]{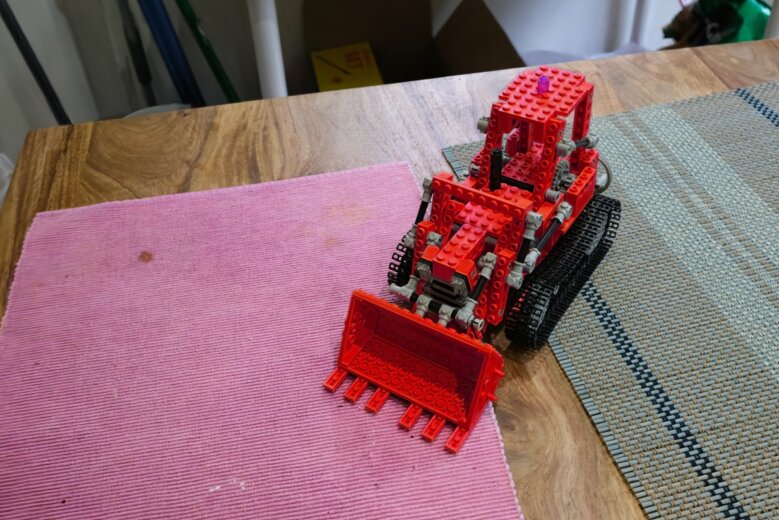}
    \includegraphics[trim={8.7cm 0 4.7cm 0}, clip, width=2.9cm]{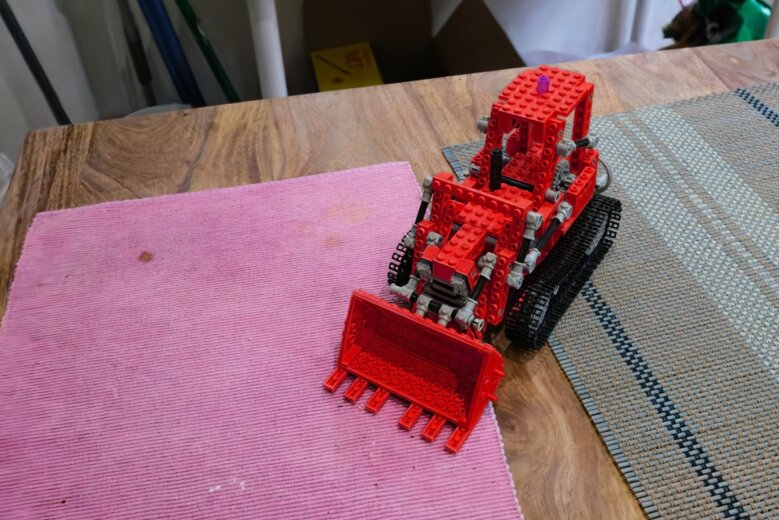}
    \includegraphics[trim={8.7cm 0 4.7cm 0}, clip, width=2.9cm]{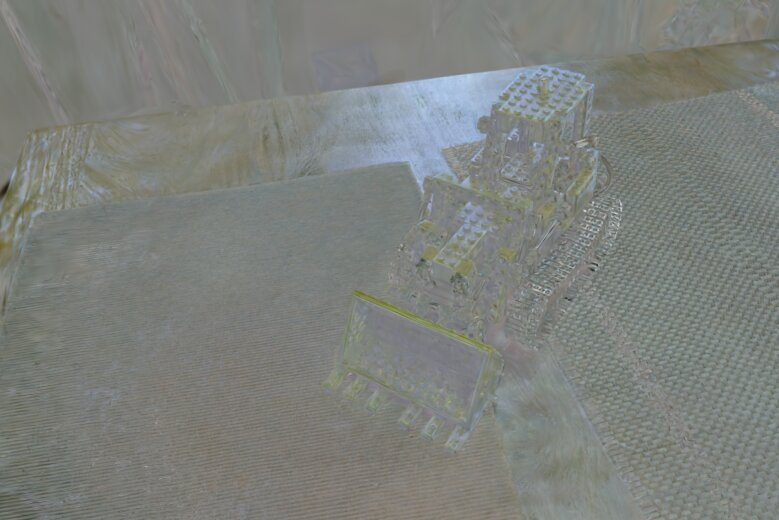} \\
\vspace{-1mm}
    \caption{\methodname~ enables -- by editing a single view -- instant recolorings of one or multiple objects in scenes modeled with 3D Gaussian Splats. Our key idea of decomposing the reflectance of the scene into diffuse and specular components allows view-dependent consistent edits. Our separation can also be used to enhance or reduce material specularity.}
    \label{fig:teaser}
\vspace{-1mm}
\end{figure*}

\noindent In this paper we introduce \methodname, the first approach for photorealistic appearance editing of scenes modeled with 3DGS. Our approach is highly efficient and user friendly. Given a scene modeled with 3DGS, the user simply provides color edits in a single view. These edits are then rapidly propagated throughout the entire scene and viewpoints within approximately two seconds. This, combined with the inherent rendering efficiency of 3DGS, enables a quasi-real-time interactive editing experience.

\begin{figure*}[t]
    \centering
	\includegraphics[width=0.98\textwidth]{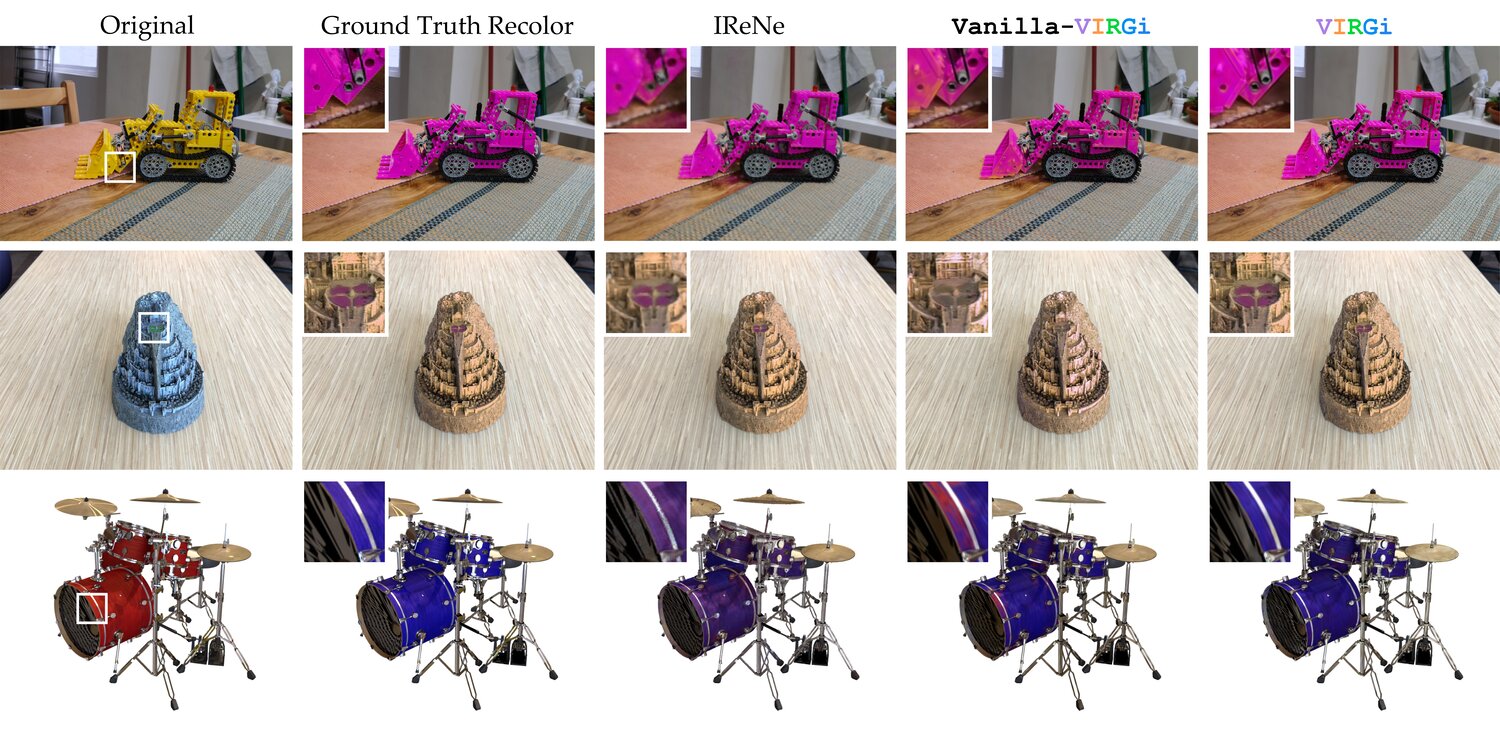} 
\vspace{-1mm}
    \caption{{\bf Comparison of recoloring methods with a ground truth recolor.} Our method \methodname~is the first one to propose a solution for 3DGS recoloring, outperforming one of the best methods in NeRFs recoloring~\cite{ireneCVPR2023} (IReNe). We propose several contributions that make our method robust and better than a baseline (\vanillamethod) obtained by na\"ively combining existing methods.
    %IReNe~\cite{ireneCVPR2023} 
    %The third column shows the results obtained with IReNe~\cite{ireneCVPR2023}. The fourth shows results of \vanillamethod. The last column shows the results obtained with \methodname. 
    }
\label{fig:motivation_figure}
\centering
\vspace{-1mm}
\end{figure*}

\noindent Our solution is built upon two primary components. First, we present a novel architecture and training scheme tailored for accurately modeling viewpoint dependencies. Traditional 3DGS systems~\cite{kerbl3Dgaussians} are trained on a per-view basis, with training batches composed of image tiles from a single viewpoint. In our approach we implement multi-view training by modifying the CUDA implementation of the rasterizer, enabling the use of different viewpoints in the same training batch. This approach, as we will show later, facilitates the decomposition of reflections into diffuse and view-dependent components (e.g. specularities), encapsulated in two independent MLPs. 
The second component of our approach is a pipeline for efficient scene recoloring. Given a target view with the desired color edition, we propose a strategy where we only fine-tune the last layer of the diffuse MLP, in conjunction with a module designed to perform soft segmentation of the edited region. This optimization ensures the consistent propagation of color edits throughout the entire scene, preserving view-dependent effects and preventing artifacts such as color bleeding commonly observed in other approaches. Since this process operates at interactive rates, users can recursively provide new edited images if required.
We validate our recoloring method across multiple datasets and conduct a comparative analysis with previous NeRF-based methodologies. \methodname~consistently outperforms these approaches not only in efficiency and interactivity  but also in quality. We demonstrate that our decoupled diffuse/view-dependent color modeling enables the creation of highly photorealistic color editions,  while also empowering the user to control specular effects, such as the strength of the highlights. Fig.~\ref{fig:teaser}, and the supplementary video showcase examples of the  results achieved with our method, consistently surpassing competing approaches as demonstrated in Fig.~\ref{fig:motivation_figure}.
To summarize, we provide the following contributions:

\begin{itemize}[leftmargin=*]
    \item The first method for conducting photorealistic, view-dependent, and interactive color edits for scenes modeled with Gaussian Splats. Our approach is designed to be user-friendly and interactive, requiring only a single color edit from one view to recolor the entire scene.
    \item A novel neural network architecture for 3DGS that learns the decoupling of color into its specular and diffuse components. This architecture enables precise color editing of scenes while preserving their view-dependent effects.
    \item A multi-view training strategy that is essential to learning the diffuse/specular color components. Incorporating information from different viewpoints during training enhances not only the learning of the decomposed output in our novel architecture, but also the overall performance of the 3DGS reconstruction.
\end{itemize}

\section{Related Work}  \label{sec:related}

%THIRD TRY ;)

\noindent 3D Gaussian Splatting~\cite{kerbl3Dgaussians} provides real-time rendering with simpler manipulation compared to Neural Radiance Fields (NeRFs)~\cite{mildenhall2020nerf} because of its explicit scene representation. However, globally editing 3DGS scenes remains an open problem, as we discuss in this section.

\vspace{1mm}
\noindent{\bf Point and 3DGS Rendering.} 
Gaussian Splats originate from point-cloud representations~\cite{gross2011point}. These techniques became very popular in 3D modeling~\cite{zwicker2002pointshop,ohtake2005multi} and as the basis for image-based rendering techniques~\cite{seitz2006comparison,thonat2016multi,baek2016multiview,snavely2006photo,chaurasia2013depth}. In recent years, Gaussian Splats~\cite{kerbl3Dgaussians} have surpassed previous point-based rendering methods by enabling real-time novel view synthesis for scenes of varying complexity. These scenes range from those with simple Lambertian materials to those with intricate light interactions, dynamic elements~\cite{yang2024deformable3dgs,wu20234dgaussians,das2023neural,lin2024gaussian,li2024spacetime,sun20243dgstream,xie2023physgaussian}, physical simulators~\cite{xie2023physgaussian}, and   avatars~\cite{li2024animatablegaussians,hugs2023,abdal2023gaussian,zheng2024gpsgaussian,hu2024gaussianavatar,xiang2024flashavatar,zhu2023ash,qian20233dgsavatar,yuan2023gavatar}.

\noindent \textcolor{review}{Many works have already addressed the memory and storage limitations of 3DGS, leading to efficient compression strategies that now allow deployment even on lightweight devices~\cite{kerbl2024hierarchical}. In particular, vector clustering and quantization-based methods have proven effective in minimizing parameter counts~\cite{fan2023lightgaussian,navaneet2023compact3d,morgenstern2023compact,niedermayr2024compressed}. Notably,~\cite{lee2023compact} employ vector quantization to construct codebooks. However, unlike other methods, their approach explicitly separates color information from geometry. A major drawback that remains to be addressed, and which we tackle in this paper, is the constraint of performing the optimization step over a single image. Unlike NeRF, where rays from different viewpoints can be sampled with minimal overhead, the 3DGS rasterization pipeline is inherently built around a single image. In this work, we adopt the representation from~\cite{lee2023compact} and extend it by augmenting the CUDA rasterizer so that optimization can be done from multiple images in a single step.}

\vspace{1mm}
\noindent{\bf Inverse Rendering in Radiance Fields.} \textcolor{review}{
NeRF-based approaches have incorporated surface normals and separated diffuse/specular components to improve reflectance modeling and enable scene relighting, as demonstrated in~\cite{verbin2022ref, wu2023nerf}. Other works~\cite{zhang2023neilf++, wu2023nefii, wu2024pbr} jointly optimize geometry, materials, and illumination, leveraging physics-based priors or path tracing with near-field indirect light to better disentangle reflective properties from environmental lighting. In the domain of 3D Gaussian Splatting, inverse rendering has been enhanced by estimating normals, materials, and indirect lighting for each Gaussian primitive~\cite{liang2024gs}. To better handle ambiguities between surface normals and roughness, a segmentation prior is introduced in~\cite{lai2025glossygs}, guiding geometry and material decomposition. Normals are estimated from the shape of each Gaussian in~\cite{shi2023gir}, where directional masking enforces orientation, enabling BRDF rendering and disentangling indirect light via voxel-based tracing. A dedicated specular shading is proposed in~\cite{ye20243d} to refine surface normals and enhance high-frequency reflections without compromising splatting efficiency.}

\vspace{1mm}
\noindent{\bf Color Editing.}
Color editing through edit propagation strategies utilizes local color cues, such as user-defined scribbles or points, to influence similar regions within the image. Previous works in this domain employed propagation rules based on color similarity metrics~\cite{endo2016deepprop,an2008appprop,chen2012manifold,li2008scribbleboost}. However,  defining effective metrics that capture both local and non-local image features posed a challenge. To address this, Convolutional Neural Networks (CNNs) proved successful for robust feature extraction~\cite{meyer2018deep,zhang2019deep}.
\noindent A different approach involves pre-computed palettes representing dominant colors in an image. These palettes serve as a basis for describing other colors through linear combinations. Some artists  prefer this method for  its control over global and local color modifications, often  combining it with object segmentation techniques~\cite{chang2015palette,tan2018efficient,Du:2021:VRS}.
\noindent In this work, we propose a method that can leverage any of these paradigms to transition from a 2D to a full 3D edit. Our method allows users to edit a single 2D image using their preferred editing tool or technique. Thanks to our efficient propagation strategy, any 2D recoloring is extrapolated to a 3DGS-based representation. 

\vspace{1mm}
\noindent{\bf Radiance Fields Color Editing.} 
Despite their brief existence, Gaussian Splats have sparked rapid development, with various manipulation techniques emerging. Text prompts are used to modify static scenes~\cite{chen2024text,liang2024luciddreamer,FangGaussianEditor} or dynamic ones~\cite{ling2024alignyourgaussians,huang2024sc}. Additionally, ~\cite{chen2023gaussianeditor} enable object selection, while~\cite{yu2023cogs} create articulated Gaussian Splats from dynamic sequences. \cite{zhou2023feature}  propose diverse edits, such as semantic segmentation, object segmentation, or text-based editing, through feature distillation.
Unlike for NeRFs~\cite{gong2023recolornerf,kuang2023palettenerf,Lee_2023_ICCV,wang2023proteusnerf}, to the best of our knowledge, there is no specific work for Gaussian Splat recoloring. Our work draws inspiration from~\cite{ireneCVPR2023}, who, by fine-tuning part of the NeRF, achieved remarkable quality and view-dependent instant recoloring. However, as we will demonstrate in the experimental section, our method, which incorporates multi-view training and explicit diffuse/specular modeling, significantly enhances the quality compared to~\cite{ireneCVPR2023}. %\francescrmk{Can we say this about our advantage w.r.t. IRENE?} \ferrmk{I think we can (based on table 2)}

\section{Background and \protect\vanillamethod}
\label{sec:method}

\noindent We ground our approach on 3D Gaussian Splatting (3DGS)~\cite{kerbl3Dgaussians}, particulary focusing on Compact 3D Gaussian Splatting (C3DGS)~\cite{lee2023compact}. 3DGS represents the radiance field by generating a collection of 3D Gaussians,  each incorporating spherical harmonics information. During the rendering process, these Gaussians are splatted onto a 2D plane  to produce the 2D image~\cite{splattingFirstPaper}. However, a notable limitation of 3DGS is its significant memory requirement, averaging  746 MB for the Mip-NeRF dataset~\cite{barron2022mipnerf360}. To address this challenge, C3DGS introduced two solutions: encoding the Gaussians using a codebook to conserve memory, and a key contribution for our purposes, utilizing a hashgrid and an $\MLP$ to encode  appearance information. Their latter contribution eliminates the necessity  to store  spherical harmonics coefficients for each Gaussian, thereby reducing the memory footprint by a factor of 15. Consequently,  the memory requirements decrease substantially to an average of 48.8 MB on Mip-NeRF dataset.

%Our proposed approach adapts to C3DGS the recoloring strategy for NeRFs from Mazzucchelli ~\etal~\shortcite{ireneCVPR2023} (IReNe), one of the top performing methods for recoloring NeRFs.
%
%IReNe asks the user to recolor a single view of the NeRF and propagates the edits to the full NeRF using a strategy based on fine-tuning part of the color MLPs, selecting the neurons in charge of the view-dependent effects, and performing soft-segmentation in 3D space. We take these three key ideas and adapt them to C3DGS.

\noindent Our proposed approach adapts the recoloring strategy for NeRFs from IReNe~\cite{ireneCVPR2023}, one of the top-performing methods in this domain. IReNe takes as input a single recolored view of the NeRF and propagates the edits to the entire NeRF using a strategy based on fine-tuning part of the color MLPs, selecting the neurons responsible for view-dependent effects, and performing soft segmentation in 3D space. We leverage some of these concepts and tailor them to suit C3DGS.

\begin{figure*}[t!]
	\centering
	\includegraphics[trim={0 3mm 0 3mm}, clip, width=0.97\textwidth]{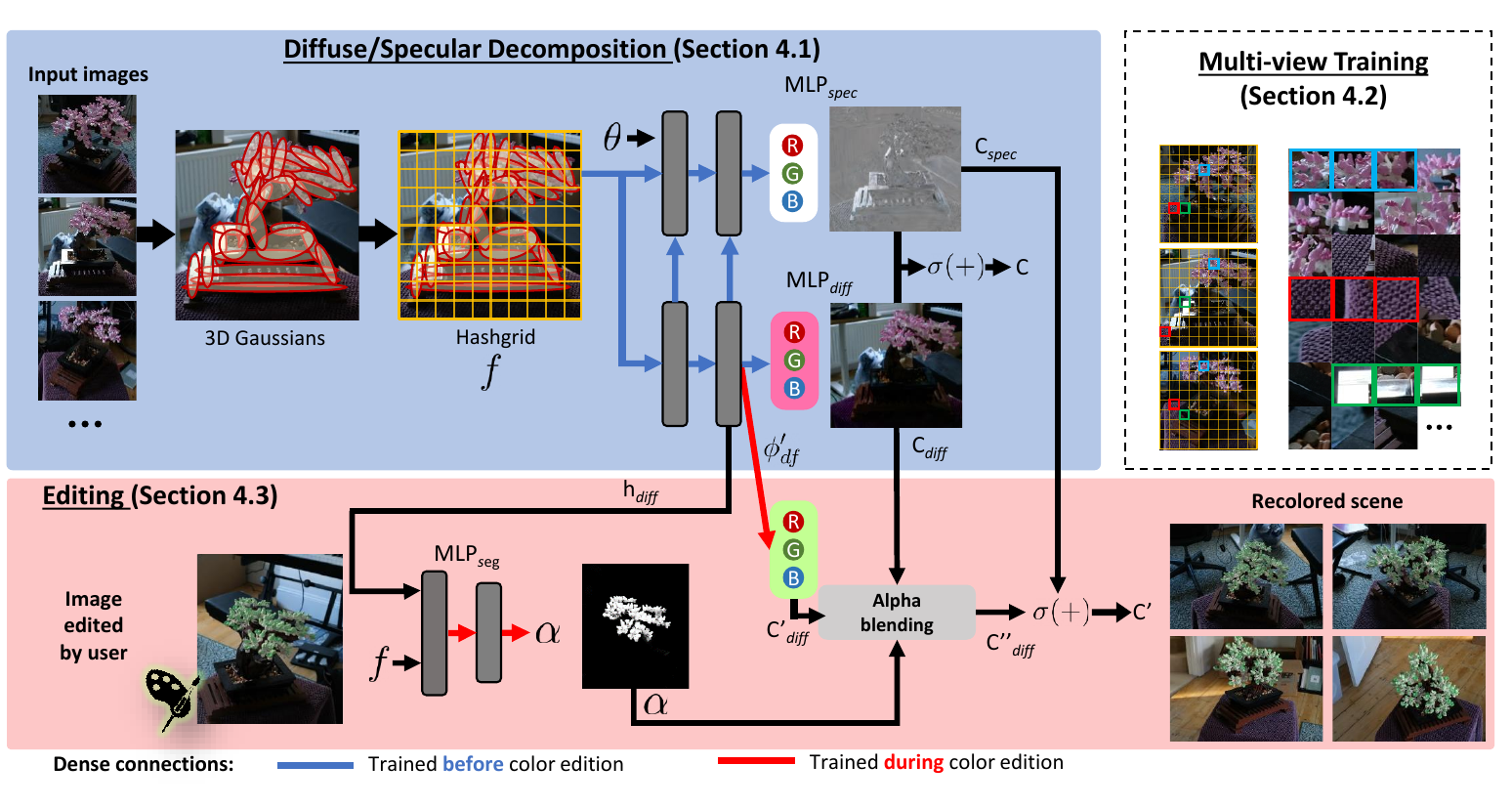}
\vspace{-2mm}
	\caption{{\bf Overview of \methodname.} 
 Our method comprises two main steps. In the initial step, we train a 3DGS architecture consisting of a Hashgrid $f$ representing the geometry and two MLPs for color modeling. We separate the observed color into a diffuse term $\MLPdiff$, which solely relies on hashgrid features $f$, and a specular term $\MLPspec$, which incorporates the view direction $\theta$. Our training strategy involves leveraging multiple views of the same scene point in a single batch, as opposed to using a single image per batch, resulting in improved PSNR compared to previous methods. In the second step, given an edited 2D image, we fine-tune the last layer of the diffuse MLP using a soft-segmentation mask $\alpha$ to achieve the target color.
 %Our method has two steps. In the first step, we train a 3DGS composed of a Hashgrid $f$ that represents the geometry, and two MLPs that model color. We decompose the observed color into a diffuse term $\MLPdiff$ that only receives as input the hashgrid features $f$ and a specular term $\MLPspec$ that includes the view direction $\theta$. During the training, we avoid explicit losses for the diffuse/specular separation, and, instead, leverage the proposed architecture and a novel training strategy for 3DGS that uses several views of the same scene point in a single batch. As opposed to using a single image per batch, our strategy proves effective for a better PSNR than previous work. In the second step, for a given edited 2D image, we propose fine-tuning the last layer of the diffuse MLP with a soft-segmentation mask $\alpha$ to obtain the target color.
 }
	\label{fig:network}
	\centering
 \vspace{-3mm}
\end{figure*}

\subsection{\normalfont\vanillamethod}
\noindent Given the differing radiance field representations between NeRFs and 3DGS, combining both strategies presents challenges. Our initial idea, dubbed \vanillamethod, attempted to integrate the MLP configuration and hashgrid features from C3DGS, along with the last layer and soft-segmentation blending from IReNe in a straightforward manner.
Specifically, \vanillamethod~utilizes a single neural network to estimate the color \Cgs~
 of each Gaussian, following the formulation proposed in \cite{lee2023compact}, expressed as:
\begin{equation}
	\centering
	\Cgs = \MLPc(\feat, \viewd | \phi_{\co}),
	\label{eq:c3dgs_color}
\end{equation}
where $\feat$ are the hashgrid features, $\viewd$ denotes the view direction, and $\phi_{\co}$ are the weights of the $\MLP$.
Then, following the approach outlined in IReNe~\cite{ireneCVPR2023}, \vanillamethod~modifies only the last layer of $\MLPc$ when an edit is performed, resulting in a new color:
\begin{align}
	\Cgs^{\prime} &= \MLP_{c}(\feat, \viewd | \phi_c^{last}).
	\label{eq:last_layer}
\end{align}
$\Cgs^{\prime}$ is blended with the original to produce the final color $\Cgs^{\prime\prime}$:
\begin{align}
	\Cgs^{\prime\prime} &= \alpha  \Cgs^{\prime} + (1- \alpha)  \Cgs.
	\label{eq:irene_blending}
\end{align}

\noindent Blending is performed as $\alpha = \MLPseg (\feat)$, which is a soft segmentation of the edited region and is learned by an MLP from the hashgrid space features. Additionally, this approach incorporates  a view-dependent neuron selection strategy within the last layer of $\MLP_{c}$. For further details, we encourage readers to refer to the original paper \cite{ireneCVPR2023}.

\noindent However, as depicted in Figure~\ref{fig:motivation_figure}, this straightforward approach does not yield optimal results, as traces of the previous color persist in the rendering of novel views. This outcome is somewhat expected, as this method lacks mechanisms to enforce multi-view consistency, leading to the entanglement of view-dependent effects such as specular highlights within the color $\MLPc$.

\noindent Our proposed solution, \methodname, aims to establish a clear separation between diffuse and view-dependent (specular) effects~\cite{kuang2023palettenerf,shafer1985using,beigpour2011object} from the initial construction of the Gaussian Splats. We introduce two pivotal concepts detailed in the following section: 
Firstly,  a decoupled diffuse-specular color estimation through distinct MLPs;  and secondly, a multi-view training strategy that leverages the principle that the diffuse color of an object with isotropic reflectance remains invariant to view changes~\cite{pharr2023physically}.

%This blending uses the soft-segmentation $\alpha = \MLPseg (\feat)$ of the edited region learnt by an MLP from the features values of the hashgrid space. This approach is combined with a view-dependent neuron selection strategy that takes place in the last layer of $\MLP_{c}$. We refer the reader to the original paper \cite{ireneCVPR2023} for the full details. 

%As shown in Figure~\ref{fig:motivation_figure}, this simple solution does not work perfectly, and the previous color still appears visible when rendering novel views. This behavior makes sense because there is nothing in that approach forcing multi-view consistency, so view-dependent effects such as specular highlights are coupled inside the color $\MLPc$. 
%
%Our proposed solution \methodname ~aims to enforce this separation between diffuse and specular (view-dependent effects) from the initial construction of the Gaussian Splats. We introduce two key ideas explained next: First, a decoupled diffuse-specular color estimation through different MLPs, and second, a multi-view training strategy that leverages the idea that the diffuse color of an object with isotropic reflectance is invariant to view changes \Elena{find references}.

\section{Method} 
\label{sec:method}
\noindent Figure~\ref{fig:network} provides an overview of our method, which operates in two main steps. First, we initiate the process by constructing a compact 3D scene representation~\cite{lee2023compact} from a collection of multi-view images. This learned representation comprises two primary components: (i) a set of hashgrid features $\feat$, which capture the geometric details of the scene, and (ii) a pair of Multi-Layer Perceptrons (MLPs) responsible for encoding the scene's appearance. Unlike C3DGS, we employ separate MLPs, $\MLPdiff$ and $\MLPspec$, to model diffuse color $\Cdiff$ and specular $\Cspec$ reflections, respectively. The architecture of this decoupling is explained in Section~\ref{subsec:decoupled}.

%Figure~\ref{fig:network} presents an overview of the method. Our method works in two steps. 
%
%First, we begin by constructing a compact 3D scene representation~\cite{lee2023compact} from a set of multi-view images.
%The learned representation comprises two components: (i) a set of hashgrid features $\feat$ which capture the scene's geometric information, and (ii) a set of Multi-Layer Perceptrons (MLPs) that encode the scene's appearance.
%
%As opposed to C3DGS, we model diffuse $\Cdiff$ and specular $\Cspec$ reflection with different MLPs ($\MLPdiff$ and $\MLPspec$, respectively). The architecture of this decoupling is explained in Section~\ref{subsec:decoupled}.
%

\noindent To train our 3DGS, we employ a novel optimized multi-view training strategy, detailed in Section~\ref{subsec:multiview}, that implicitly encourages corresponding 3D points to occupy nearby locations within the neural feature space. This novel approach, unlike conventional training with single images per batch~\cite{kerbl3Dgaussians,lee2023compact}, captures more consistent and better reflections, as demonstrated in the experiments section. Once the original scene is learned, users can utilize any existing tool to recolor one view of the scene $\rgbedit$. Subsequently, our optimized fine-tuning process, described in Section~\ref{subsec:editing}, ensures that the color edits are consistently propagated throughout the entire scene at interactive rates.

%To train our initial 3DGS, we employ a novel optimized multi-view training strategy, detailed in Section~\ref{subsec:multiview}, that implicitly encourages corresponding 3D points to occupy nearby locations within the neural feature space. This novel strategy, in contrast to simply training with single images per batch~\cite{kerbl3Dgaussians,lee2023compact}, captures more consistent and better reflections, as we demonstrate in the experiments section.

%Once the original scene is learned, users can leverage any existing tool to recolor one view of the scene $\rgbedit$. Then, our optimized fine-tuning process, explained in Section~\ref{subsec:editing}, ensures that the color edits are consistently propagated throughout the entire scene at interactive rates. 

\begin{figure}[t]
\begin{center}
\begin{tabularx}\columnwidth{XXX}
\centering \hspace{-1mm} Original image & 
\centering \hspace{-3mm} Monoview & 
\centering \hspace{-3mm} Multiview
\end{tabularx}
\vspace{-4mm}
\end{center}
    \centering
    \includegraphics[trim={440px 550px 100px 70px}, clip, width=0.32\columnwidth]{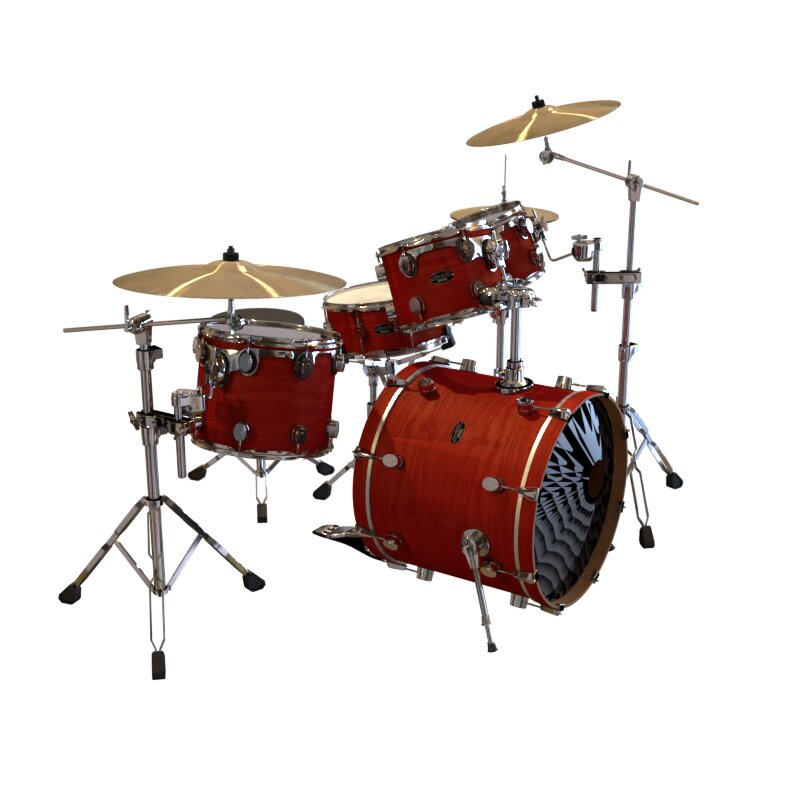} 
    \includegraphics[trim={440px 550px 100px 70px}, clip, width=0.32\columnwidth]{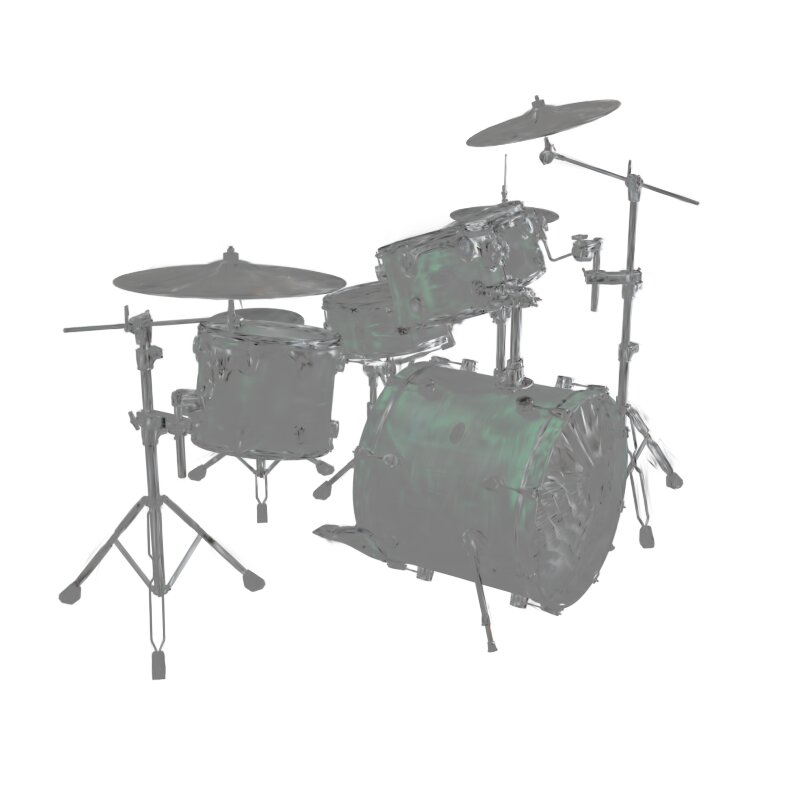} 
    \includegraphics[trim={440px 550px 100px 70px}, clip, width=0.32\columnwidth]{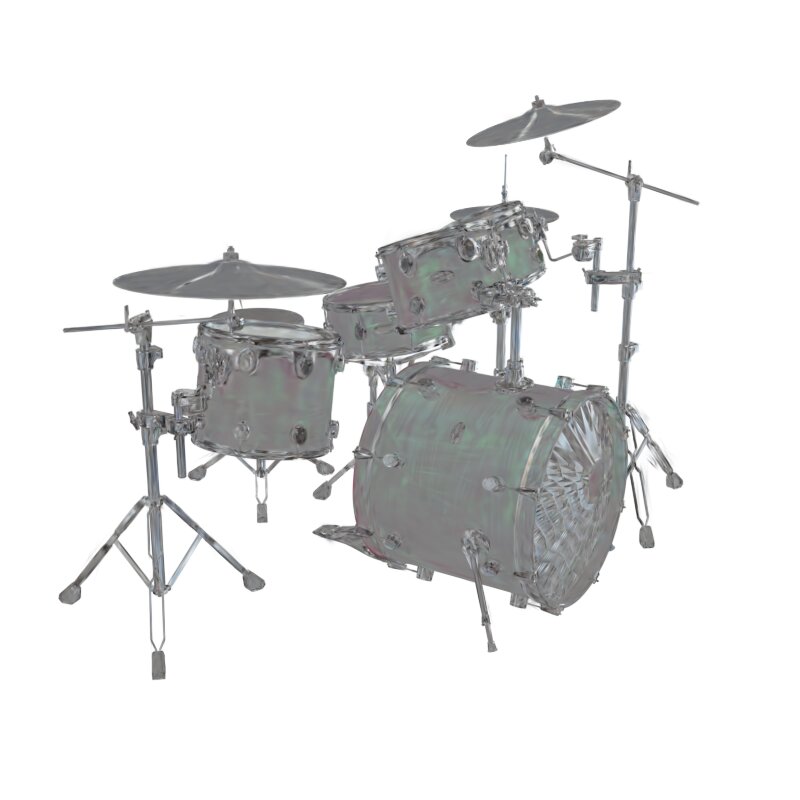} \\
    \vspace{-1px}
    \includegraphics[trim={480px 260px 0px 300px}, clip, width=0.32\columnwidth]{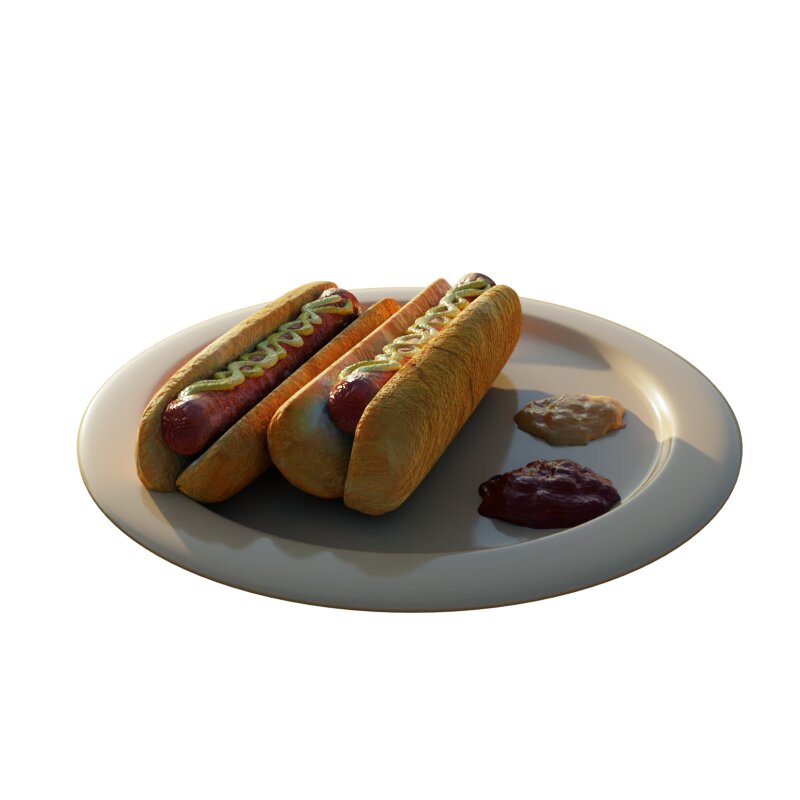}
    \includegraphics[trim={480px 260px 0px 300px}, clip, width=0.32\columnwidth]{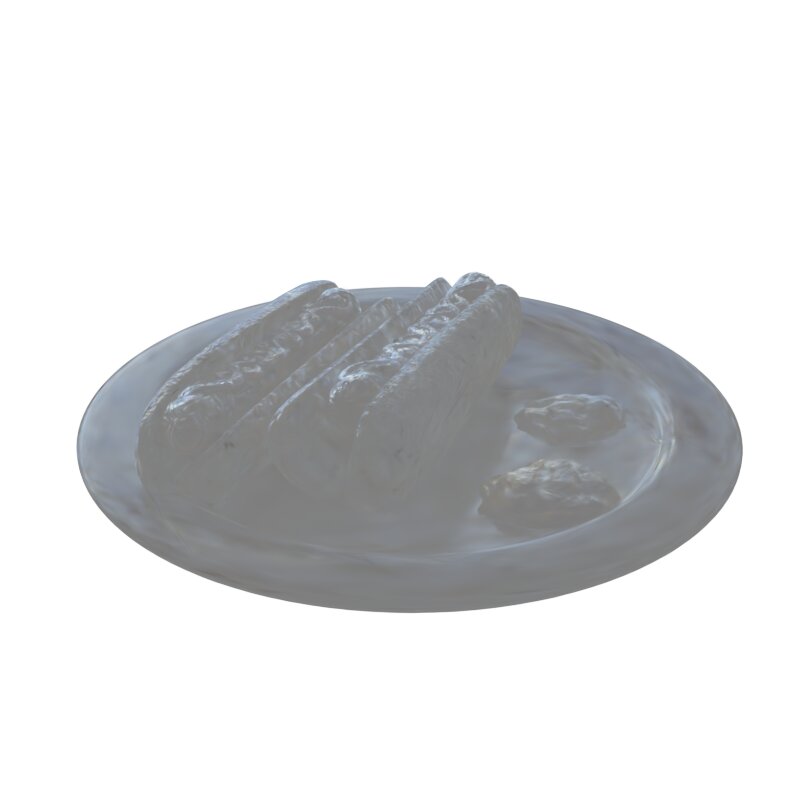}
    \includegraphics[trim={480px 260px 0px 300px}, clip, width=0.32\columnwidth]{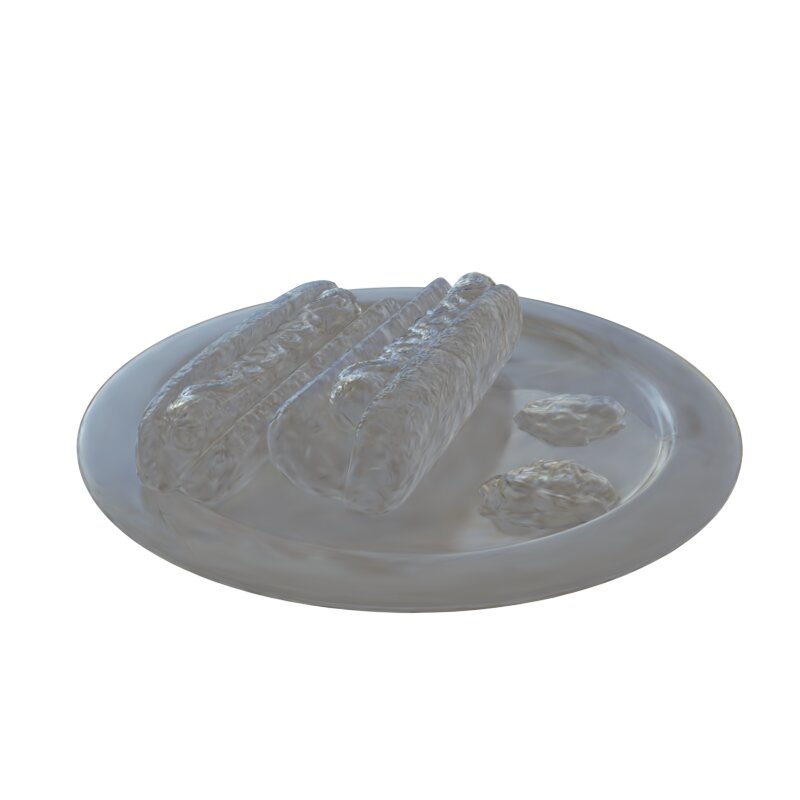}  \\
\vspace{-1mm}
    \caption{{\bf Specularity Comparison: Monoview vs. Multiview.} Training with a single viewpoint poses challenges in learning specularities due to limited sample diversity, multiview training  facilitates an easier understanding of specular effects.
    }
    \label{fig:multiview_helps_decomposition}
    \centering
    \vspace{-4mm}
\end{figure}

\subsection{Diffuse/Specular Decomposition Architecture}
\label{subsec:decoupled}

\noindent Separating diffuse $\Cdiff$ and view-dependent color $\Cspec$  (or specular) in independent MLPs~\cite{ireneCVPR2023,kuang2023palettenerf} is essential for ensuring view-dependent consistency in the edits. This separation reflects the underlying assumption that object color is primarily captured by the diffuse component which is reasonable for many common scenarios.
%It's worth noting that while we use the term ``specular component'', in practice, this component encompasses any view-dependent effect beyond specularity (\eg, Fresnel, etc.).

\noindent To model the view dependency, $\MLPdiff$ only utilizes the hashgrid features $\feat$, whereas $\MLPspec$ incorporates both the hashgrid features and the view direction $\viewd$. The outputs of these components are combined using a sigmoid  function $\sigma(\cdot)$, expressed formally as:
\begin{align}
	\Cdiff &= \MLPdiff(\feat | \phi_{df}) \\
	\Cspec &= \MLPspec(\feat, \viewd | \phi_{sp}, \phi_{df}) \\
	\Cgs &= \sigma(\Cdiff + \Cspec)
	\label{eq:color_sigmoid}
\end{align}
where  $\phi_{sp}$ and $\phi_{df}$ denote the specular and diffuse weights of the MLPs, respectively, and $\Cgs$ represents the RGB color of the Gaussian. 

\noindent The standard architecture of $\MLPspec$ and $\MLPdiff$ consists of three fully-connected layers. However, we found it necessary to introduce additional residual connections between them. Specifically, we connected the first layer of $\MLPdiff$ to the second layer of $\MLPspec$, and the second layer of $\MLPdiff$ to the third layer of $\MLPspec$. These connections are unidirectional, allowing only $\MLPspec$ to access the content of $\MLPdiff$.

\noindent The rationale behind this architecture is as follows: while $\MLPspec$ theoretically accounts for all view-dependent reflectance (since it receives $\viewd$ as input), $\MLPdiff$ should focus solely on learning what is view-point invariant.  Without these connections, the specular component tends to be noisier, making it challenging to learn an accurate reflection. With the added connections, the specular component gains a better understanding of which aspects of the reflection to prioritize.

\noindent To train the pair of MLPs we use the following loss function:
\begin{align}
	\mathcal{L} &= \lVert \Cgs - \rgbedit \rVert_{1} + \lambda_{spec} \lVert\Cspec\rVert_{2}^2
\end{align}
This loss function uses an $L_1$ photometric loss with a smoothness term on the specular component. The value of $\lambda_{spec}$ gradually decreases from $0.25$ to $0.05$ during training. Initially, this penalty encourages the network to prioritize explaining color variations using the diffuse component ($\Cdiff$) alone. Since the diffuse MLP ($\MLPdiff$) lacks view direction information, it cannot effectively capture view-dependent effects. As training progresses, we reduce the penalty on $\MLPspec$ to allow the network to also account for the view-dependent part.

%While the architecture of $\MLPspec$ and $\MLPdiff$ is the standard one with three fully-connected layers, we observed that it was necessary to add extra residual connections between them. In particular, we connected the second layer of $\MLPdiff$ to the third layer of $\MLPspec$. Importantly, these connections are uni-directional, and only $\MLPspec$ \textit{sees} the content of $\MLPdiff$. 

%While the architecture of $\MLPspec$ and $\MLPdiff$ is the standard one with three fully-connected layers, we observed that it was necessary to add extra residual connections between them. In particular, we connected the first layer of $\MLPdiff$ to the second layer of $\MLPspec$, and, the second layer of $\MLPdiff$ to the third layer of $\MLPspec$. Importantly, these connections are uni-directional, and only $\MLPspec$ \textit{sees} the content of $\MLPdiff$. 
%
%The intuition behind this architecture is that while $\MLPspec$ can in theory explain all the view-dependent reflectance (it receives $\viewd$ as input), $\MLPdiff$ should be restricted to learn what is invariant to the view-point. Further, not having the connections makes the specular component more noisy, struggling to learn a proper reflection. With the connections, the specular components knows better which parts of the reflection needs to focus on. 

%\Elena{let's see if we can build a figure that supports this claim} \Adrian{We have the diffuse and specular outputs of several images, if that works, then yes. Otherwise, we would need to do extra experiments.}
%

\subsection{Multi-view Training}
\label{subsec:multiview}

% This idea, which was previously used for NeRFs~\cite{mueller2022instant}, has two benefits. -- Wouldn't say that Instant-NGP's contribution was multiview
% First, it favors points in the 3D world with the same reflectance to be also close in the neural space,

\noindent An essential aspect of our method is the multi-view training strategy. Unlike existing approaches in Gaussian Splats~\cite{kerbl3Dgaussians,lee2023compact}, which utilize one image per batch, our method samples multiple views simultaneously. This strategy offers two key advantages. Firstly, it encourages points in the 3D world with similar reflectance properties to be close together in the feature space, thereby aiding convergence and enhancing overall quality. Secondly, it assists in implicitly decomposing the reflection into diffuse and specular components, which proves advantageous for the recoloring task.

\noindent We adapted the CUDA implementation of the fast rasterizer to use tiles from different images within the same training step.
Additionally, to expedite training time without altering the geometry during recoloring, 
we initialize the 3D Gaussians using C3DGS and freeze them. 
%
%we freeze the 3D Gaussians. 
This ensures interactivity, enabling faster operations as there's no need to recompute the gaussian ordering for alpha blending during rasterization. %We initialize the 3D gaussians using a pretrained C3DGS model and precompute the gaussian ordering per tile in each image. 
With the precomputed ordering, we create mini-batches composed of five tiles from different images, where the same gaussian is observed from diverse viewpoints. The training batch is comprised of as many 5-sample mini-batches as can be accommodated within the GPU's VRAM.

\noindent As we demonstrate in the experimental section,  this modification alone significantly enhances the default rendering quality of different baselines (see Table~\ref{tab:multiviewVSmono}). Additionally, multi-view training is essential for effectively learning a robust decomposition of both diffuse and view-dependent components. In the original 3DGS approach, training is conducted using a single image per batch, meaning each Gaussian is updated based on this singular viewpoint. However, by including five samples of each Gaussian within the same training batch, the update process endeavors to account for all these viewpoints simultaneously. This aids the $\MLPdiff$ in discerning the shared color aspects (diffuse) and the viewpoint-dependent characteristics (specular) among these five tiles. In Fig.~\ref{fig:multiview_helps_decomposition}, we provide two examples where the benefits of utilizing multi-view training for learning the specular components are evident.

%This modification alone significantly enhances the reconstruction quality compared to the baselines, as demonstrated in Table~\ref{tab:multiviewVSmono}. Both our method and the original C3DGS~\cite{lee2023compact} exhibit superior PSNR values when employing the multi-view training strategy. Notice in the table that combining our novel specular/diffuse architecture with the multi-view approach does not degrade the quality of the synthesized images compared to C3DGS, while providing valuable separation of the color components. 

%Indeed, multi-view training is essential for effectively learning a robust decomposition of both diffuse and view-dependent components. In the original 3DGS approach, training is conducted using a single image per batch, meaning each Gaussian is updated based on this singular viewpoint. However, by including five samples of each Gaussian within the same training batch, the update process endeavors to account for all these viewpoints simultaneously. This aids the $\MLPdiff$ in discerning the shared color aspects (diffuse) and the viewpoint-dependent characteristics (specular) among these five tiles. In \francescrmk{missing figure} Fig.~\ref{fig:multiview_helps_decomposition}, we provide two examples where the benefits of utilizing multi-view training for learning the specular components are evident.

\newcolumntype{Y}{>{\centering\arraybackslash}X}
\begin{table*}
\begin{center}
{
\begin{tabularx}\textwidth{XlY YYY YYY YYY}
\toprule
 & & \multicolumn{3}{c}{Mip NeRF 360~\cite{barron2022mipnerf360}} & \multicolumn{3}{c}{LLFF~\cite{mildenhall2019llff}} & \multicolumn{3}{c}{NeRF Synthetic~\cite{mildenhall2020nerf}} \\
\midrule
Method & & PSNR$\uparrow$ & SSIM$\uparrow$ & LPIPS$\downarrow$ & PSNR$\uparrow$ & SSIM$\uparrow$ & LPIPS$\downarrow$ &  PSNR$\uparrow$ & SSIM$\uparrow$ & LPIPS$\downarrow$ \\
\midrule

%\multicolumn{2}{l}{INGP (ref)}     & - & - & - & - &- & - & - & - & - \\
%\midrule
%\multicolumn{2}{l}{\textcolor{review}{SHs}} & 23.56 & 0.808 & 0.249  & 26.45 & 0.875 & 0.158 & 26.86 & 0.941 & 0.053\\

\multicolumn{2}{l}{PaletteNeRF}      & 22.24 & 0.720 & 0.278 & 24.19 & 0.800 & 0.187 & 29.92 & 0.960 & 0.038\\
\multicolumn{2}{l}{RecolorNeRF}     & - & - & - & 24.52 & 0.736 & 0.333 & 29.00 & 0.943 & 0.048\\
IReNe                              & & 26.80 & 0.714 & 0.292 & 26.51 & 0.823 & 0.136 & 30.33 & 0.959 & 0.035\\
\multicolumn{2}{l}{\vanillamethod}   & \underline{29.19} & \underline{0.881} & \underline{0.150} & \underline{28.01} & \underline{0.905} & \underline{0.129} & \textbf{32.13} & \textbf{0.975} & \textbf{0.023}\\
\methodname                        & & \textbf{29.74} & \textbf{0.886} & \textbf{0.146} & \textbf{29.10} & \textbf{0.915} & \textbf{0.119} & \underline{31.65} & \underline{0.971} & \underline{0.027}\\
\bottomrule
\end{tabularx}
}
\end{center}
\vspace{-3mm}
\caption{\small{\textbf{Quantitative color edition results.} \textbf{Best}, \underline{Second}. Results using the recolored ground truth images from~\cite{ireneCVPR2023}.}}
\vspace{-2mm}
\label{table:main}
\end{table*}

\newcolumntype{Y}{>{\centering\arraybackslash}X}
\begin{table}
%\vspace{-6mm}
\begin{center}
{
\begin{tabularx}\columnwidth{ccccccc}
\toprule
  & & & & Mip NeRF 360 & LLFF & NeRF Synthetic \\
\midrule
DC & MV & DS & & PSNR$\uparrow$ & PSNR$\uparrow$ & PSNR$\uparrow$ \\
\midrule

\xmark & \xmark & \xmark & & 29.19 & 28.01 & \textbf{32.13} \\
\xmark & \cmark & \xmark & & 29.45 & 28.54 & \underline{31.90} \\
\xmark & \cmark & \cmark & & 29.56 & 28.62 & 31.78 \\
\cmark & \xmark & \xmark & & 28.58 & 27.66 & 30.93 \\
\cmark & \cmark & \xmark & & \underline{29.61} & \textbf{29.15} & 31.50 \\
\cmark & \cmark & \cmark & & \textbf{29.74} & \underline{29.10} & 31.65 \\

\bottomrule
\end{tabularx}
}
\end{center}
\vspace{-3mm}
\caption{\small{\textbf{Ablation study. Recoloring results on all datasets of our ablated method}.  \textbf{Best}, \underline{Second}. \textbf{DC} indicates if the output is decoupled or not. \textbf{MV} indicates if the model has been trained using our multi-view rasterization \textbf{DS} indicates if the diffuse's network $\MLPdiff$ output has been introduced into the soft segmentation network $\MLPseg$.}
}
\vspace{-3mm}
\label{table:ablation}
\end{table}

\vspace{-4mm}

\subsection{Editing}\label{subsec:editing}

\noindent The recoloring process involves the user modifying a designated target view $\rgbedit$, followed by fine-tuning a portion of the architecture to accommodate the desired changes.
Drawing from~\cite{ireneCVPR2023}, we implement last-layer adaptation on the diffuse component (as per Equation~\ref{eq:last_layer}) and soft-segmentation (as per Equation~\ref{eq:irene_blending}). 
We update the weights $\phi^{\prime}_{df}$ of the last layer of $\MLPdiff$ to derive a new diffuse color $\Cdiff^{\prime}$. This new color is then alpha-blended with $\Cdiff$ using the soft-segmentation mask, denoted as $\alpha$, resulting in the final diffuse color denoted  $\Cdiff^{\prime \prime}$. It's worth noting that our method differs in the way the soft-segmentation mask is obtained compared to~\cite{ireneCVPR2023}.  In our approach, we establish a connection between the activations of last layer of the diffuse MLP, denoted as $\hdiff$, and the segmentation MLP. This integration aims to introduce additional contextual information to improve the quality of segmentation. As demonstrated in the experiments section, this enhancement contributes to the overall performance of our approach. 
\begin{align}
    \alpha &= \MLPseg (\feat,\hdiff), \\
    \Cdiff^{\prime} &= \MLPdiff(\feat | \phi^{\prime}_{df}), \\
    \Cdiff^{\prime \prime} &= \alpha  \Cdiff^{\prime} + (1- \alpha) \Cdiff, \\
    \Cgs^{\prime} &= \sigma(\Cdiff^{\prime \prime} + \Cspec)
\end{align}
Upon receiving the user's edit, we fine-tune the color and segmentation MLPs ($\MLPdiff$, $\MLPspec$, and $\MLPseg$) to accommodate the desired changes with the following loss function:
\begin{align}
	\mathcal{L} &= \lVert \Cgs^{\prime} - \rgbedit \rVert_{1}
\end{align}
This loss function uses an $L_1$ photometric loss between the rendered recolored version $\Cgs^{\prime}$ and the user's edit $\rgbedit$.

\section{Results} \label{sec:results}

\subsection{Quantitative results}

\begin{figure*}[t]
\begin{center}
\begin{tabularx}\textwidth{XXXX}
\centering \hspace{0mm} 1. Unedited GT & 
\centering \hspace{0mm} 2. Edited GT & 
\centering \hspace{0mm} 3. IReNe & 
\centering \hspace{-1mm} 4. \methodname
\end{tabularx}
\end{center}

     \centering
    \includegraphics[width=0.24\textwidth]{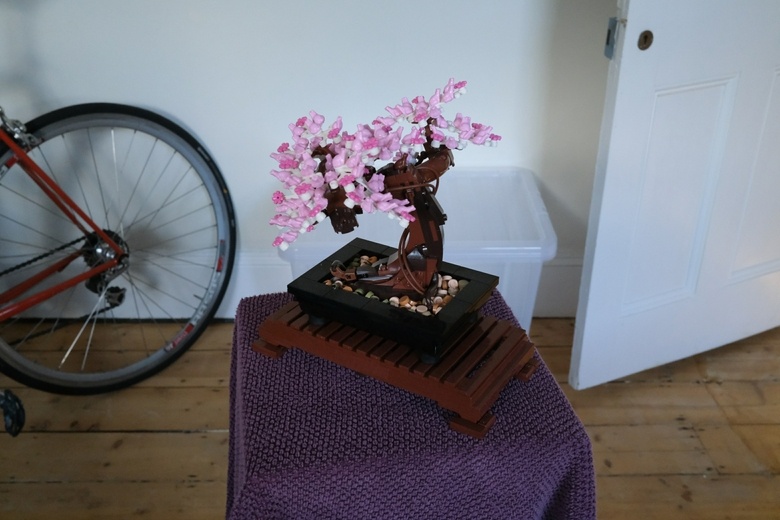}    \includegraphics[width=0.24\textwidth]{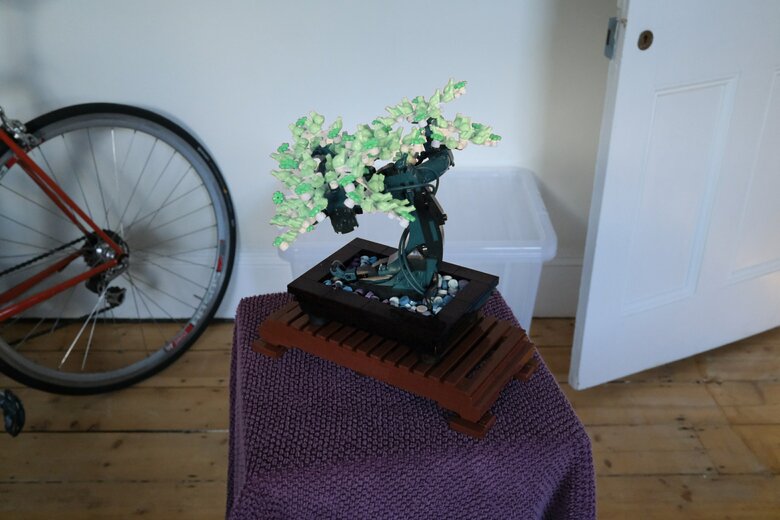}  
    \includegraphics[width=0.24\textwidth]{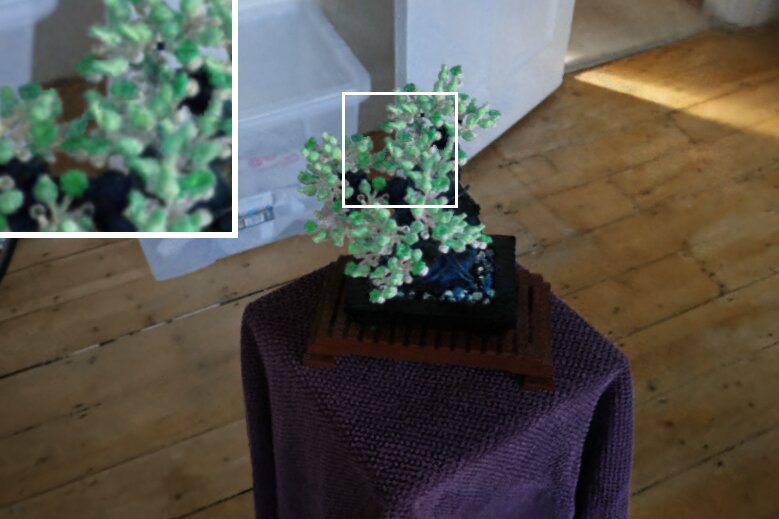} 
    \includegraphics[width=0.24\textwidth]
    {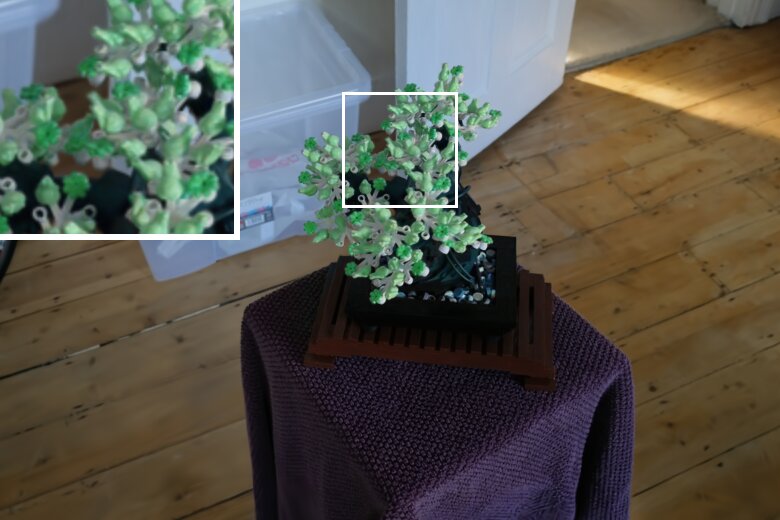} \\

    \centering
    \includegraphics[width=0.24\textwidth]{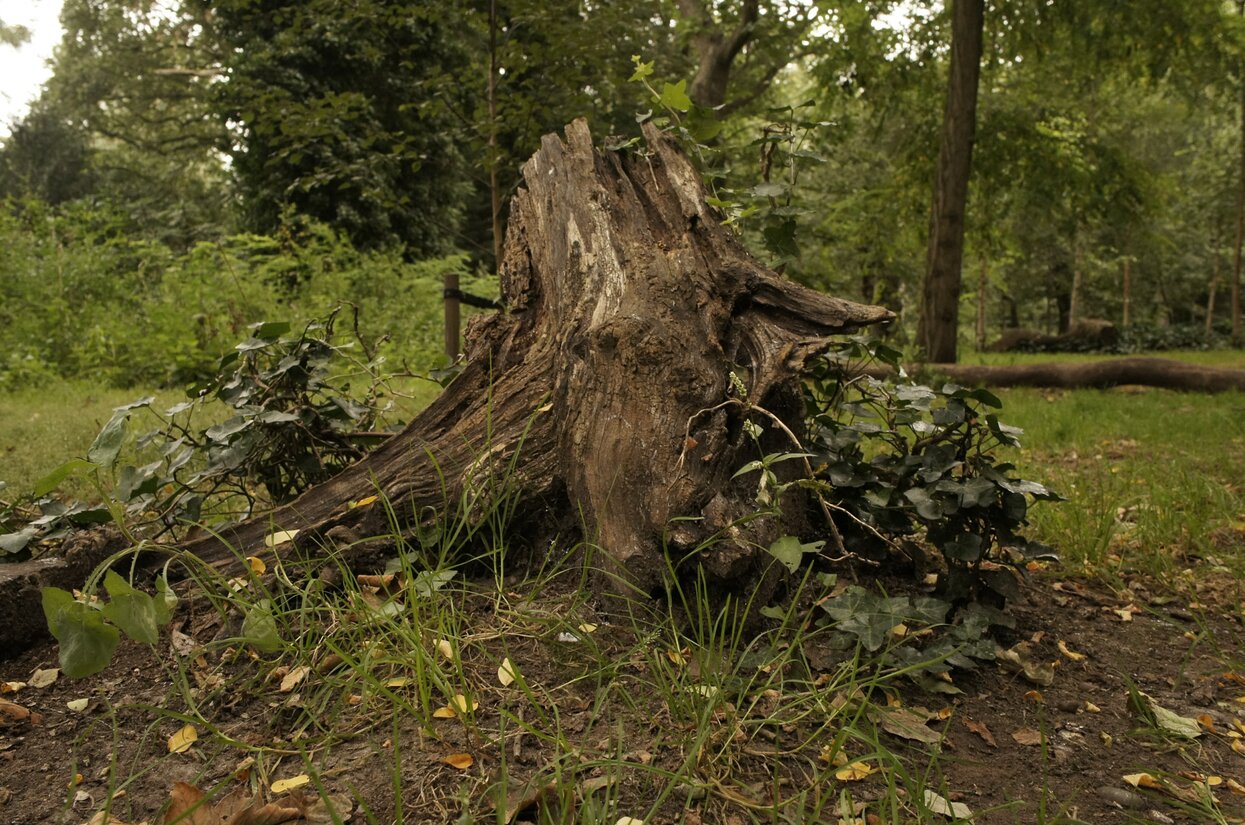}    \includegraphics[width=0.24\textwidth]{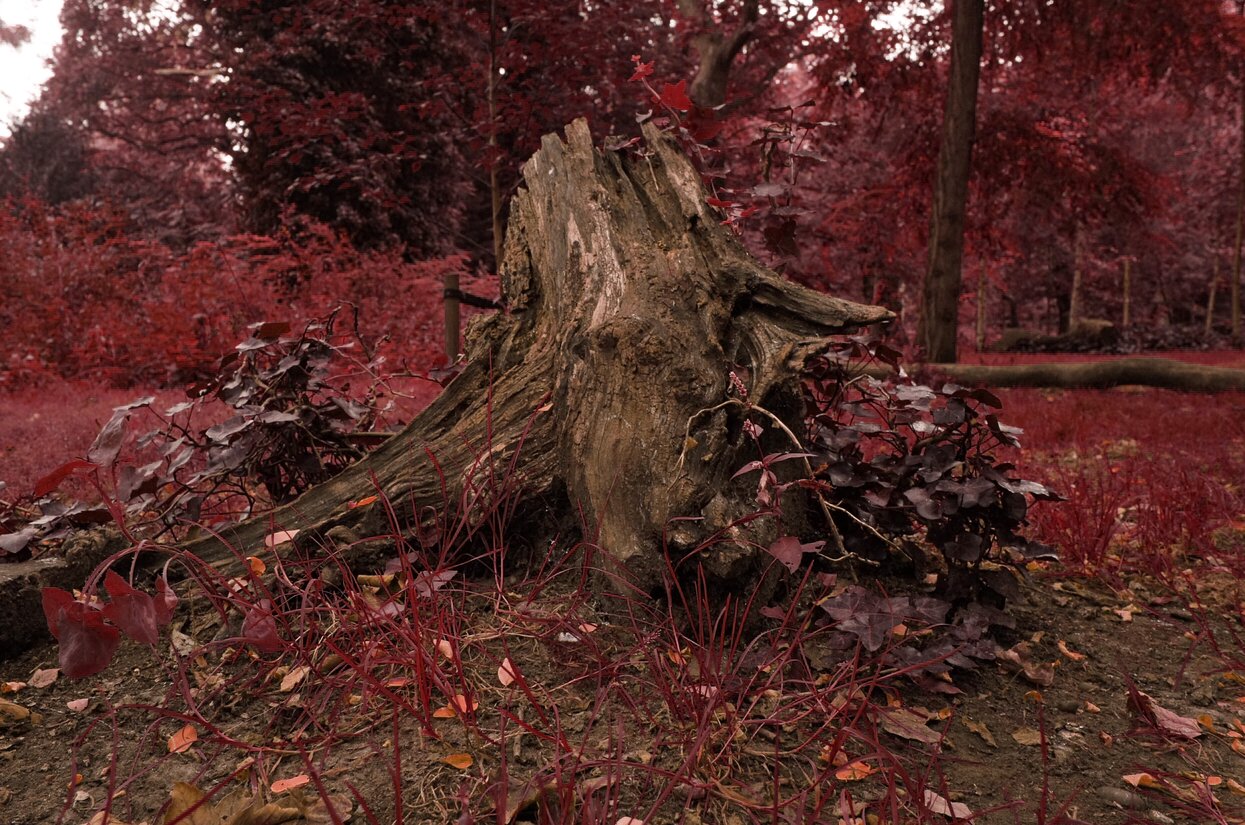} 
    \includegraphics[width=0.24\textwidth]{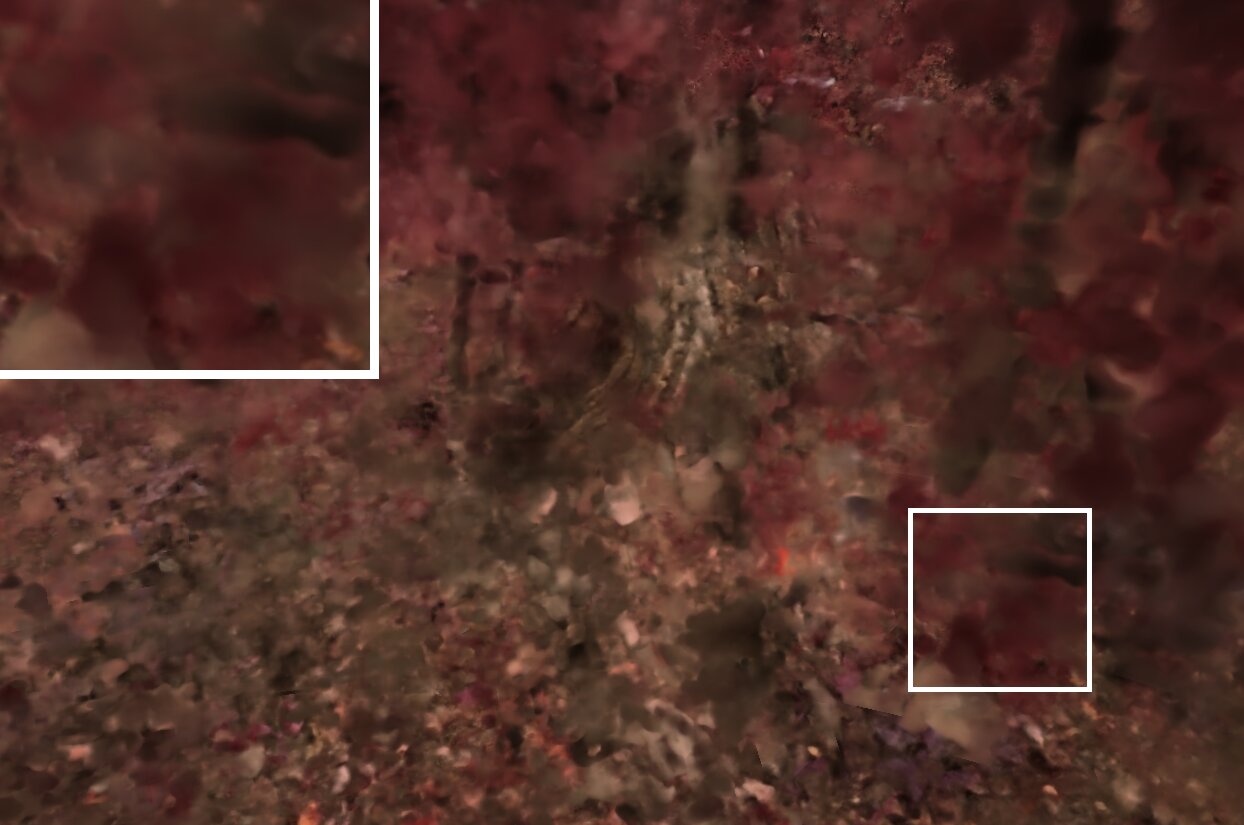} 
    \includegraphics[width=0.24\textwidth]{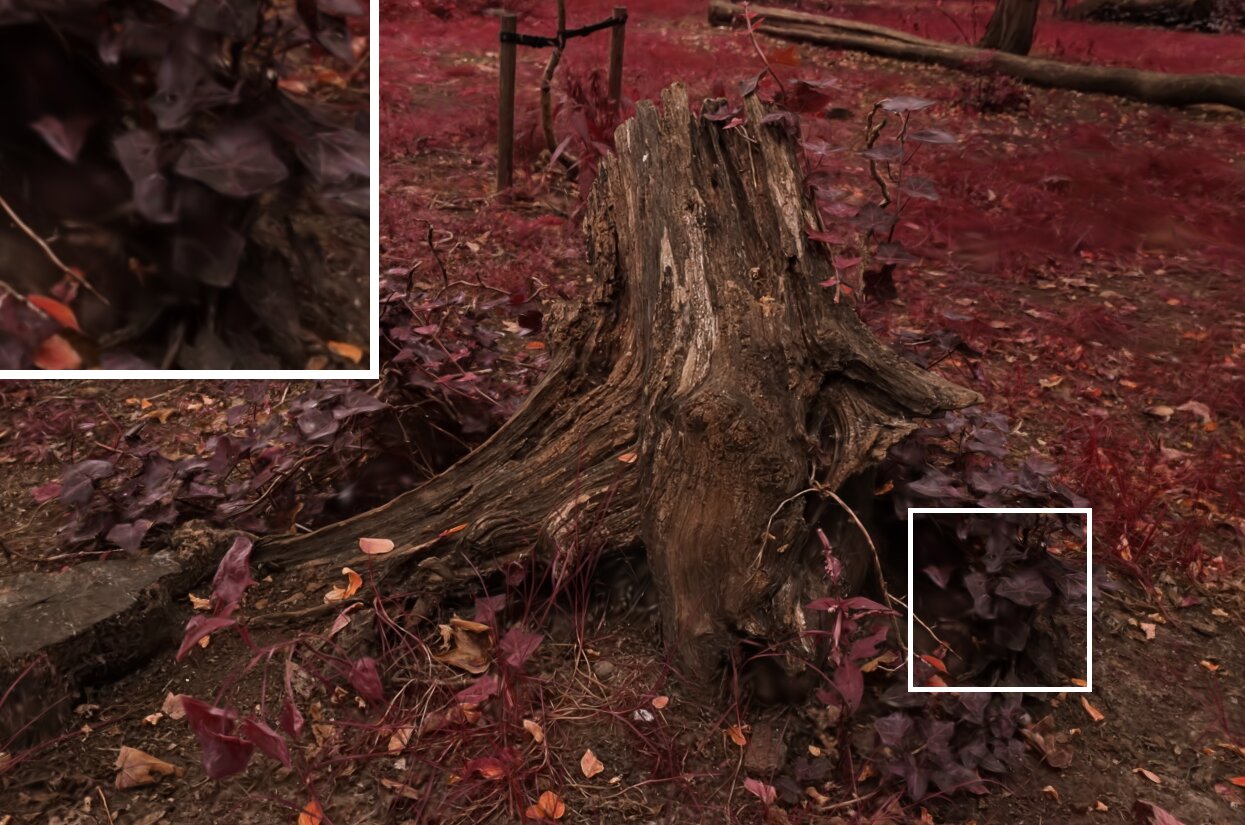} \\

    \centering
    \includegraphics[width=0.24\textwidth]{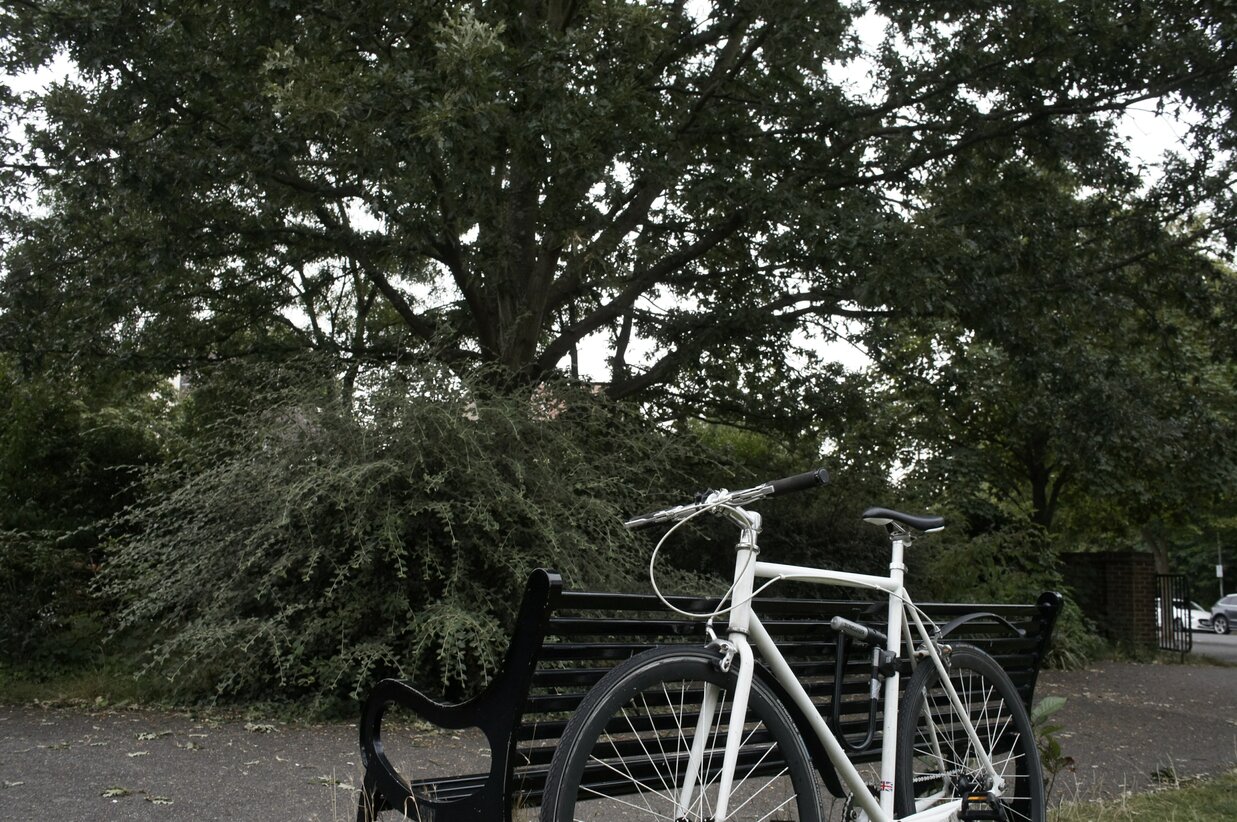}    \includegraphics[width=0.24\textwidth]{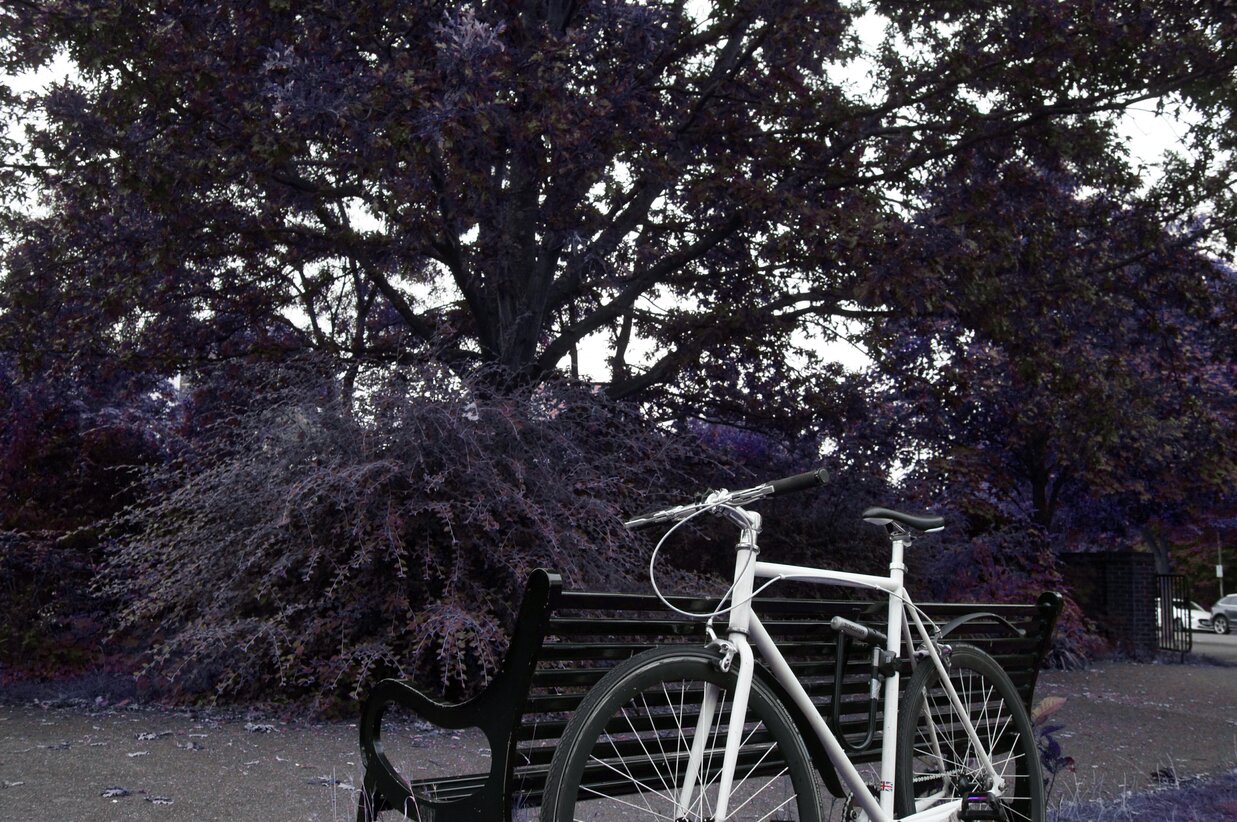} 
    \includegraphics[width=0.24\textwidth]{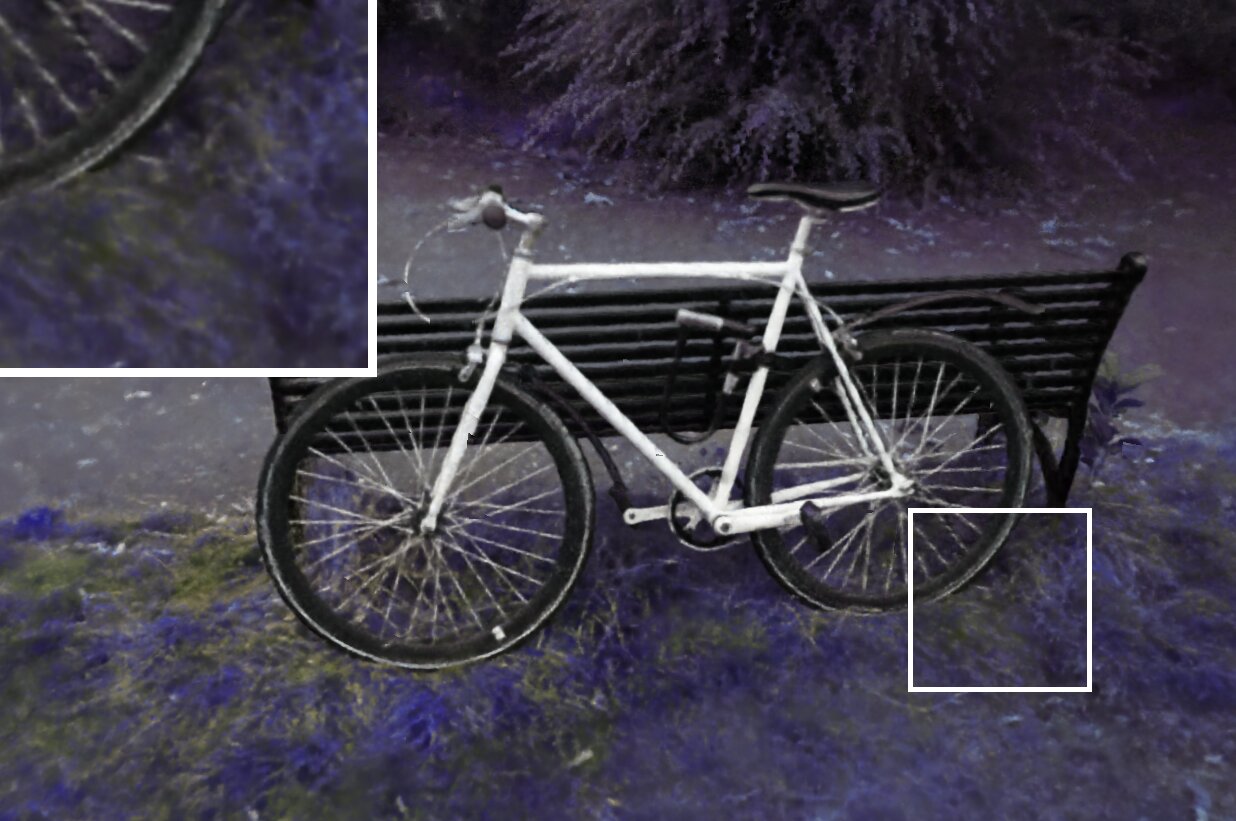} 
    \includegraphics[width=0.24\textwidth]{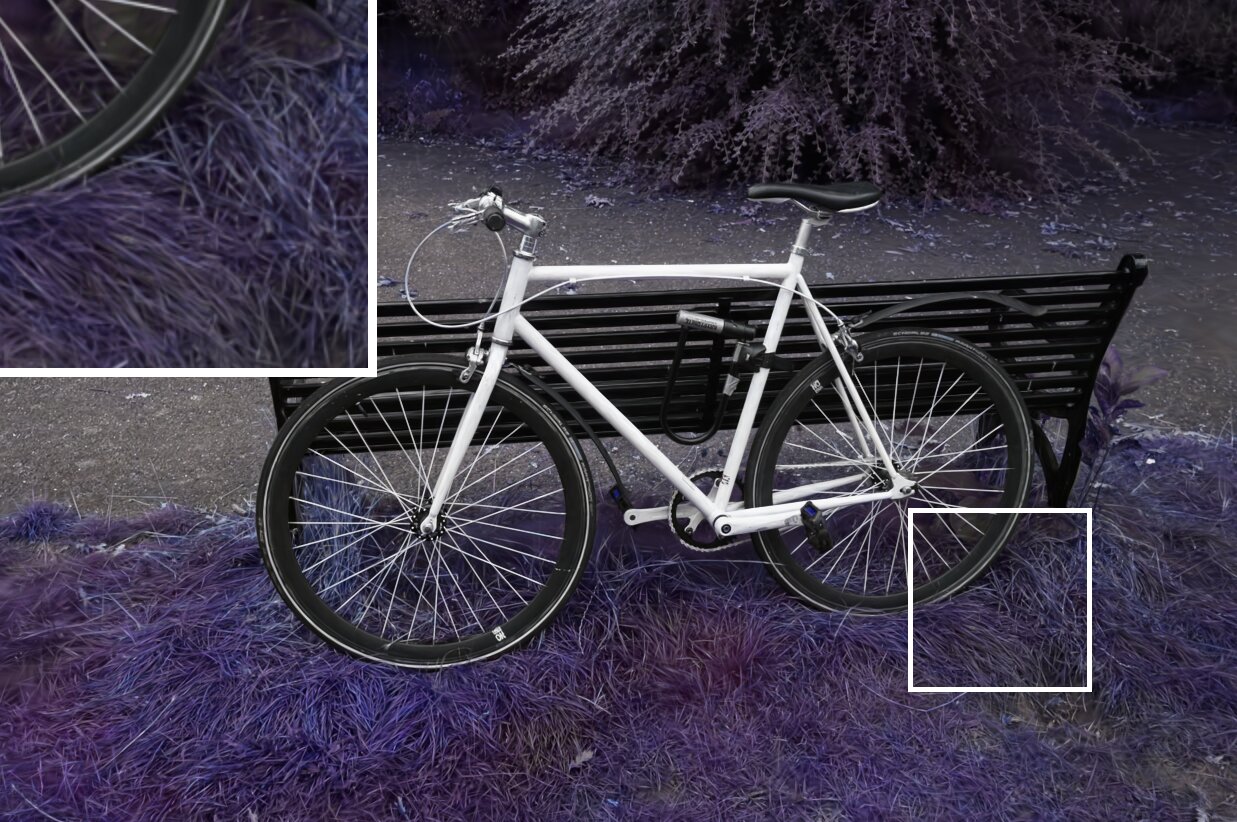} \\

    \centering
    \includegraphics[width=0.24\textwidth]{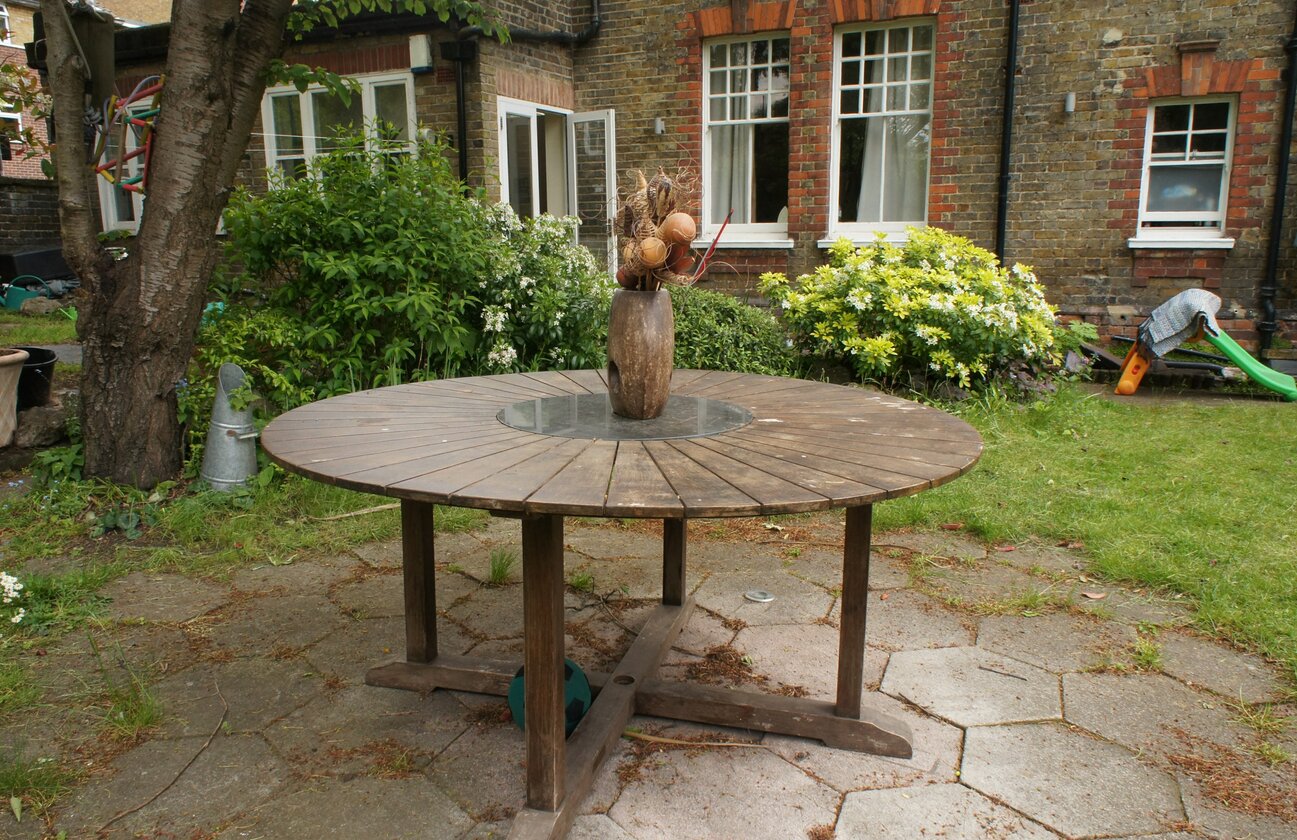}    \includegraphics[width=0.24\textwidth]{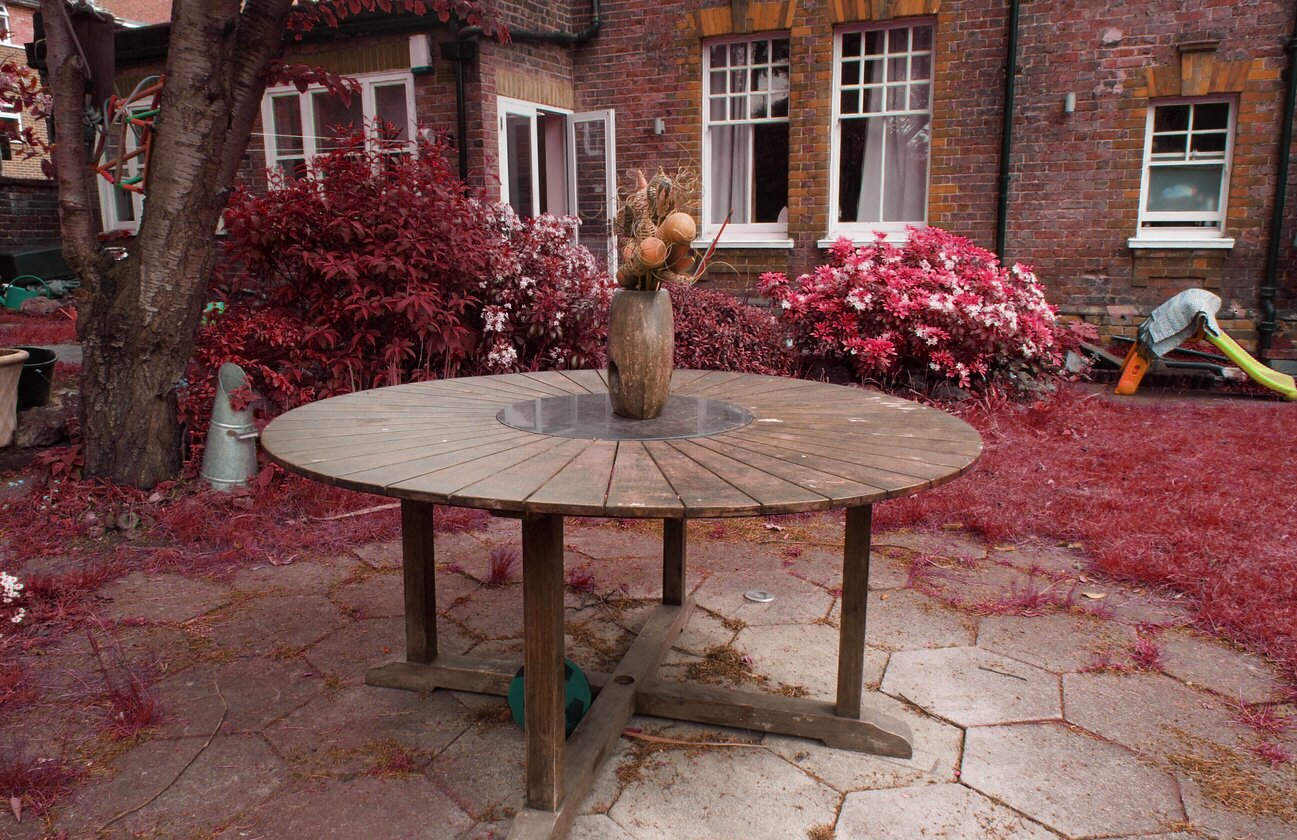} 
    \includegraphics[width=0.24\textwidth]{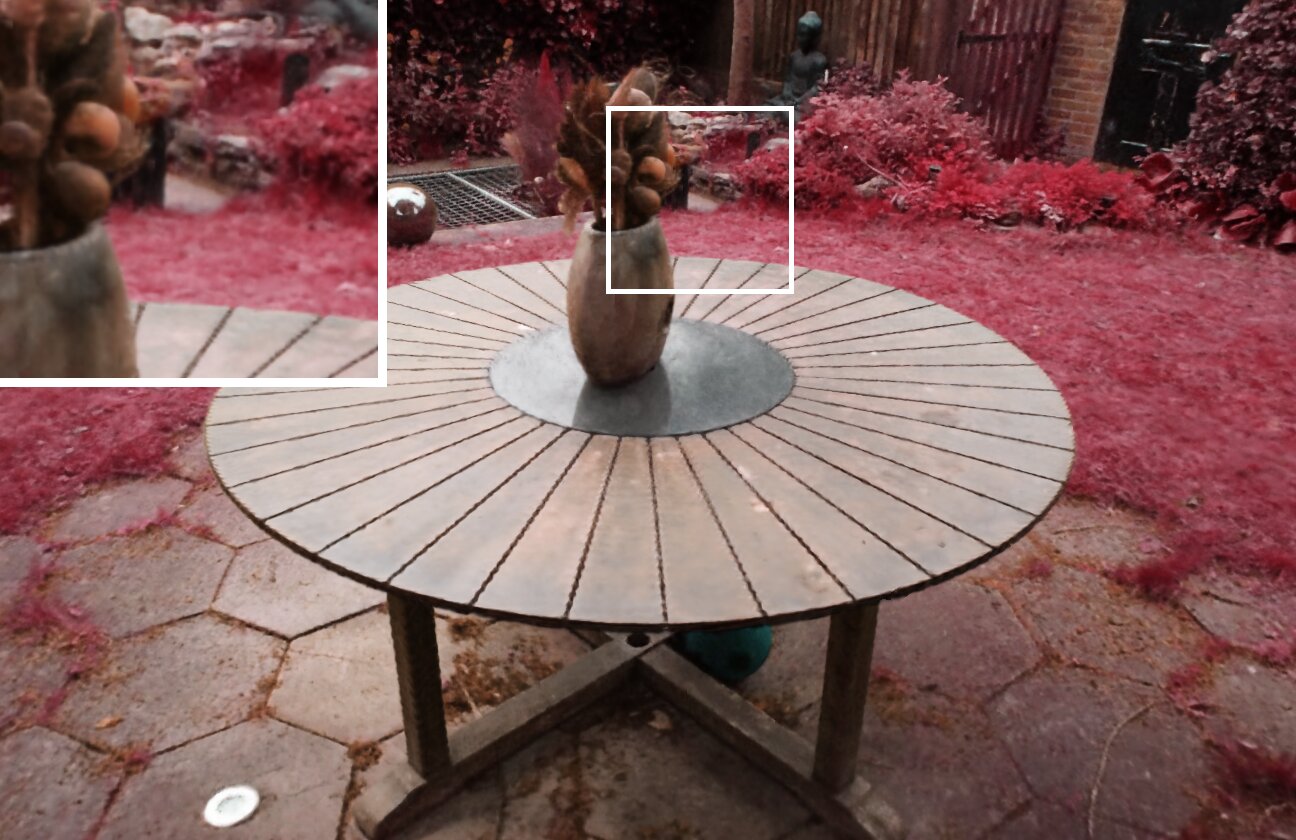} 
    \includegraphics[width=0.24\textwidth]{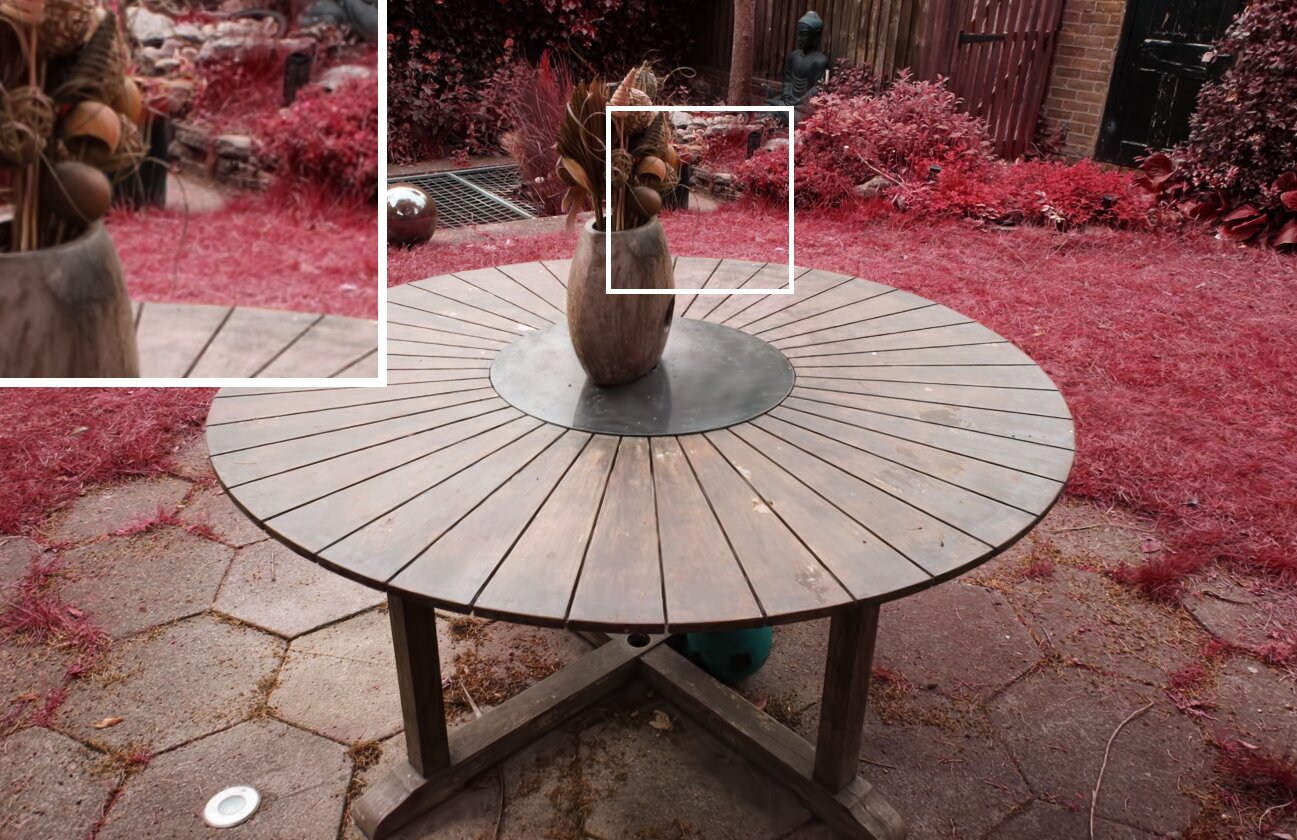} \\

        \centering
    \includegraphics[width=0.24\textwidth]{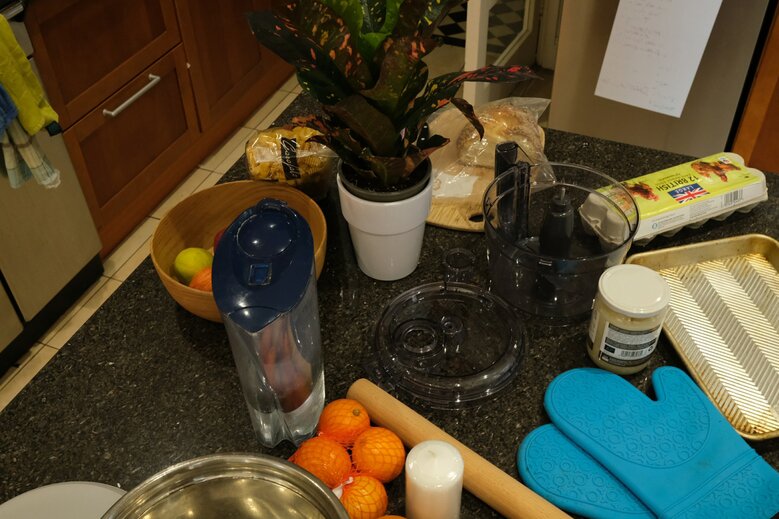}    \includegraphics[width=0.24\textwidth]{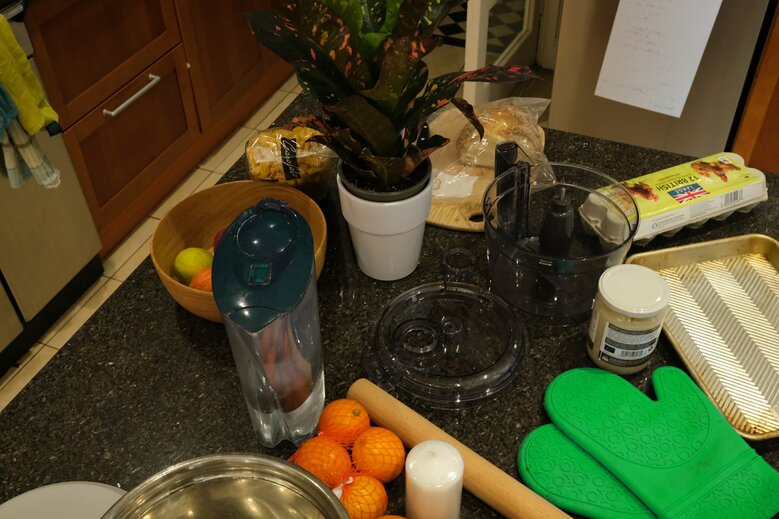} 
    \includegraphics[width=0.24\textwidth]{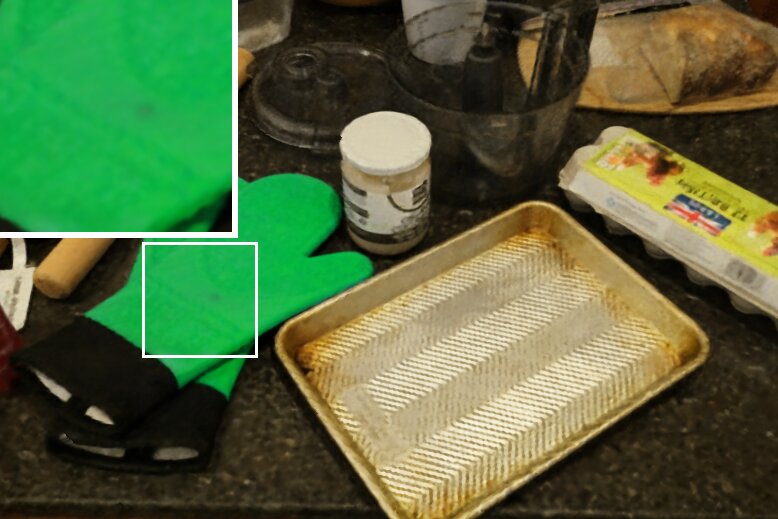} 
    \includegraphics[width=0.24\textwidth]{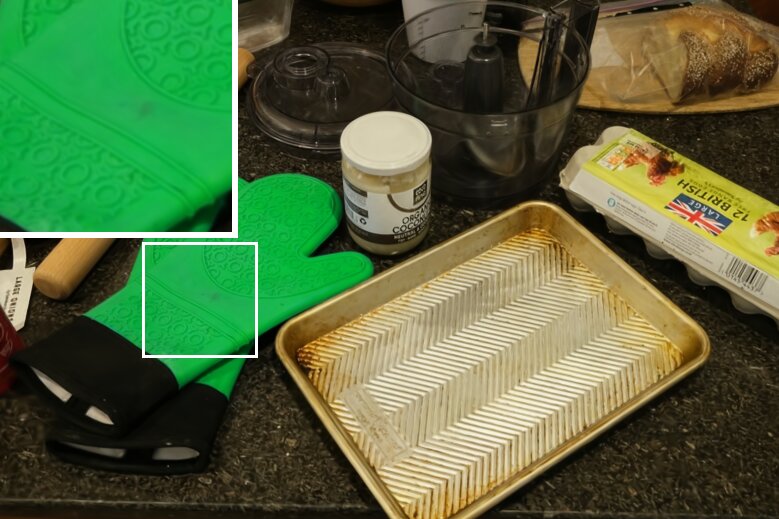} \\

    \centering
    \includegraphics[width=0.24\textwidth]{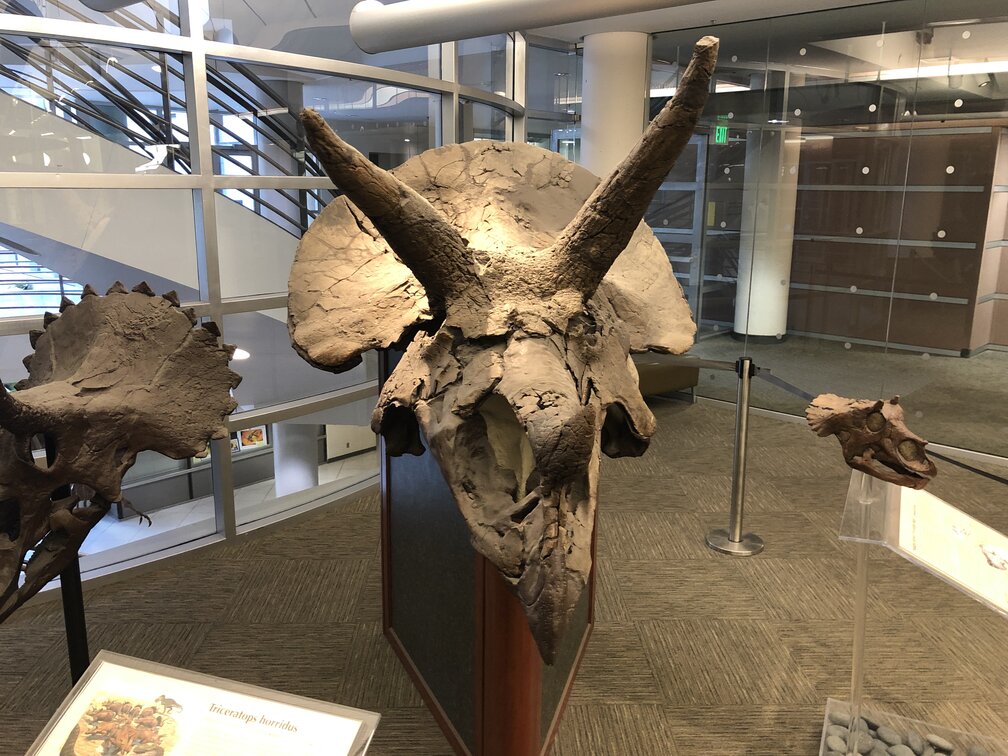}    \includegraphics[width=0.24\textwidth]{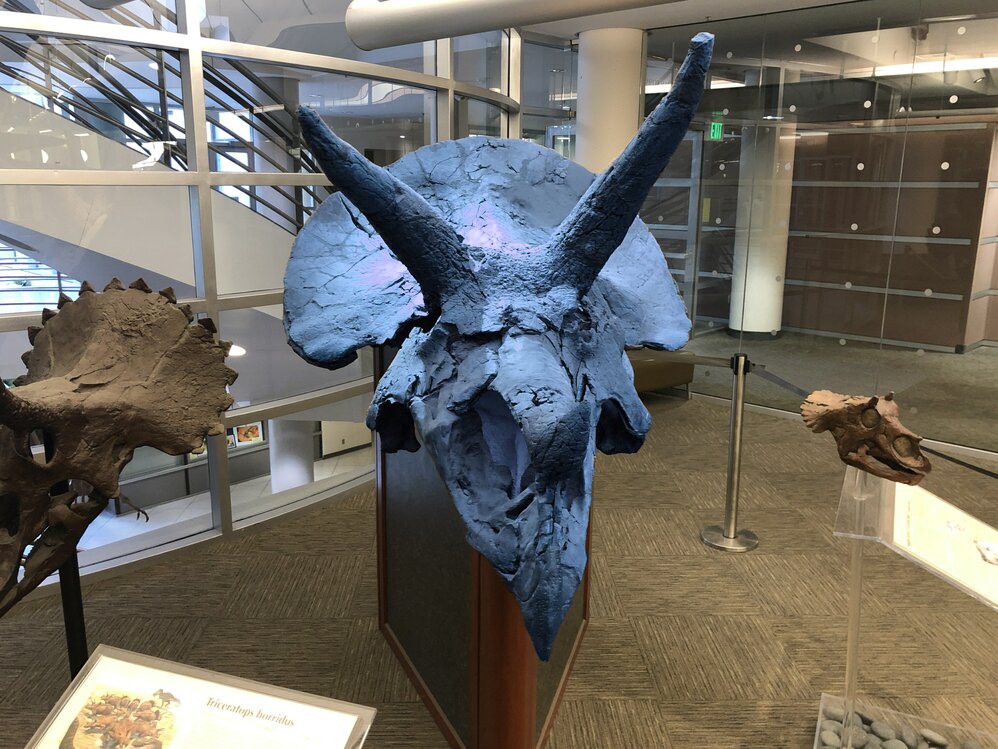}
    \includegraphics[width=0.24\textwidth]{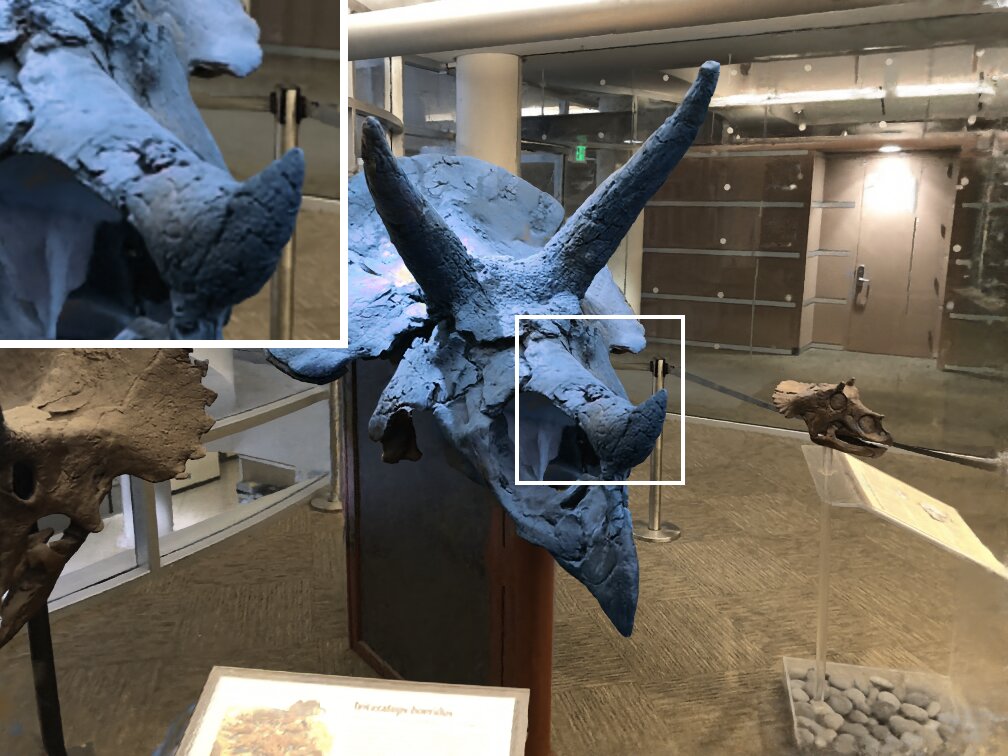} 
    \includegraphics[width=0.24\textwidth]{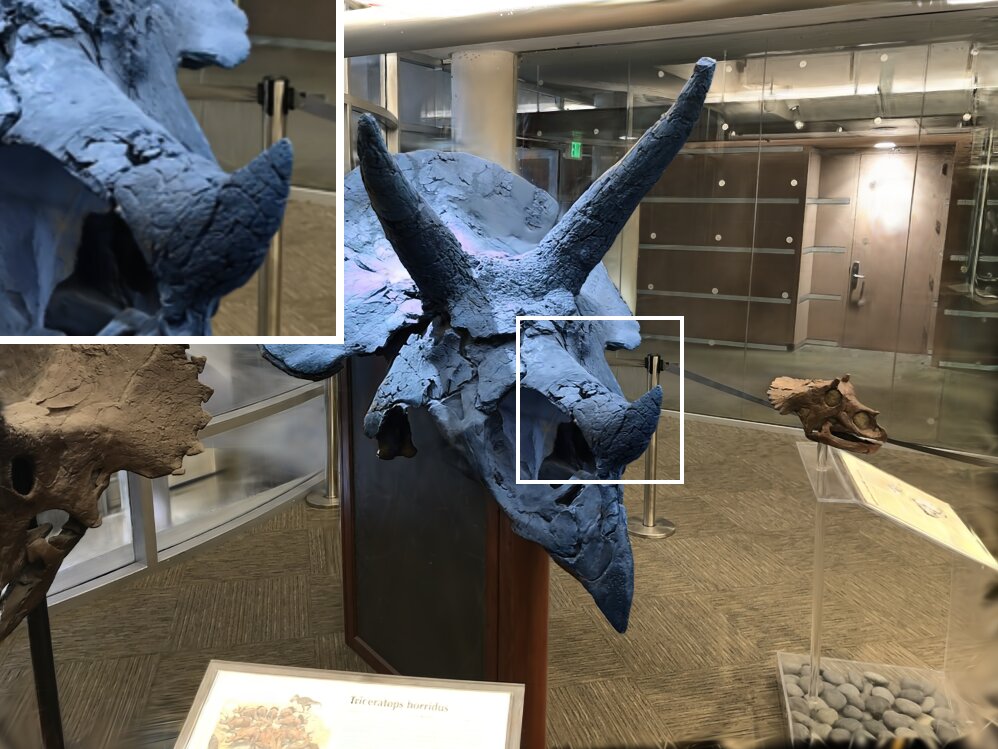} \\
    
    \centering
    \includegraphics[width=0.24\textwidth]{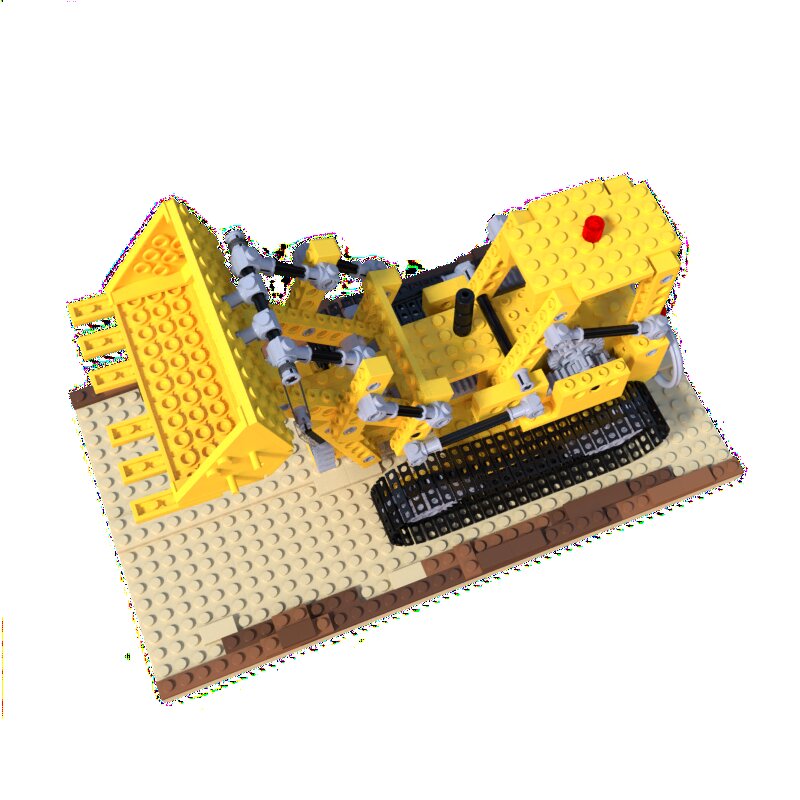}    \includegraphics[width=0.24\textwidth]{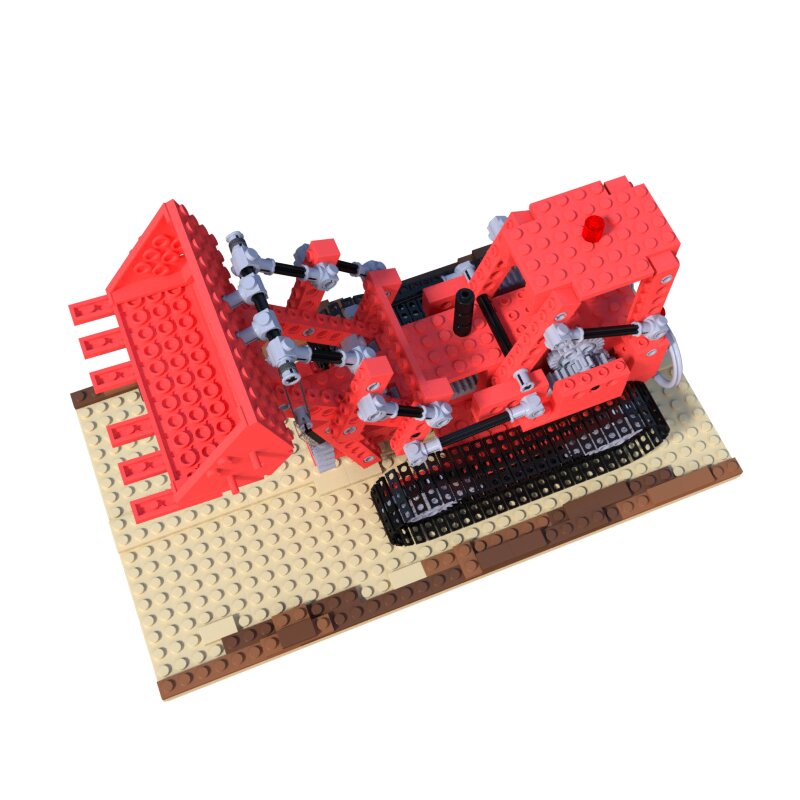}
    \includegraphics[width=0.24\textwidth]{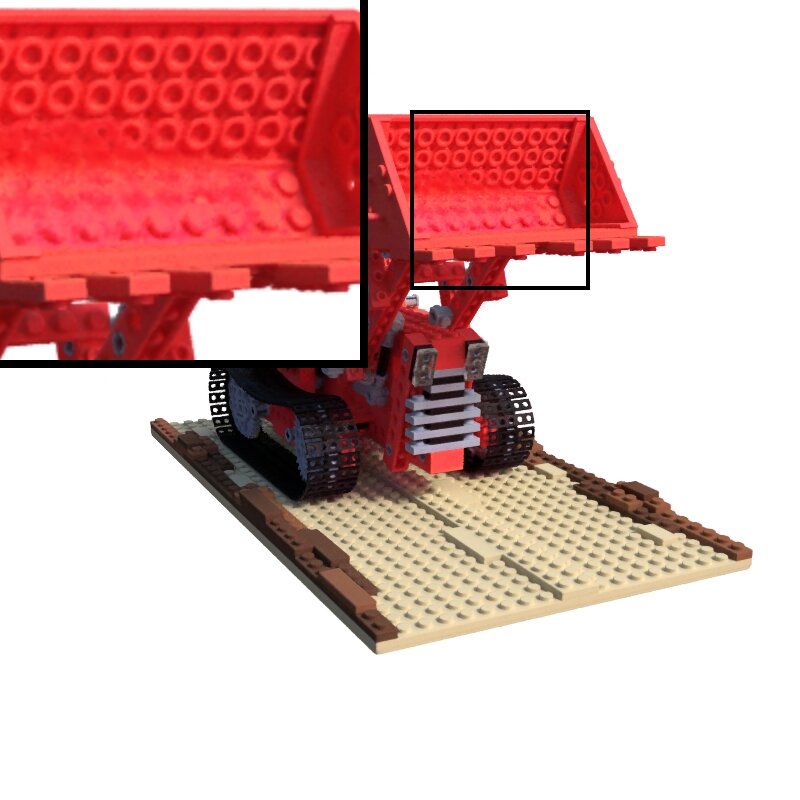} 
    \includegraphics[width=0.24\textwidth]{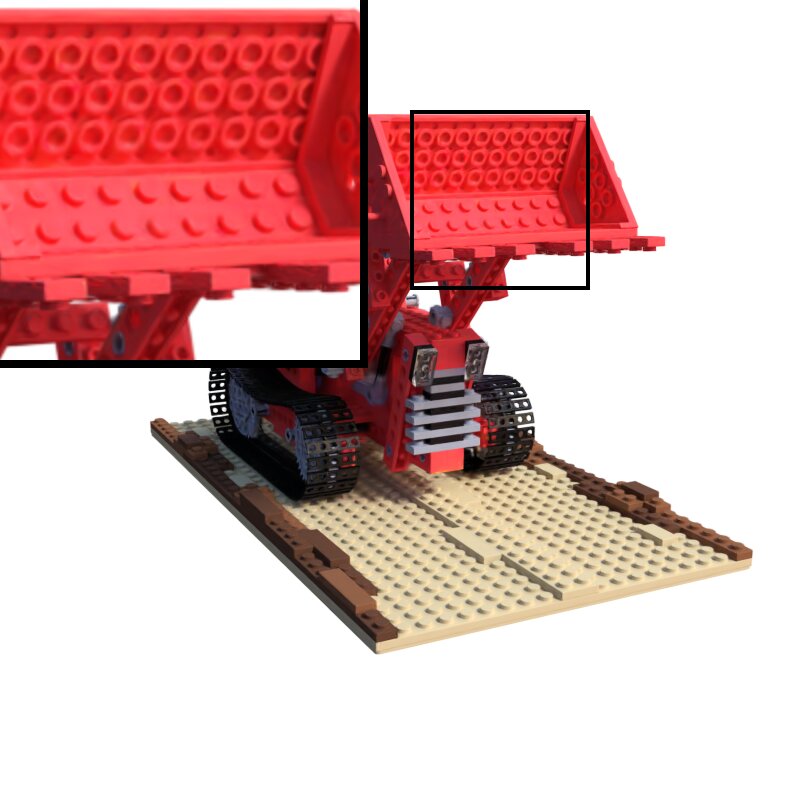} \\
    
    \caption{Some qualitative results of the IReNe\cite{ireneCVPR2023} dataset.}
    \label{fig:comparison_mip360}

\end{figure*}

\noindent{\bf Comparison with state-of-the-art.}  In Table~\ref{table:main}, we present the results of recoloring on the dataset provided by~\cite{ireneCVPR2023}, which includes manually recolored scenes using photoshop in LLFF~\cite{mildenhall2019llff}, MipNeRF-360~\cite{barron2022mipnerf360}, and recolored scenes from Synthetic NeRF~\cite{mildenhall2020nerf} using Blender. The original code from ~\cite{kerbl3Dgaussians} did not converge in the flower and leafs scenes from LLFF, so those scenes were not evaluated for any of the reported methods. We compare our results against several recent baselines, including PaletteNeRF~\cite{kuang2023palettenerf}, RecolorNeRF~\cite{gong2023recolornerf}, and IReNe~\cite{ireneCVPR2023}. We provide results from our proposed method, \methodname, as well as from the \vanillamethod. 
%\textcolor{review}{Additionally, we provide another baseline, referred to as Spherical Harmonics (SHs), where we perform gradient descent on the non-view-dependent component of the SHs learned in 3DGS. Further details are provided in the supplementary material.}
In Fig.~\ref{fig:comparison_mip360} some qualitative results from the dataset can be seen. 
The evaluation metrics employed are standard radiance field metrics (PSNR, SSIM, and LPIPS), and the split between training and testing, as well as the recoloring editions, follows the protocol established in~\cite{ireneCVPR2023}.
Importantly, \methodname~demonstrates a notable and consistent improvement over other radiance field recoloring methods, underscoring its efficacy and robustness.

%In Table.~\ref{table:main} we show results on recoloring on the dataset provided by~\cite{ireneCVPR2023}. These are recolored scenes from Synthetic NeRF~\cite{mildenhall2020nerf}, LLFF~\cite{mildenhall2019llff} and MipNeRF-360~\cite{barron2022mipnerf360}. We show results compared to several recent baselines: PaletteNeRF~\cite{kuang2023palettenerf}, RecolorNeRF~\cite{gong2023recolornerf} ad IReNe~\cite{ireneCVPR2023}. We also provide results from our method {\methodname} as well as the results from \vanillamethod. It can be seen that {\methodname} introduces a significant improvement with respect to other radiance field recoloring methods. The metrics used are the standard radiance field metrics (PSNR, SSIM and LPIPS) the split between training and testing and the recoloring editions are the same as the one in the original work~\cite{ireneCVPR2023}. 

\begin{table}[t]
\centering
\begin{tabular}{lccc}
\hline
Dataset (PSNR$\uparrow$)    & Mip NeRF 360     & LLFF              & NeRF Synthetic   \\ \hline
Mono-view C3DGS             & 28.96            & 25.63             & 35.60            \\ \hline
Multi-view C3DGS            & \textbf{29.47}   & \textbf{25.87}    & \textbf{35.84}    \\ \hline
Mono-view \methodname       & 28.43            & 25.73             & 33.87             \\ \hline
Multi-view \methodname      & \textbf{29.30}   & \textbf{25.97}    & \textbf{35.76}    \\ \hline
\end{tabular}
\vspace{-1mm}
\caption{\textbf{Multi-View VS Mono-View novel view synthesis}. Training both the original C3DGS as well as {\methodname} using the proposed multi-view approach enhances the novel view synthesis results. We underline that \methodname~also provides a decomposition of the image into diffuse and specular.
%\francescrmk{I'd remove the following claim. 0.87 vs 0.51 is not that significant.}In {\methodname}, the difference of using multi-view training in Mip NeRF 360 yields an average difference of $0.87$ PSNR, while for C3DGS it is of $0.51$, this illustrates how critical multi-view information is to learn the best color decoupling available.
}
\label{tab:multiviewVSmono}
\vspace{-1mm}
\end{table}

\begin{comment}

\newcolumntype{Y}{>{\centering\arraybackslash}X}
\begin{table*}
\begin{center}
{
\begin{tabularx}\textwidth{XlYYYYYYYYYYY}
\toprule
 & & \multicolumn{3}{c}{NeRF Synthetic~\cite{mildenhall2020nerf}} & \multicolumn{3}{c}{LLFF~\cite{mildenhall2019llff}} & \multicolumn{3}{c}{Mip NeRF 360~\cite{barron2022mipnerf360}} \\
\midrule
Method & &  PSNR$\uparrow$ & SSIM$\uparrow$ & LPIPS$\downarrow$ & PSNR$\uparrow$ & SSIM$\uparrow$ & LPIPS$\downarrow$ &  PSNR$\uparrow$ & SSIM$\uparrow$ & LPIPS$\downarrow$ \\
\midrule

\multicolumn{2}{l}{}          & - & - & - & - &- & - & - & - & - \\
\midrule
C3DGS Mono-view         & & 35.67 & - & - & 25.68 & - & - & 29.00 & - & -\\
C3DGS Multi-view        & & 35.89 & - & - & 25.90 & - & - & 29.46 & - & -\\
IREGA Mono-view         & & 33.95 & - & - & 25.74 & - & - & 28.47 & - & -\\
IREGA Multi-view        & & 35.83 & - & - & 25.94 & - & - & 29.34 & - & -\\
\bottomrule
\end{tabularx}
}
\end{center}
%\vspace{-4mm}
\caption{\small{\textbf{Reconstruction.} }}
%\vspace{-3mm}
\label{table:main}
\end{table*}

\end{comment}

\vspace{1mm}
\noindent{\bf Method robustness.} The dataset provided by~\cite{ireneCVPR2023} contains 11 distinct recolored images for each scene in the MipNeRF-360 and LLFF datasets. To demonstrate the robustness of our approach to variations in the input image, we conducted 11 evaluations for each scene, each time using a different recolored ground-truth image as input. This methodology allows us to assess that the quality of our method remains consistent regardless of the specific input image chosen by the user for editing.
\noindent The average recoloring PSNR for 11 iterations when modifying the edited image for \methodname~was (MipNeRF-360: $29.27$, LLFF: $28.87$), while for \vanillamethod~it was (MipNeRF-360: $28.71$, LLFF: $27.91$). Remarkably, the average performance in both cases closely aligns with the standard experiment in Table~\ref{table:main}, using the images defined by~\cite{ireneCVPR2023} as the input image. %\francescrmk{mmm it may sound weird that we have the same performance as IRENE, isn't it?} \ferrmk{I think Adrian meant that the quality of the output varies very few regarding the image chosen to recolor in table 1 (which is the standard procedure set by IRENE). I tried to rewrite it.}

%\noindent{\bf Method robustness.}  The dataset provided by ~\cite{ireneCVPR2023} includes 11 different recolored images for each of the scenes in MipNeRF-360 and LLFF datasets. To show that our approach is robust to which of the recolored ground-truth images is used as input, we evaluated our method 11 different times on each scene, changing the input image each time. In this way, we can assess that the final quality of our method does not depend on the image the user selects to edit.
% We also performed an experiment to show the robustness of our approach to the selection of the edition image. We run the experiments on both datasets with real images, MipNeRF-360 and LLFF, and performed the recoloring changing the training image, we computed all combinations for both datasets. The average recoloring PSNR when modifying the edited image for {\methodname} was (MipNeRF-360: $29.27$, LLFF: $28.87$), and for {\vanillamethod} was (MipNeRF-360: $28.71$, LLFF: $27.91$). In both cases the average performance is similar to the one from the standard experiments in ~\cite{ireneCVPR2023}.
%The average recoloring PSNR for the 11 iterations when modifying the edited image for {\methodname} was (MipNeRF-360: $29.27$, LLFF: $28.87$), and for {\vanillamethod} was (MipNeRF-360: $28.71$, LLFF: $27.91$). In both cases the average performance is similar to the one from the standard experiments in ~\cite{ireneCVPR2023}.

\vspace{1mm}
\noindent{\bf Ablation study.} We analyze the individual contribution of each component with two ablation experiments. The first experiment (Table~\ref{table:ablation}), analyzes the impact of 
%First, showcased in Table~\ref{table:ablation}, delves into three key aspects of our method: 
the decoupled output (\textbf{DC}), the multi-view rasterization (\textbf{MV}), and the integration of diffuse neuron activation into the segmentation network (\textbf{DS}). 
%This analysis serves as an ablation of the recoloring results presented in Table~\ref{table:main}. 
Notably, our method \methodname~emerges as the best-performing recoloring method overall. 
Adding multi-view information (\textbf{MV}) consistently enhances performance across all configurations. Moreover, the decoupled architecture (\textbf{DC}) demonstrates superior results with the utilization of multi-view. Remarkably, when multi-view is employed, the decoupled results consistently outperform the original non-decoupled color estimation. Additionally, inputting the diffuse neuron activations to the segmentation (\textbf{DS}) boosts performance in all cases except for a singular instance in the LLFF dataset. This outcome is reasonable, as the LLFF dataset comprises frontal recordings with limited viewpoint changes, thereby posing challenges in understanding diffuse and specular aspects of the scene.

%\vspace{1mm}
%\noindent{\bf Ablation study.} To analyze the contribution of each of the parts of the propose method to the final quality, we have performed two ablation experiments. The first is shown in Table~\ref{table:ablation}, where three main contributions of our method, the decoupled output \textbf{DC}, the multi-view rasterisation \textbf{MV}, and the introduction of the diffuse neurons activation to the segmentation network \textbf{DS}, are analyzed. The experiment is an ablation of the recoloring results of Table~\ref{table:main}. We can see that the best overall recoloring method is the full pipeline of {\methodname}. Multiview \textbf{MV} information improves performance in all configurations. As it may be expected, decoupled output learning \textbf{DC} requires multiview to be able to properly separate diffuse and specular components of the final color. Remarkably, when multiview is used the decoupled results are always better than the original non-decoupled color estimation. The use of the diffuse neuron activation pattern to enhance the segmentation \textbf{DS} increases performance in all cases except for a single case in the LLFF dataset. This is understandable as the LLFF dataset consists of frontal recordings that do not provide enough viewpoint changes from which to understand diffuse and specular parts of the scene.

\noindent In our second ablation experiment (Table~\ref{tab:multiviewVSmono}), we demonstrate how multiview optimization 
not only enhances the quality of the recoloring but also inherently 
improves the modeling of view-dependent effects in 3DGS. Specifically, we compare the reconstruction accuracy (without recoloring) of the original C3DGS~\cite{lee2023compact} trained with and without multiview, and apply the same procedure to evaluate \methodname. 
%
%However, it's important to note that for \methodname, we assess overall rendering quality performance of the original scene rather than recoloring. 
Notably, multiview consistently yields performance improvements across all cases. Furthermore, this experiment confirms that the proposed modifications to C3DGS~\cite{lee2023compact} for color editing do not lead to any noticeable reduction in the original quality.

%On our second ablation experiment, we show how multiview optimization improves not only the quality of our recoloring approach but also inherently yields a better view-dependent effects modeling in with gaussian splatting, see Table.~\ref{tab:multiviewVSmono}. In this experiment we use the standard split between training and testing from the original datasets, thus the slight deviation in metrics. We show a comparison between the original C3DGS~\cite{lee2023compact} trained with and without multiview information, and, the same procedure is followed for {\methodname}. However, note that for {\methodname} we are not testing recoloring but the overall rendering quality performance of the original scene. We can see that multiview consistently provides an increase in performance in all cases. Finally, this experiment also proof that the proposed modifications to C3DGS~\cite{lee2023compact} to allow for color edition, does not have any noticeable reduction on the original quality.% What can also be seen in this approach is that by adapting the gaussian splatting approach to allow for recoloring to be possible there is not a drop in rendering quality.

\begin{figure}[t]
    \centering
    \includegraphics[trim={0 10mm 0 0mm}, clip, width=0.325\columnwidth]{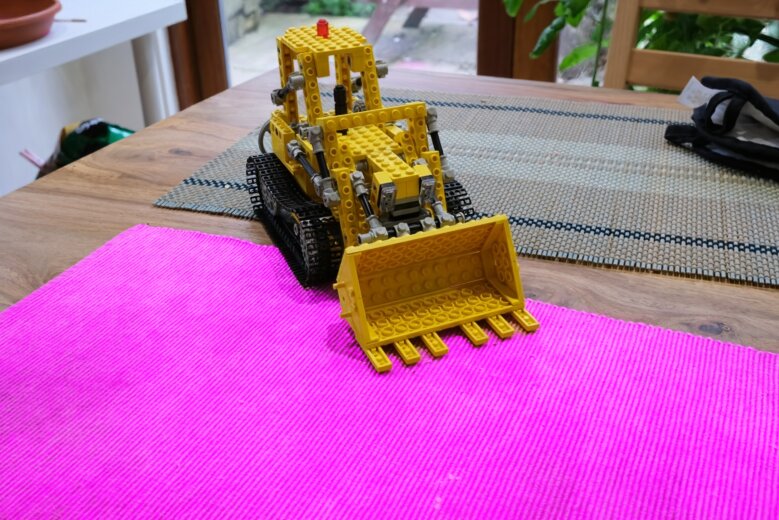} 
    \includegraphics[trim={0 10mm 0 0mm}, clip, width=0.325\columnwidth]{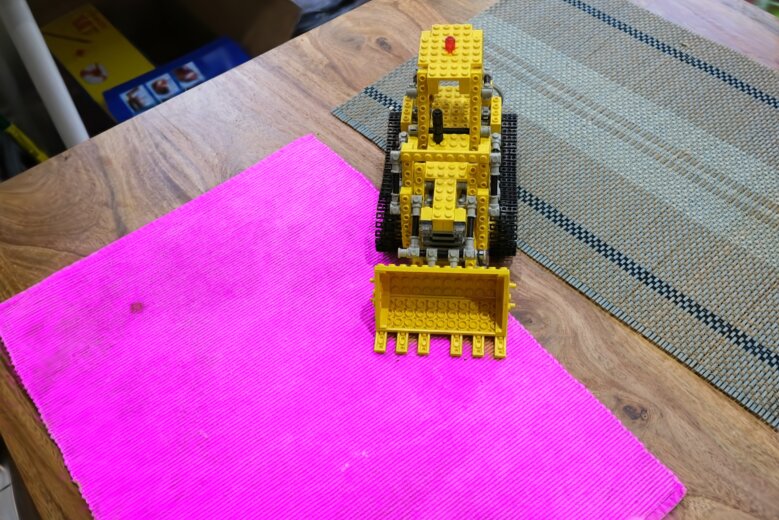}
    \includegraphics[trim={0 10mm 0 0mm}, clip, width=0.325\columnwidth]{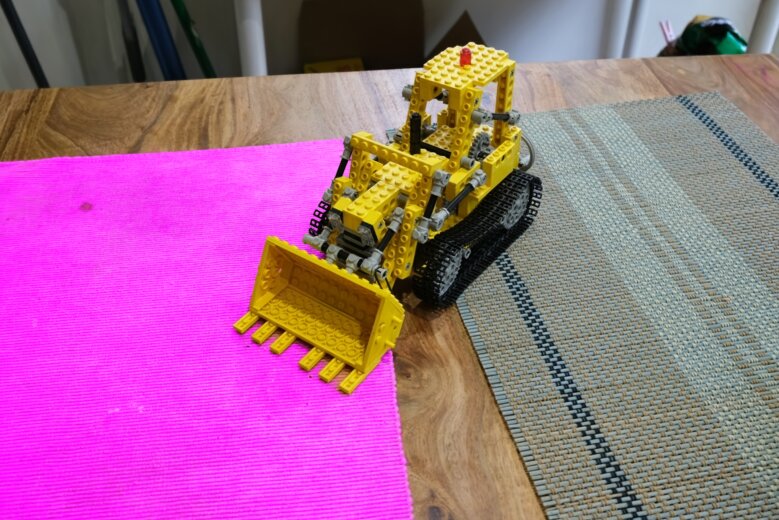}  \\
    \includegraphics[trim={0 10mm 0 0mm}, clip, width=0.325\columnwidth]{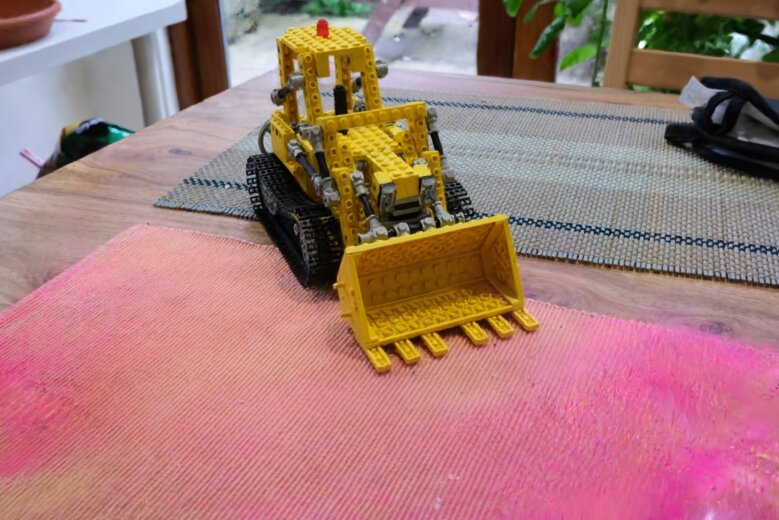} 
    \includegraphics[trim={0 10mm 0 0mm}, clip, width=0.325\columnwidth]{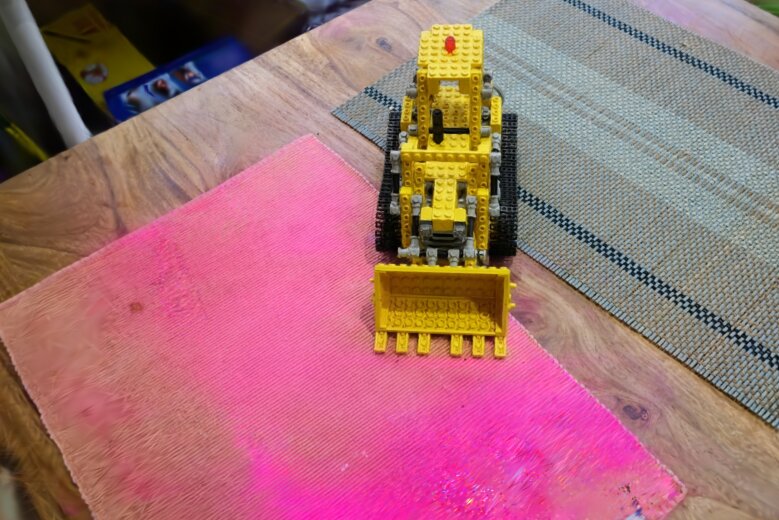}
    \includegraphics[trim={0 10mm 0 0mm}, clip, width=0.325\columnwidth]{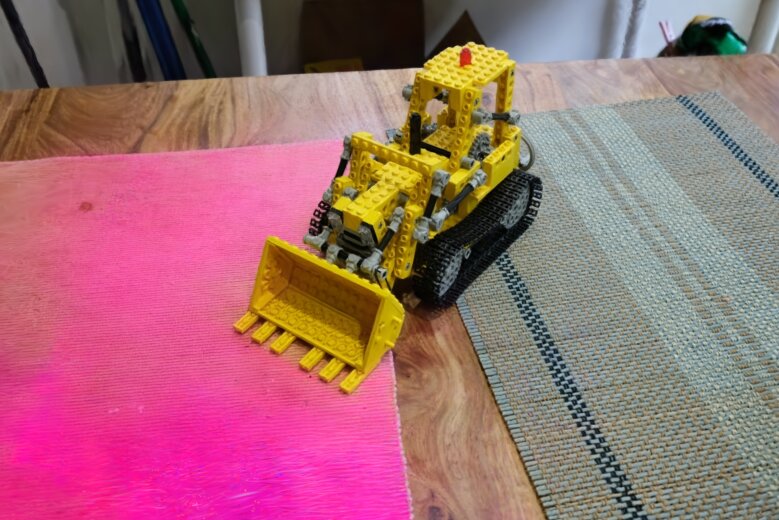}  \\
    
    \includegraphics[trim={0 10mm 0 0mm}, clip, width=0.325\columnwidth]{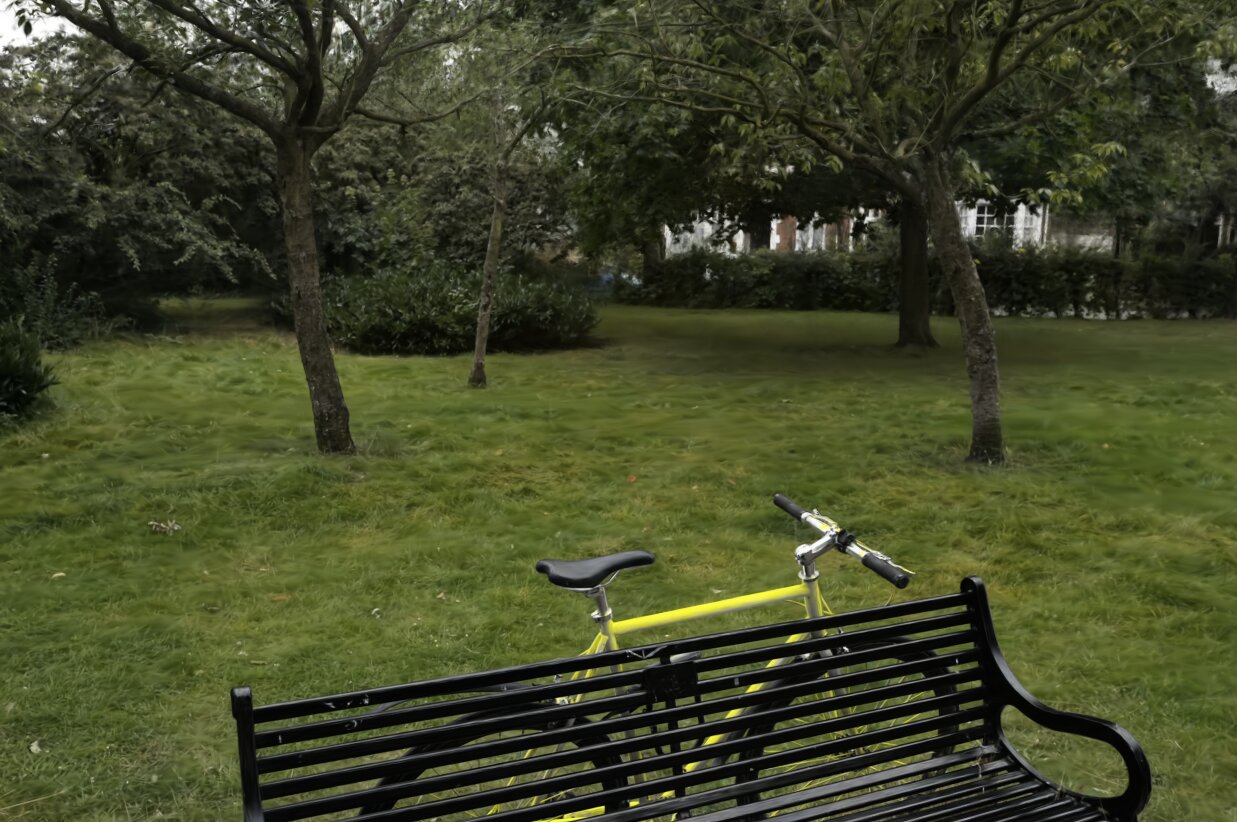} 
    \includegraphics[trim={0 0mm 0 10mm}, clip, width=0.325\columnwidth]{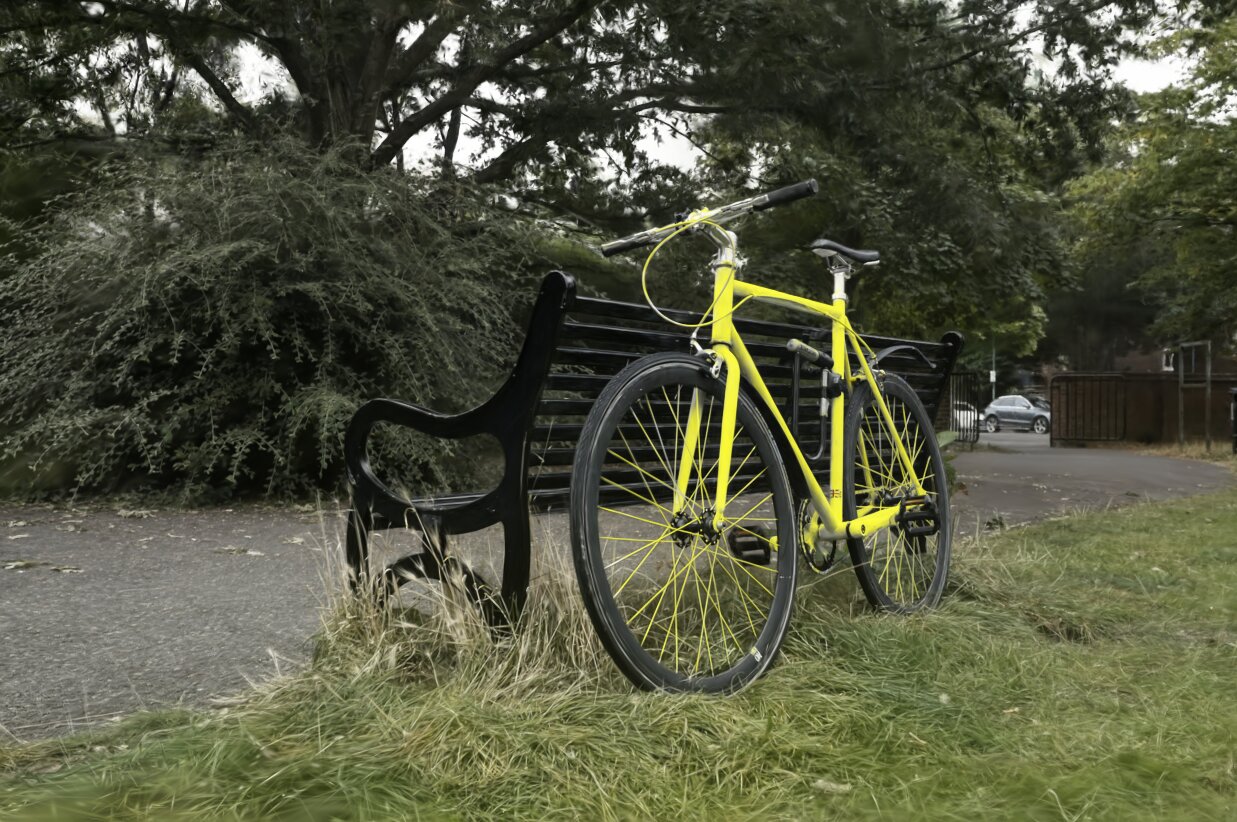}
    \includegraphics[trim={0 10mm 0 0mm}, clip, width=0.325\columnwidth]{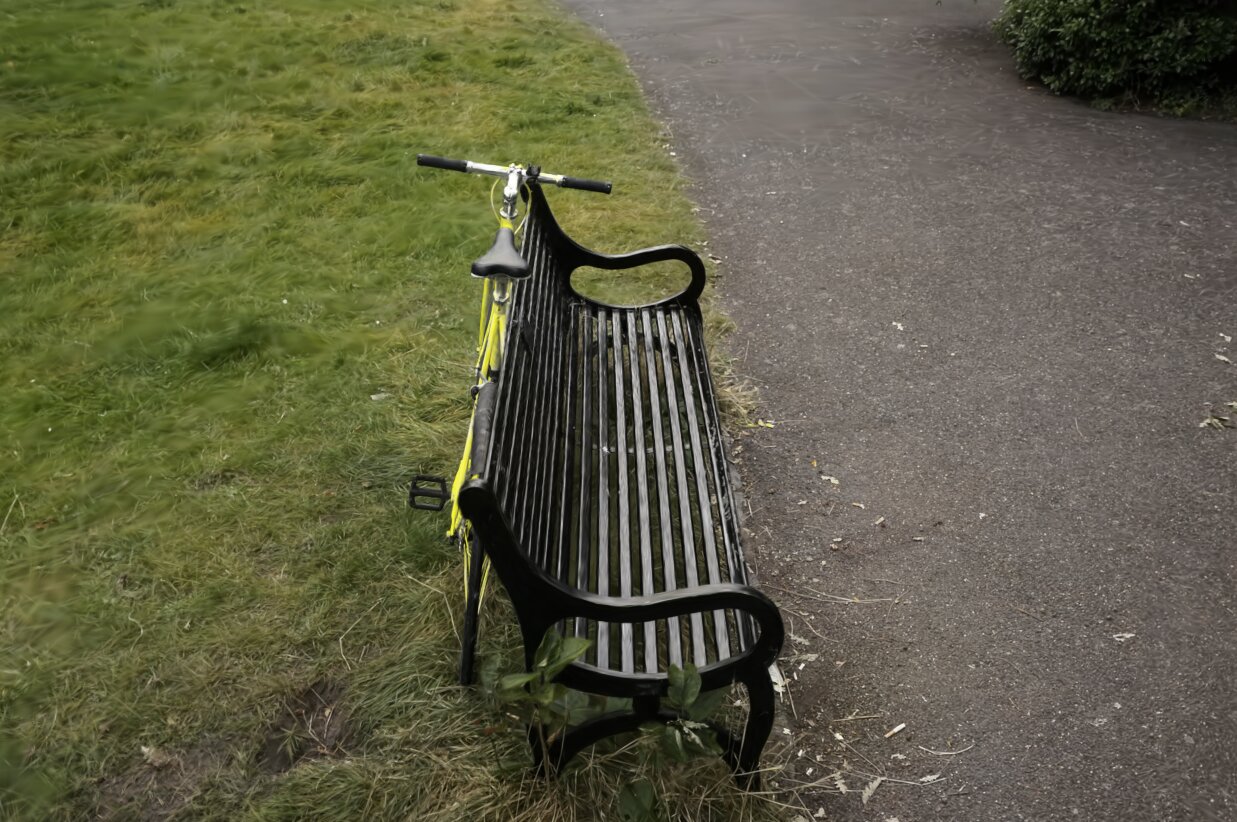}  \\
    \includegraphics[trim={0 10mm 0 0mm}, clip, width=0.325\columnwidth]{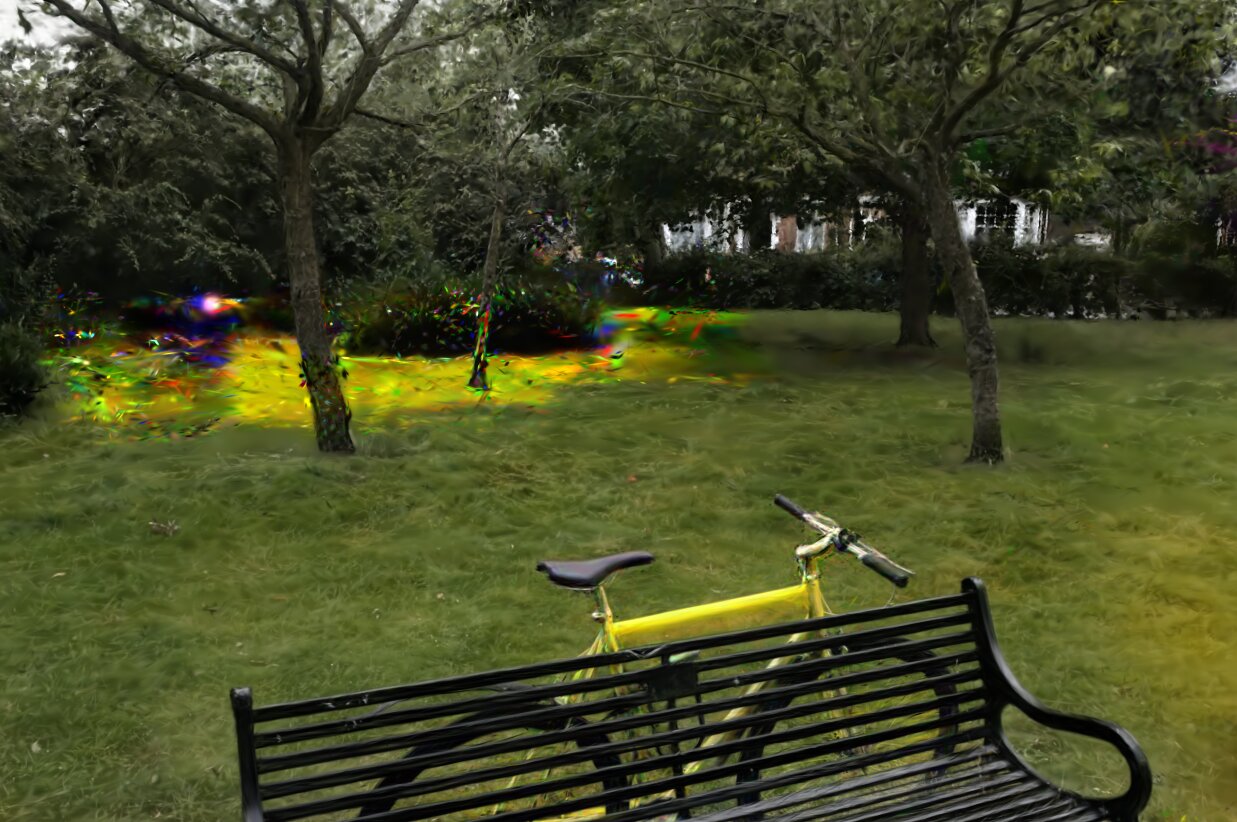} 
    \includegraphics[trim={0 0mm 0 10mm}, clip, width=0.325\columnwidth]{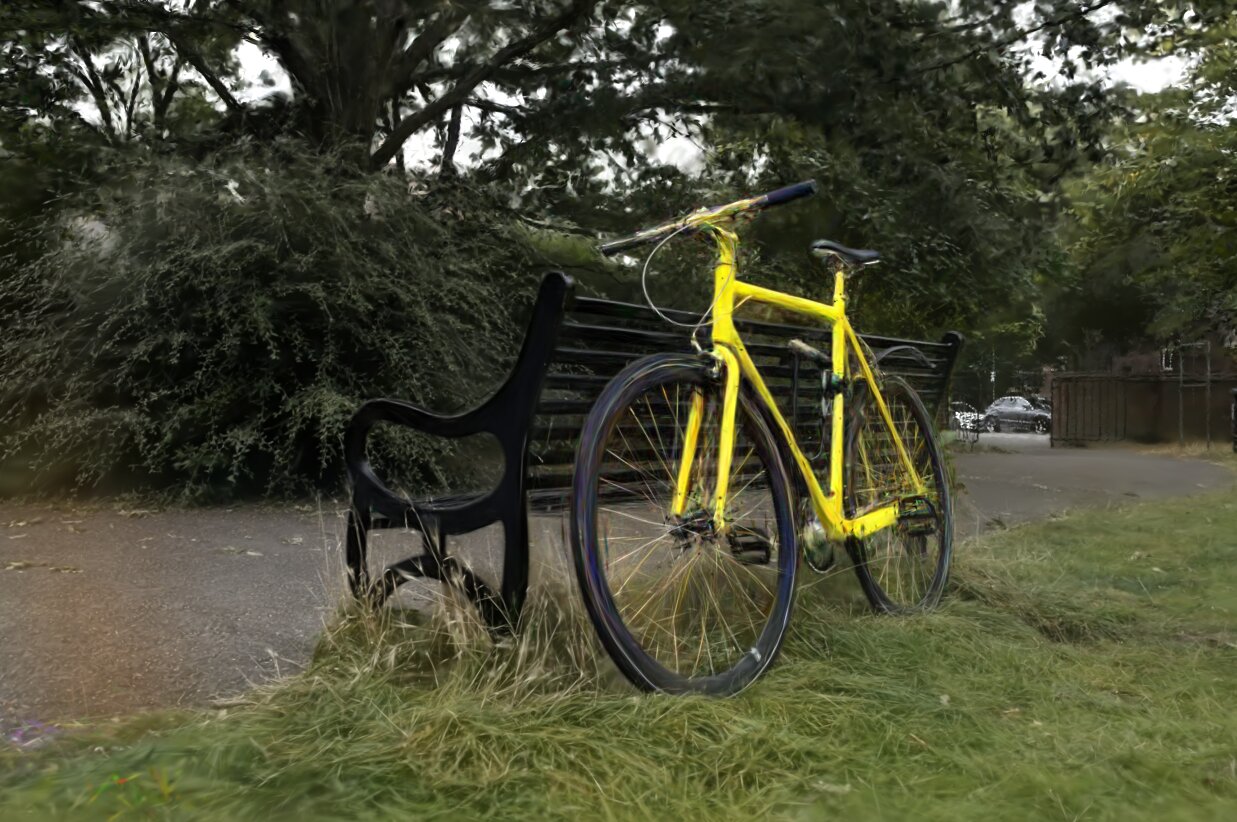}
    \includegraphics[trim={0 10mm 0 0mm}, clip, width=0.325\columnwidth]{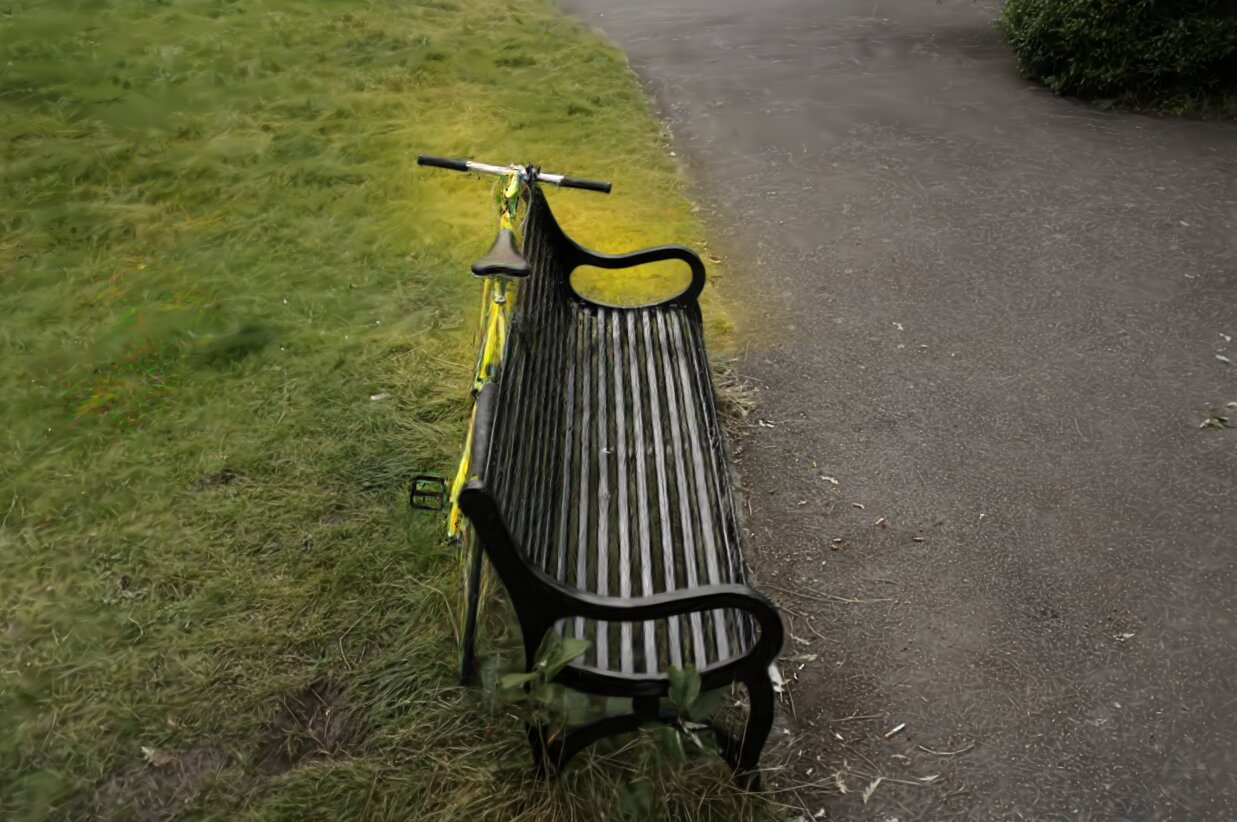}  \\
    
    \includegraphics[trim={0 10mm 0 0mm}, clip, width=0.325\columnwidth]{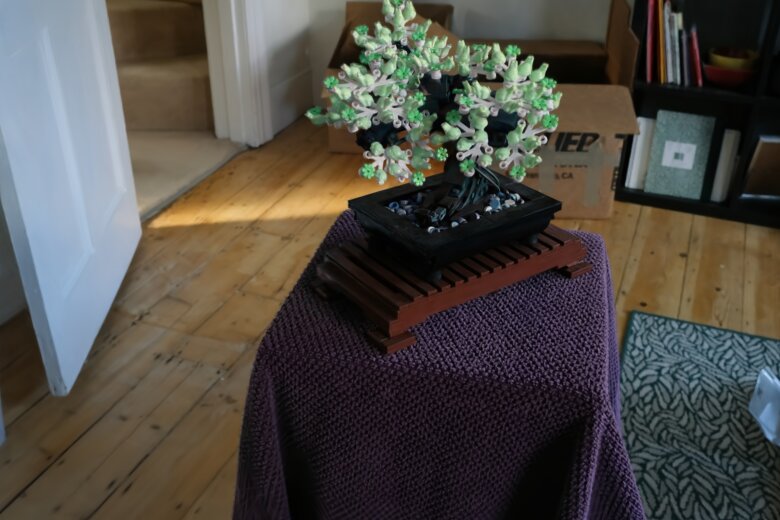} 
    \includegraphics[trim={0 10mm 0 0mm}, clip, width=0.325\columnwidth]{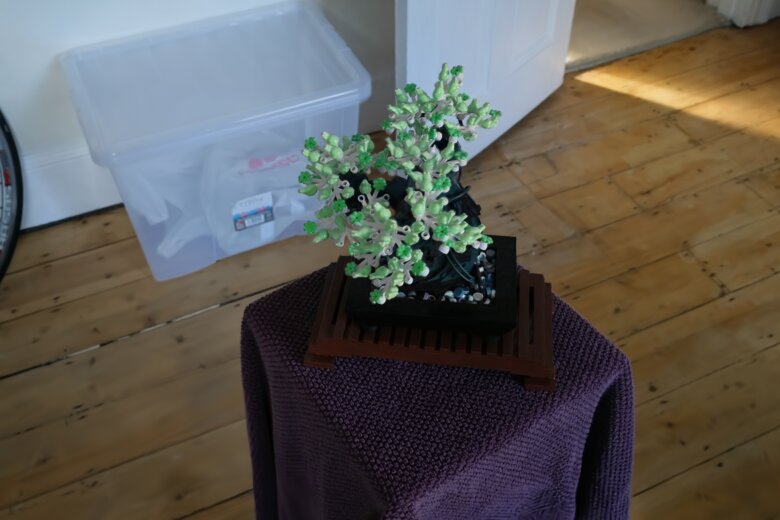}
    \includegraphics[trim={0 10mm 0 0mm}, clip, width=0.325\columnwidth]{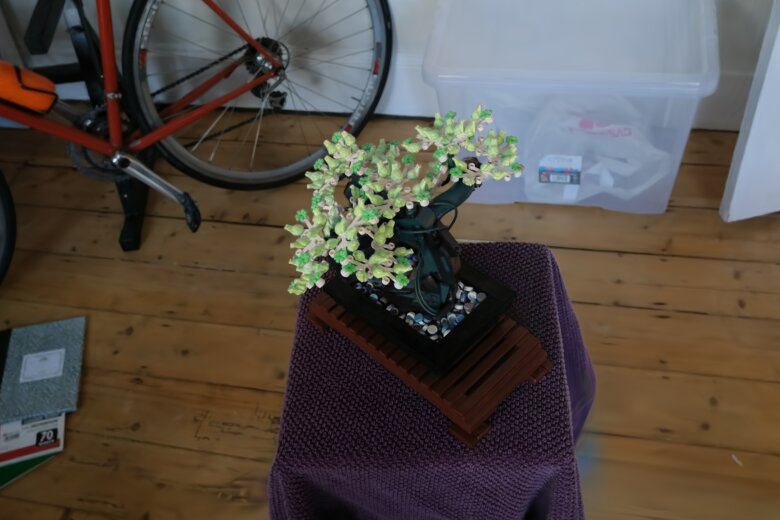}  \\
    \includegraphics[trim={0 10mm 0 0mm}, clip, width=0.325\columnwidth]{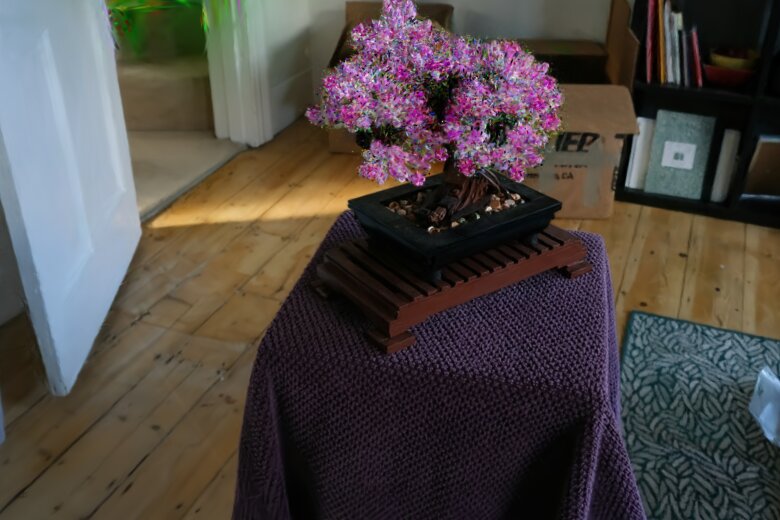} 
    \includegraphics[trim={0 10mm 0 0mm}, clip, width=0.325\columnwidth]{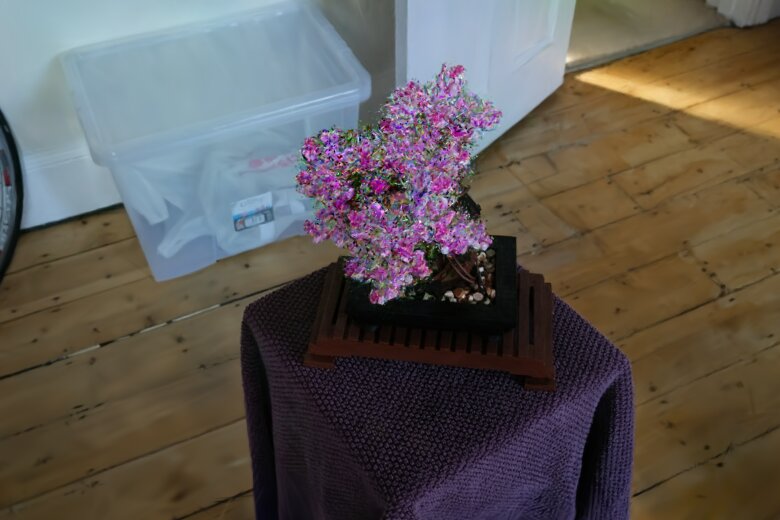}
    \includegraphics[trim={0 10mm 0 0mm}, clip, width=0.325\columnwidth]{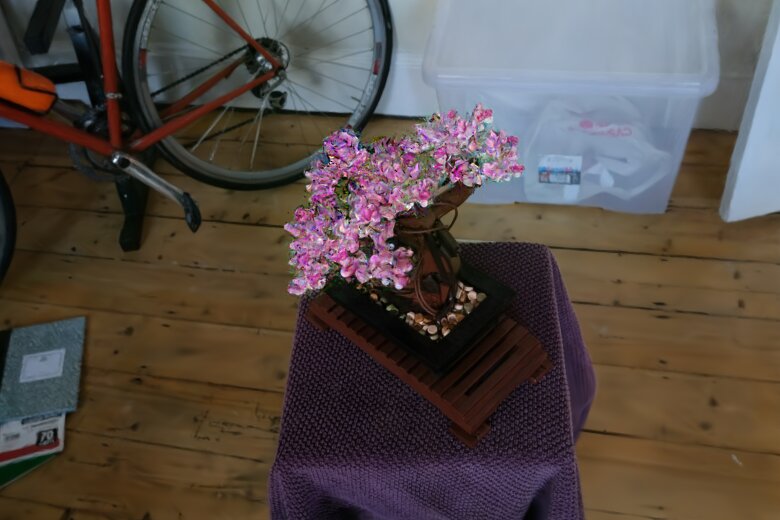}  \\

\vspace{-1mm}
    \caption{{\bf Comparison between  \methodname~(upper row per case) and Gaussian Editor~\cite{chen2023gaussianeditor} (bottom row per case).} Gaussian Editor requires the user to initialize a mask using SAM~\cite{Kirillov2023SegmentA} before applying color edits based on provided prompts: ``kitchen - make it magenta'', ``bicycle - make it yellow'', and ``bonsai - make it green''. Gaussian Editor exhibits several drawbacks, such as color bleeding into surrounding areas in the first two scenes and not understanding the scene structure when recoloring the third scene. Notably, Gaussian Editor takes an average of 612 seconds to process these scenes, while {\methodname} completes the task in just 2 seconds on average.
    %{\bf Comparison with GaussianEditor~\cite{chen2023gaussianeditor}.} Comparison between \methodname, in the upper row for each case, and Gaussian Editor, on the lower row for each case. For GaussianEditor a mask has to be initialised by the user using SAM~\cite{Kirillov2023SegmentA}. Prompts were provided to edit the masked object: kitchen - "make it magenta", bicycle - "make it yellow", bonsai - "make it green". GaussianEditor suffers from several problems like recoloring bleeding out into the rest of the scene as seen in the first two scenes. In the third scene GaussianEditor does not understand the structure of the scene and how to properly perform the recoloring of it. The average computation time of GaussianEditor on those scenes is of 612 seconds (+10 minutes) while {\methodname} takes an average of 2 seconds.
    }
    \label{fig:GEditor_comparison}
    \centering
    \vspace{-2mm} 
\end{figure}

\subsection{Qualitative results}

\begin{figure*}[t]

\begin{center}
\begin{tabularx}\textwidth{XXXXXX}
\centering \hspace{3mm} Original image & 
\centering \hspace{-1mm} Ground truth recolor & 
\centering \hspace{-1mm} {\methodname} segmentation & 
\centering \hspace{-2mm} {\methodname} edition &
\centering \hspace{-2mm} {\methodname} segmentation & 
\centering \hspace{-3mm} {\methodname} edition
\end{tabularx}
\end{center}

\vspace{-1mm}

    \centering
    \includegraphics[width=0.16\textwidth]{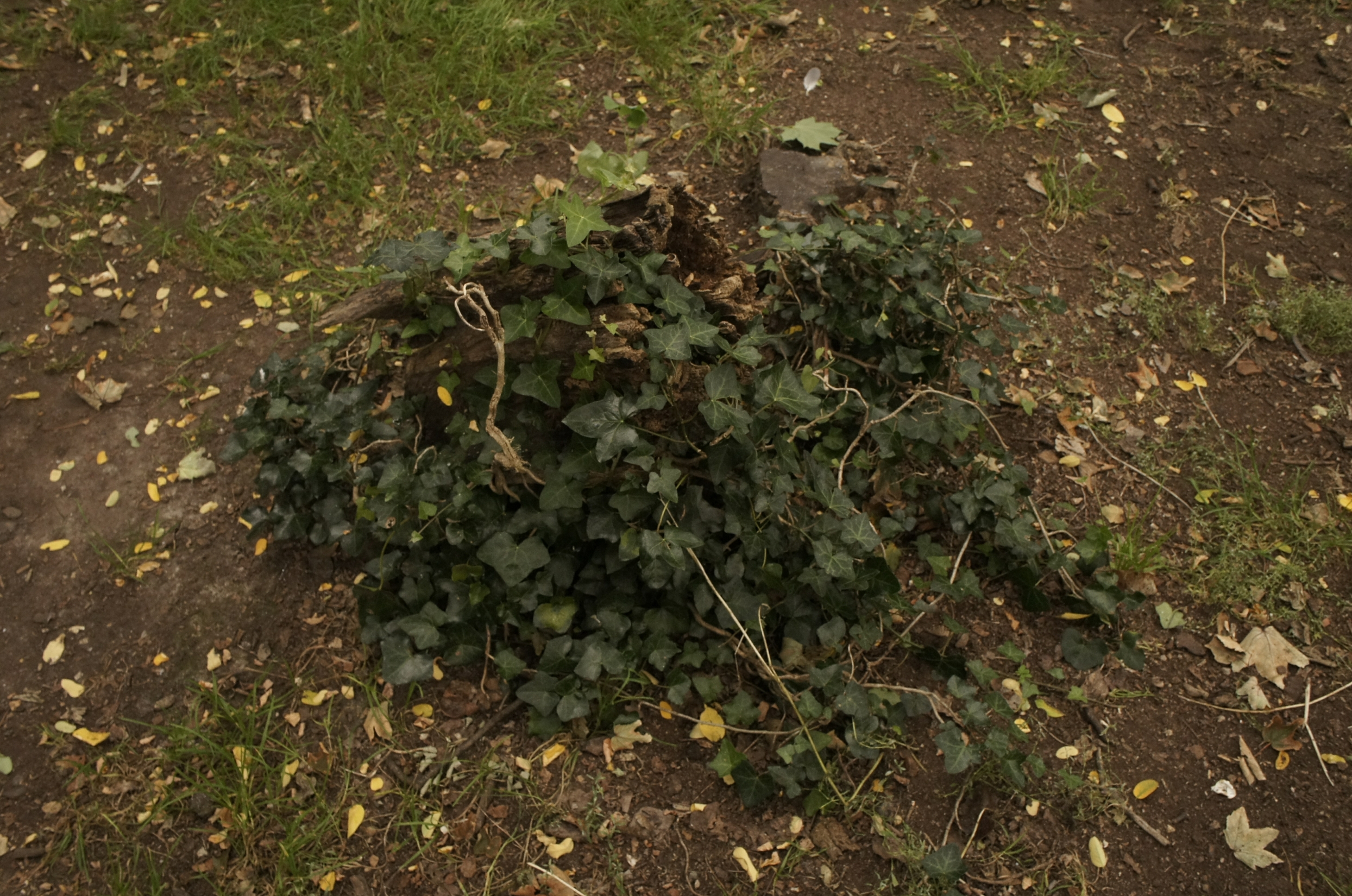}    \includegraphics[width=0.16\textwidth]{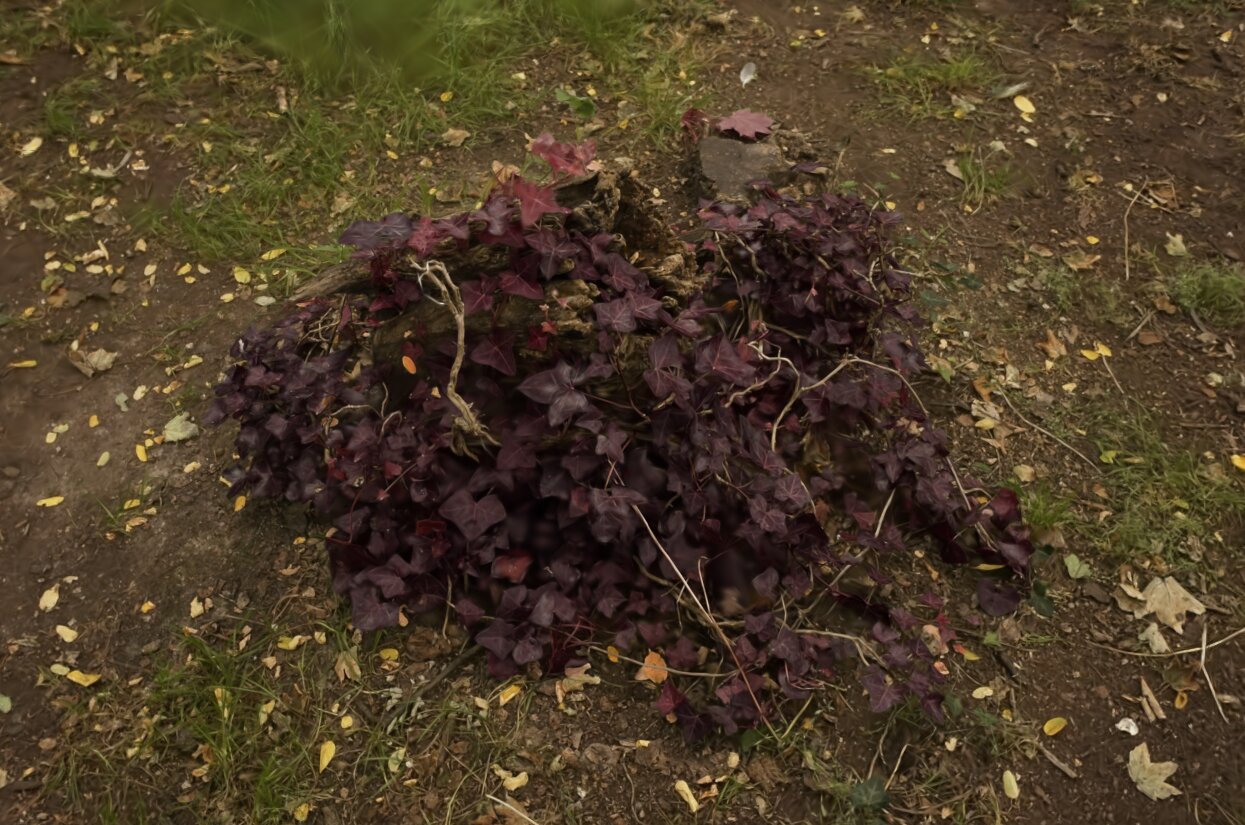} 
    \includegraphics[width=0.16\textwidth]{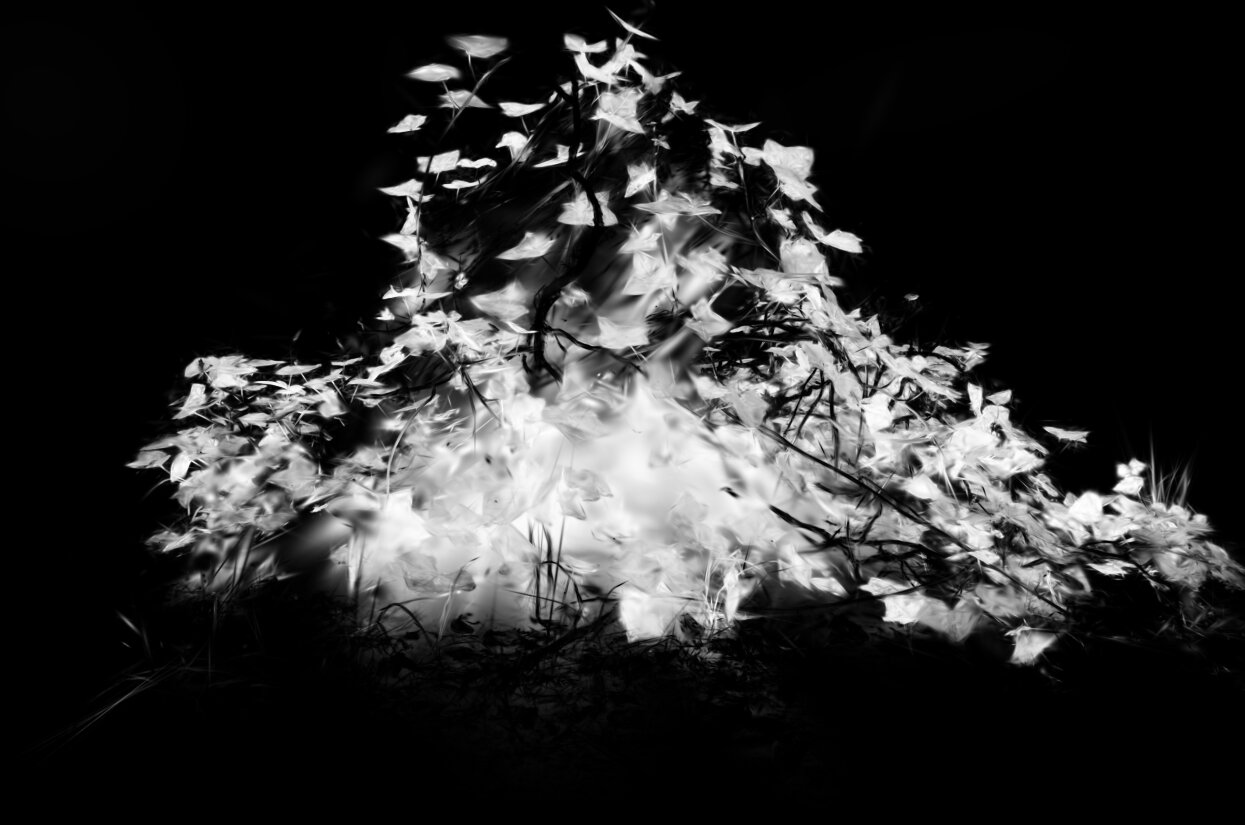} 
    \includegraphics[width=0.16\textwidth]{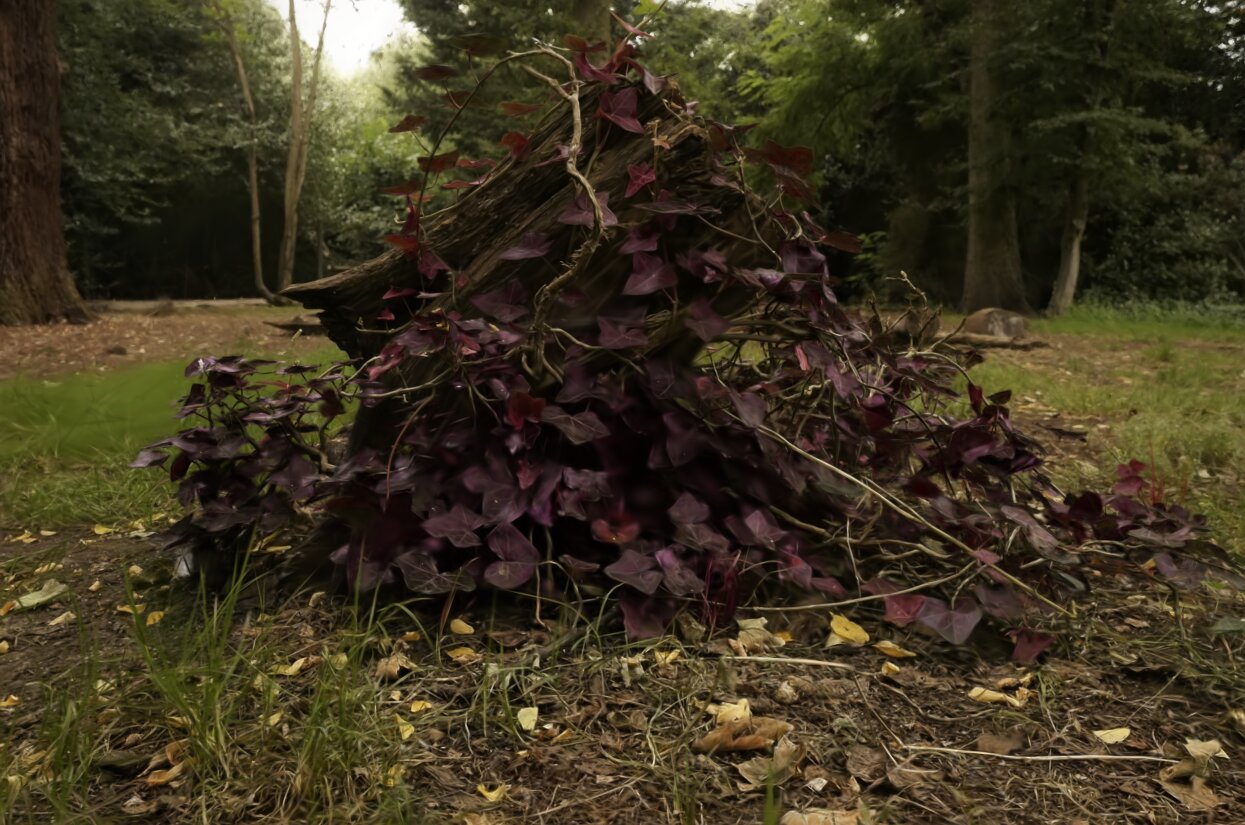} 
    \includegraphics[width=0.16\textwidth]{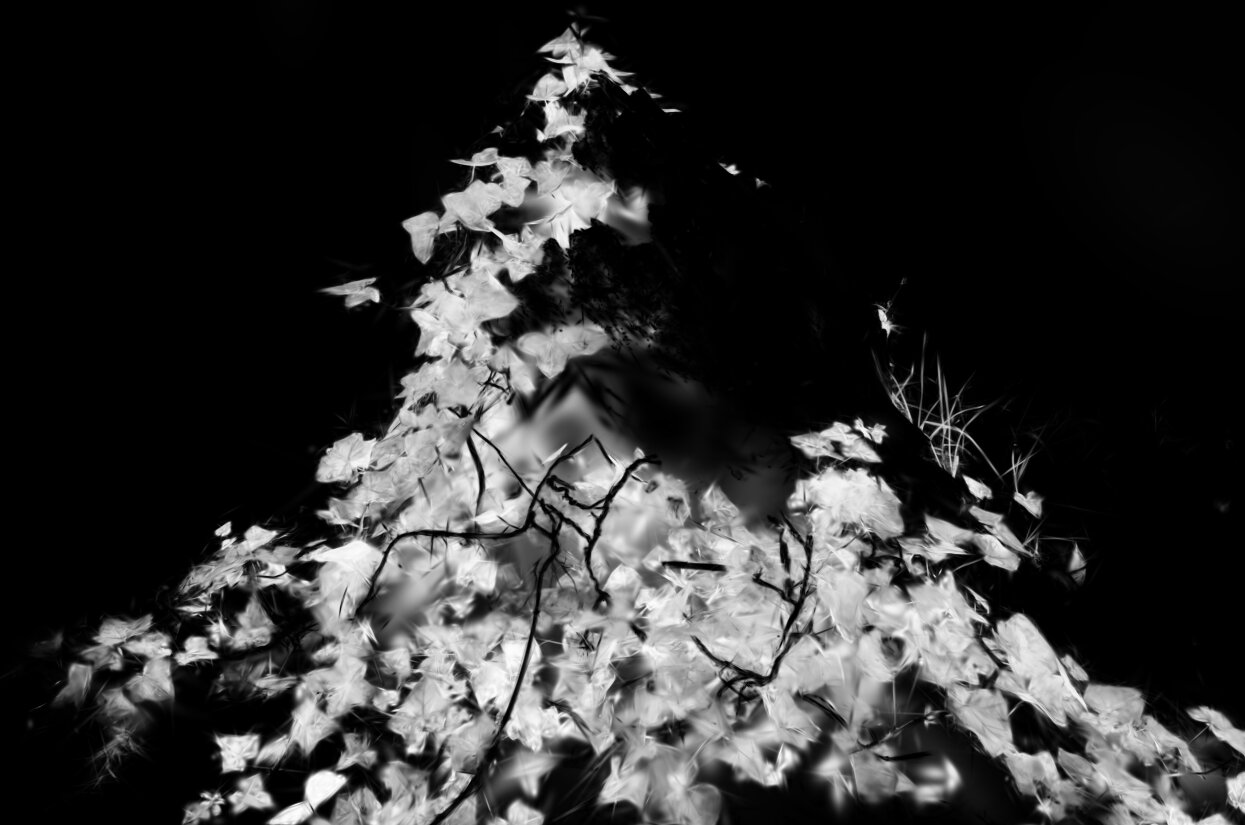}
    \includegraphics[width=0.16\textwidth]{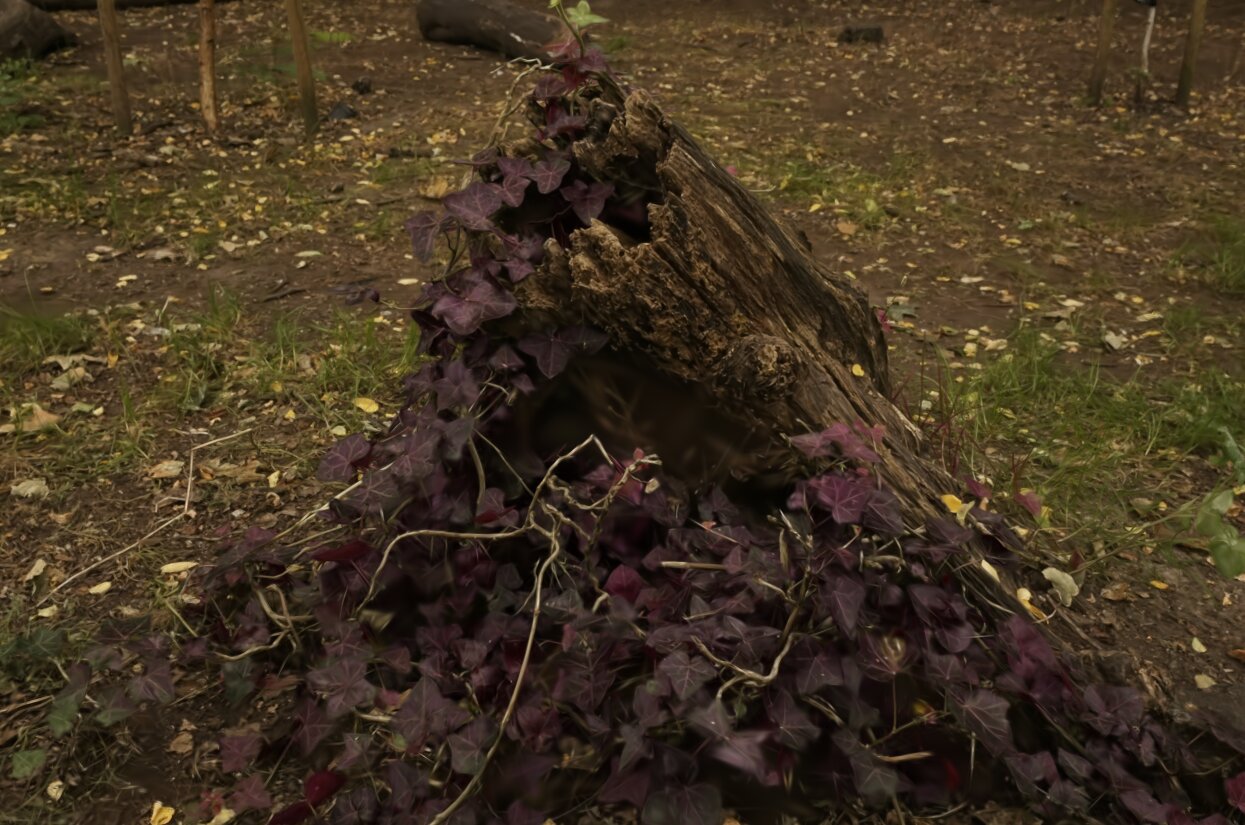} \\

    \includegraphics[width=0.16\textwidth]{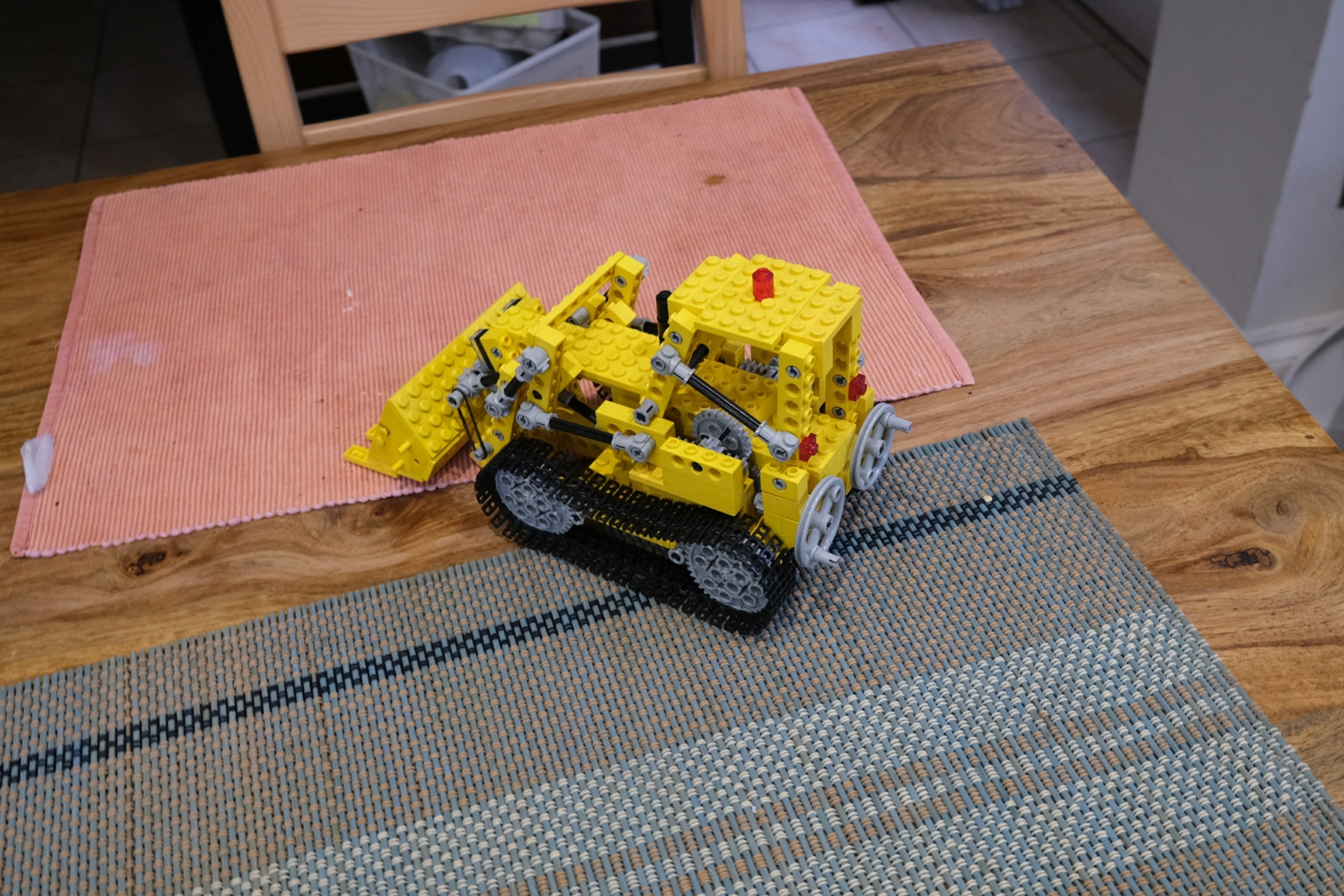}    \includegraphics[width=0.16\textwidth]{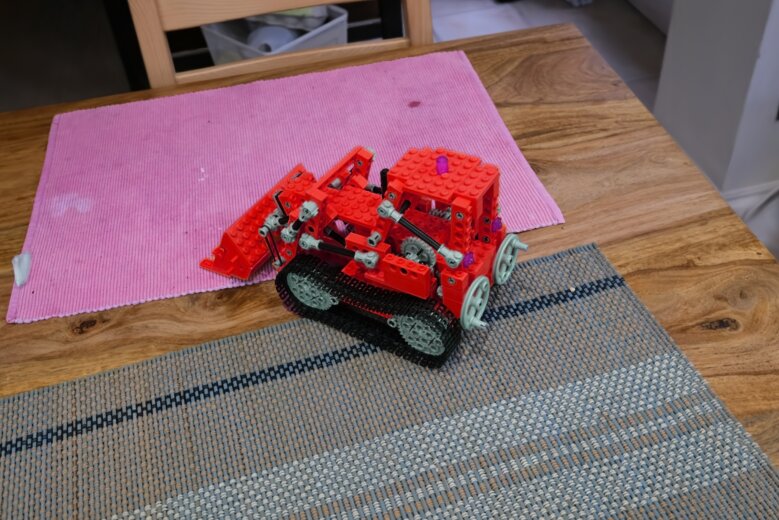} 
    \includegraphics[width=0.16\textwidth]{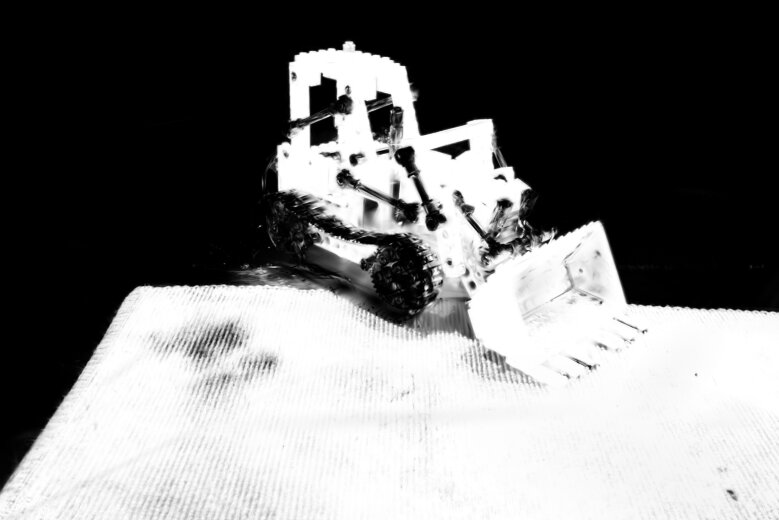} 
    \includegraphics[width=0.16\textwidth]{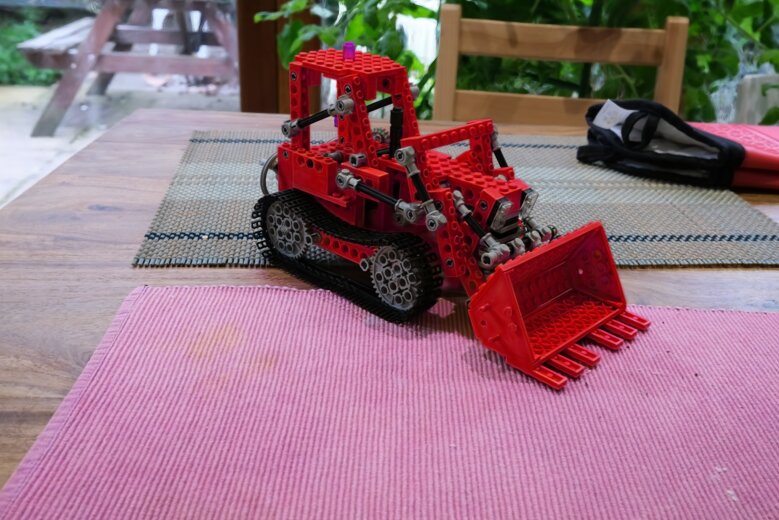} 
    \includegraphics[width=0.16\textwidth]{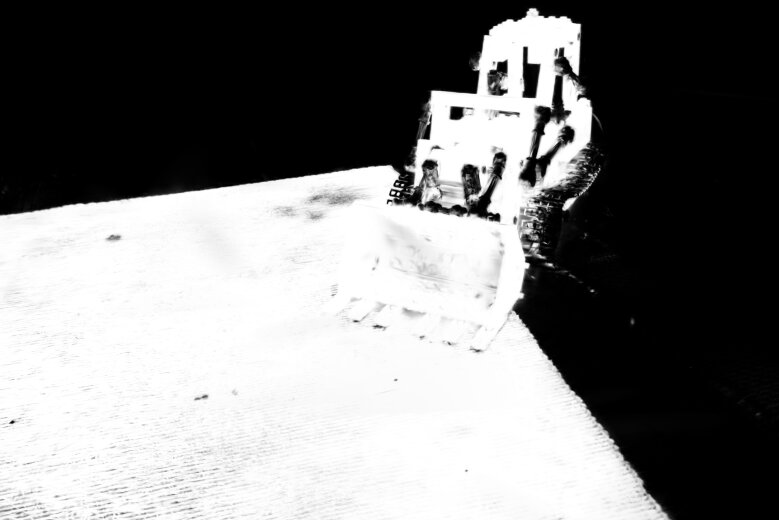} 
    \includegraphics[width=0.16\textwidth]{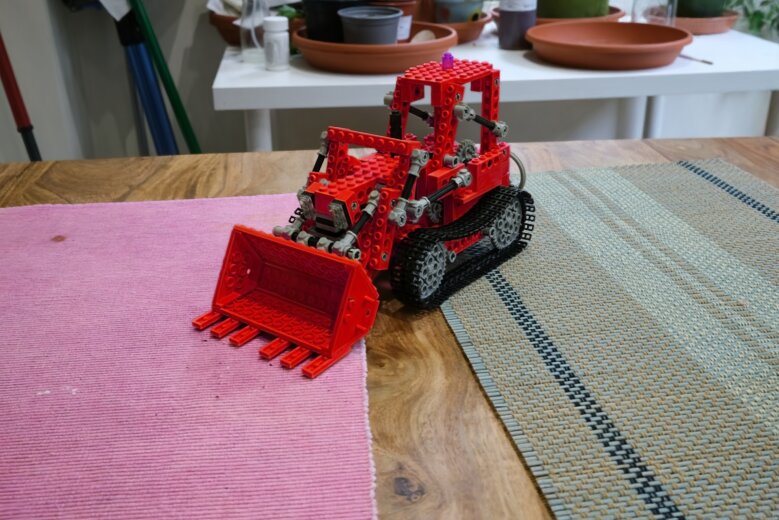} \\

    \includegraphics[width=0.16\textwidth]{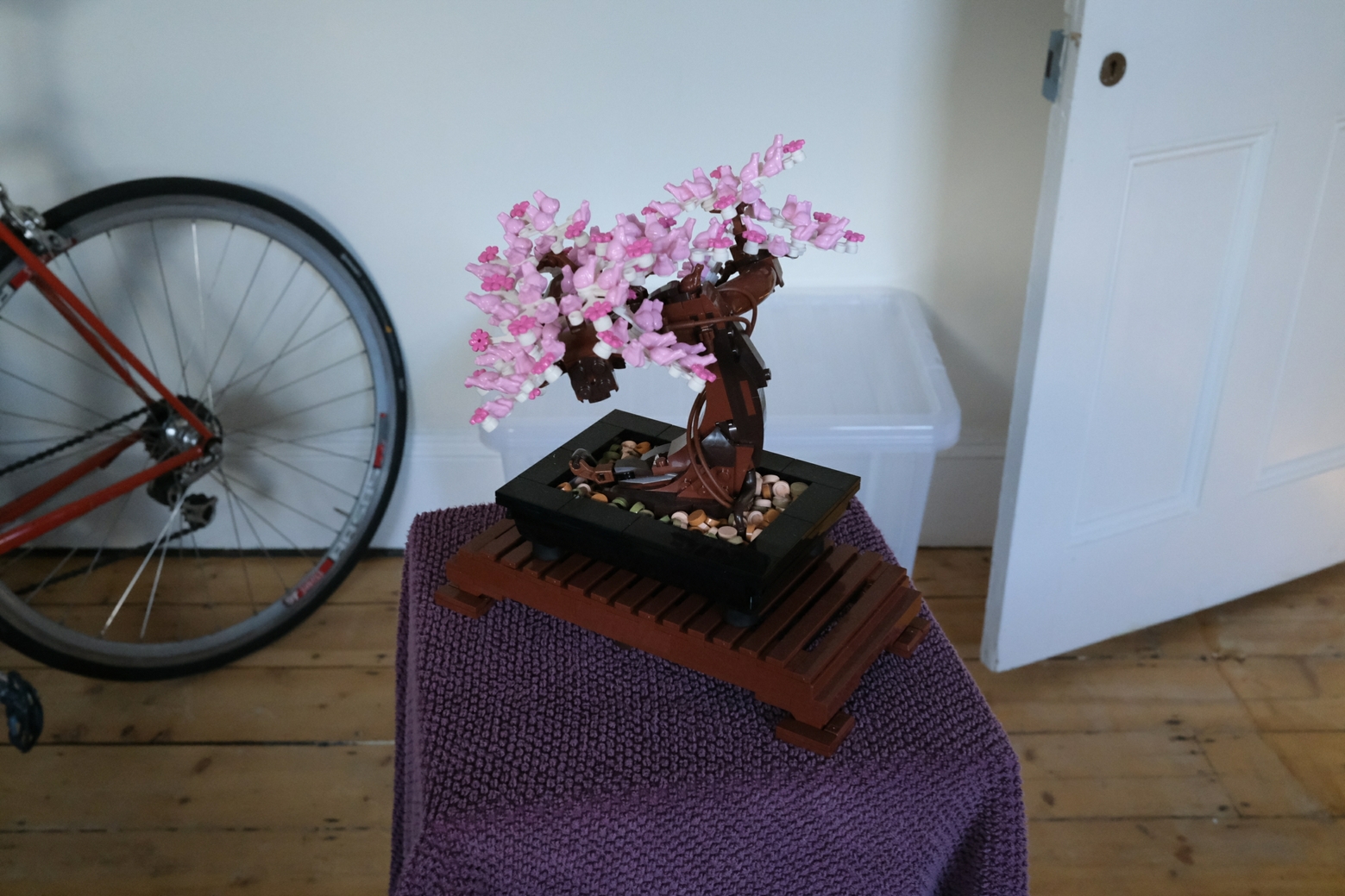}    \includegraphics[width=0.16\textwidth]{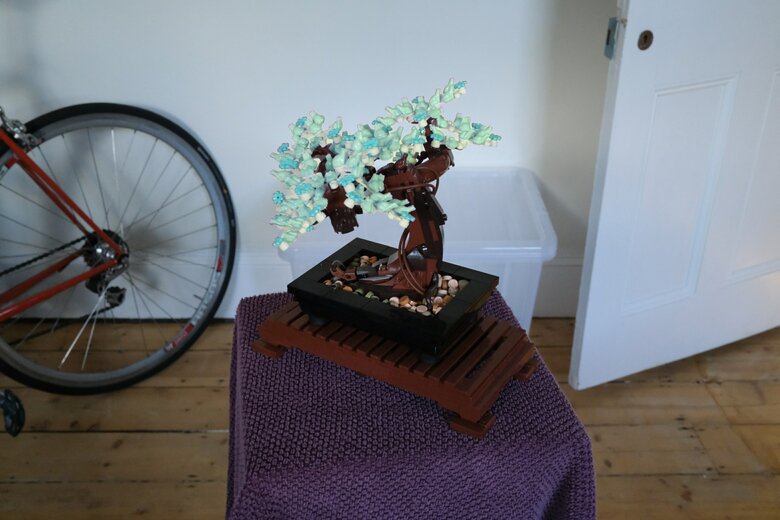} 
    \includegraphics[width=0.16\textwidth]{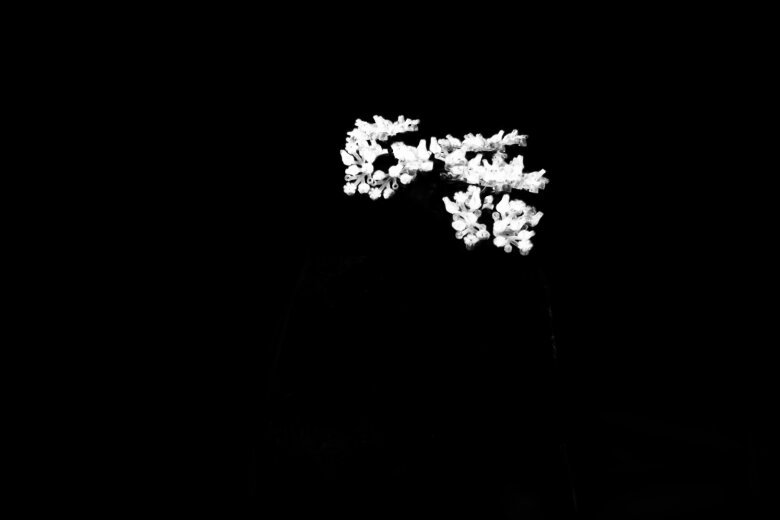} 
    \includegraphics[width=0.16\textwidth]{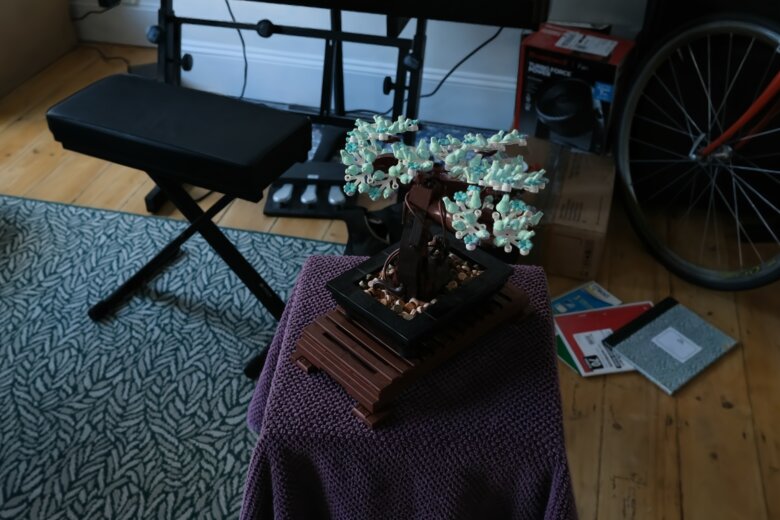}  
    \includegraphics[width=0.16\textwidth]{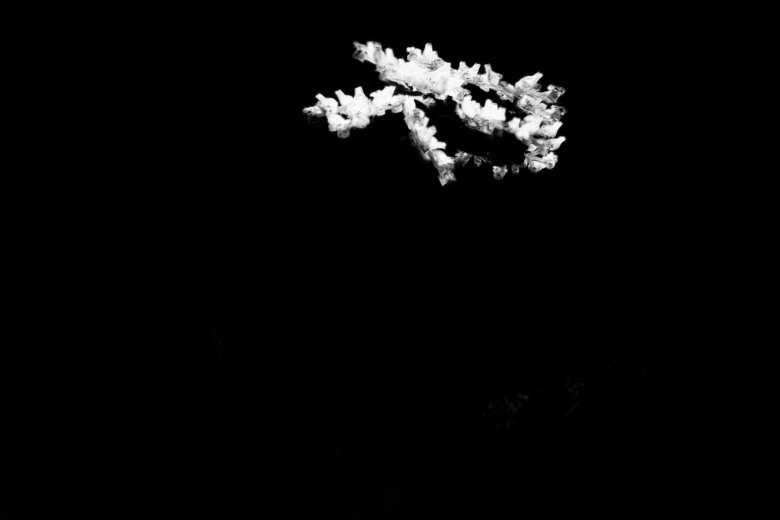}
    \includegraphics[width=0.16\textwidth]{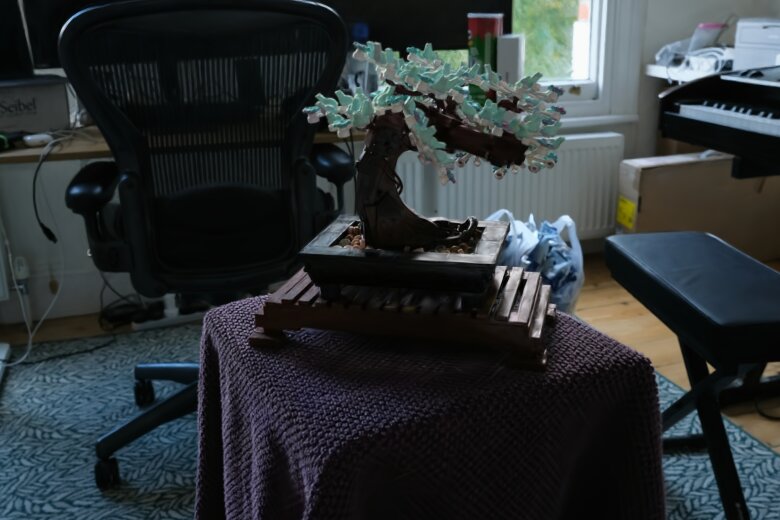} \\

    \includegraphics[width=0.16\textwidth]{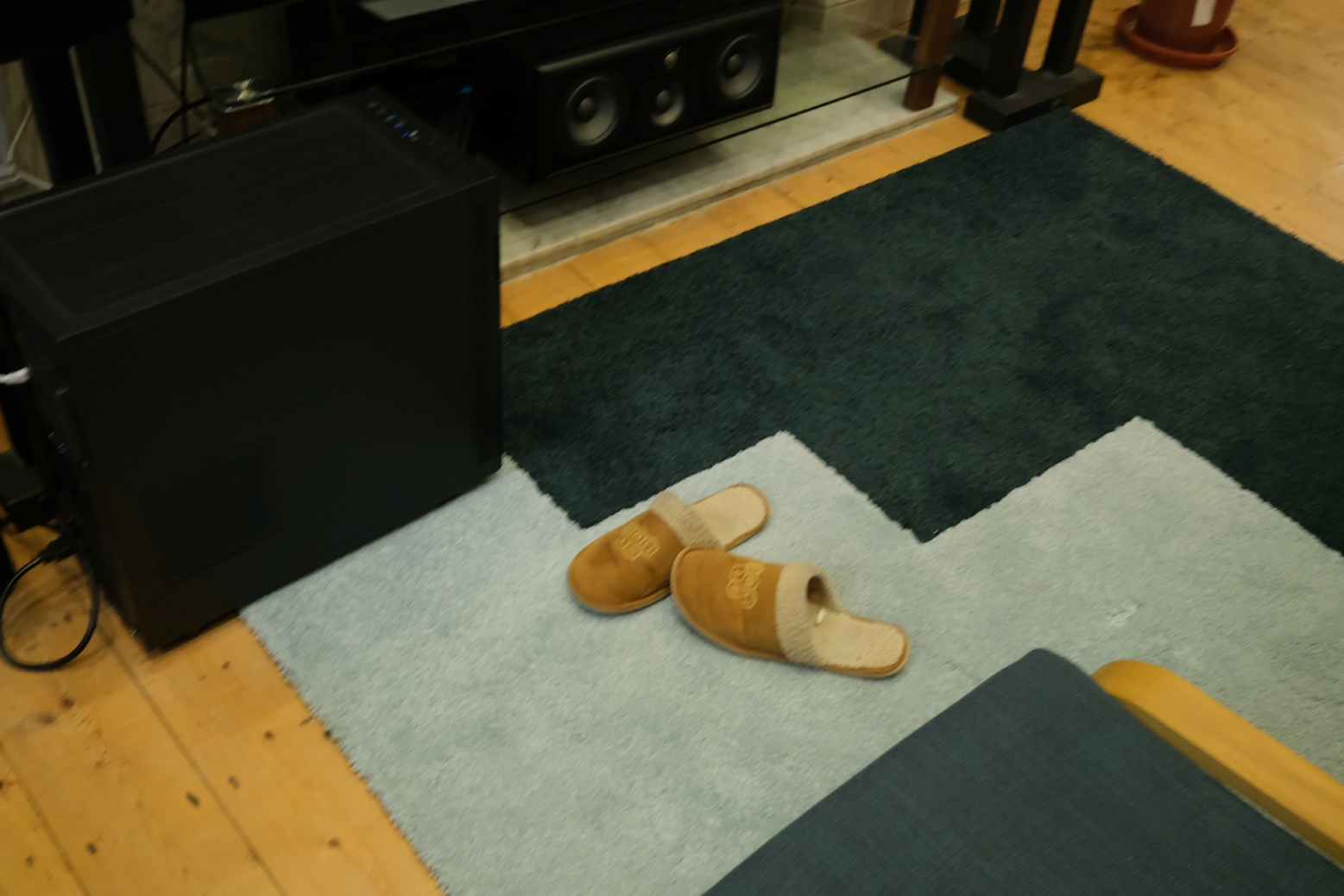}    \includegraphics[width=0.16\textwidth]{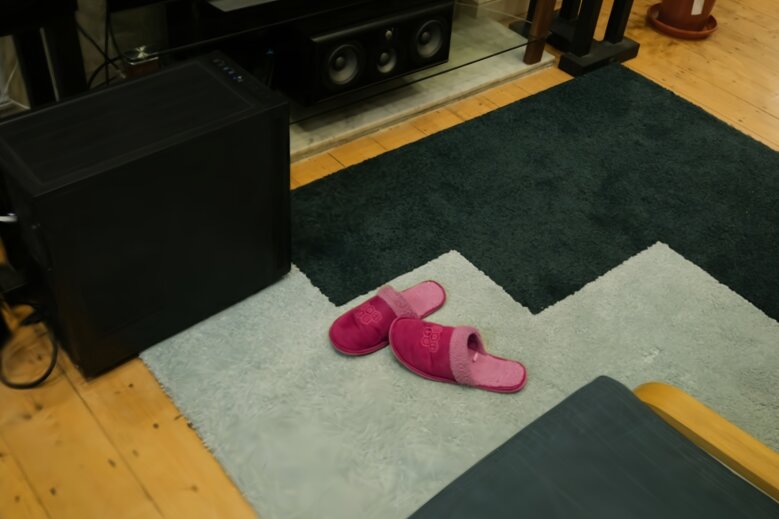} 
    \includegraphics[width=0.16\textwidth]{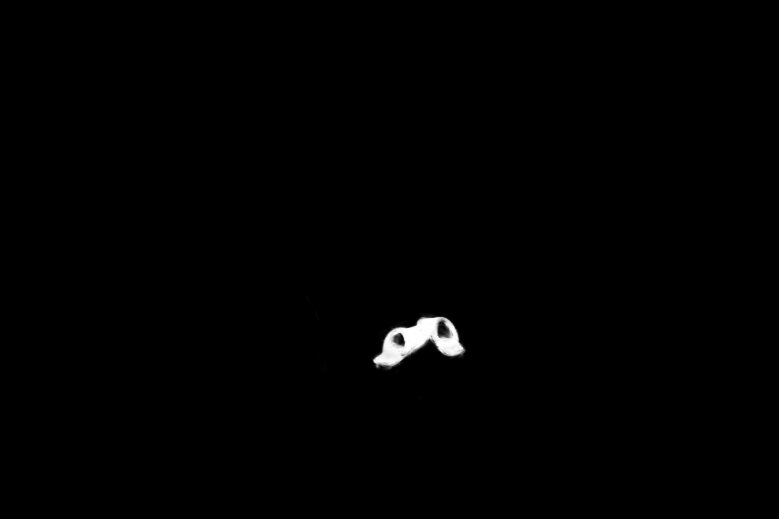} 
    \includegraphics[width=0.16\textwidth]{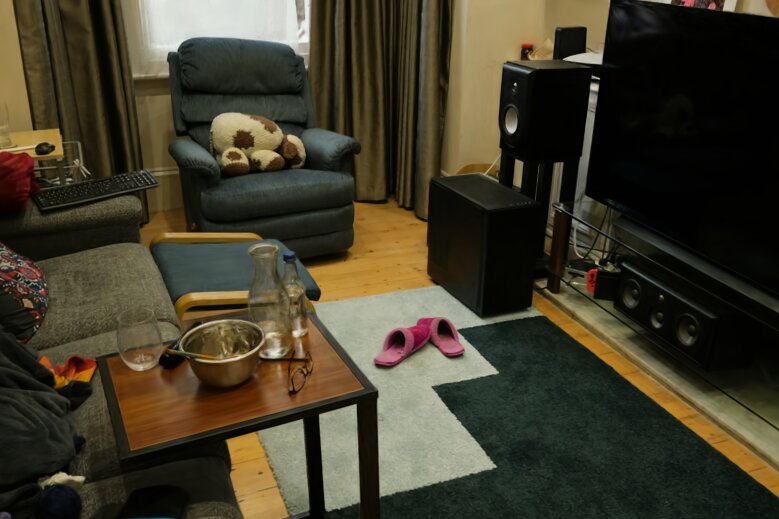} 
    \includegraphics[width=0.16\textwidth]{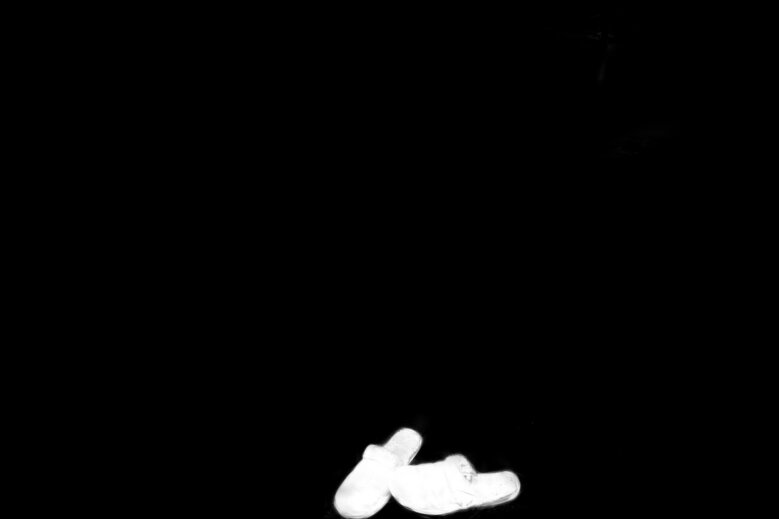}
    \includegraphics[width=0.16\textwidth]{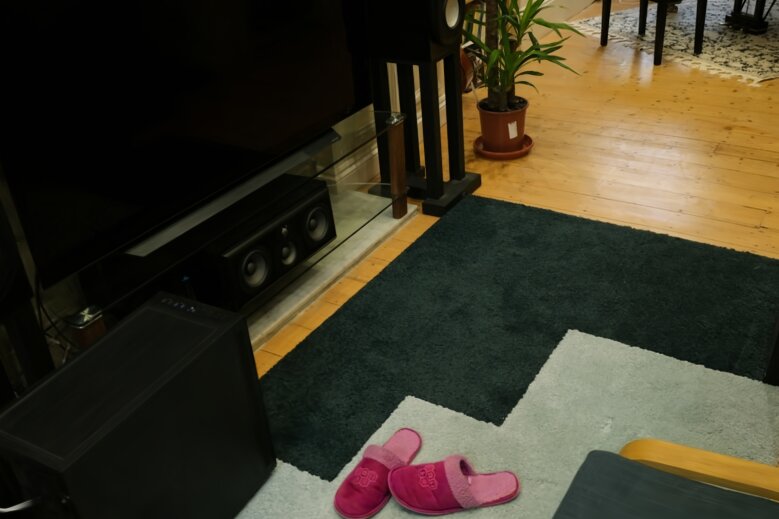} \\

    \includegraphics[width=0.16\textwidth]{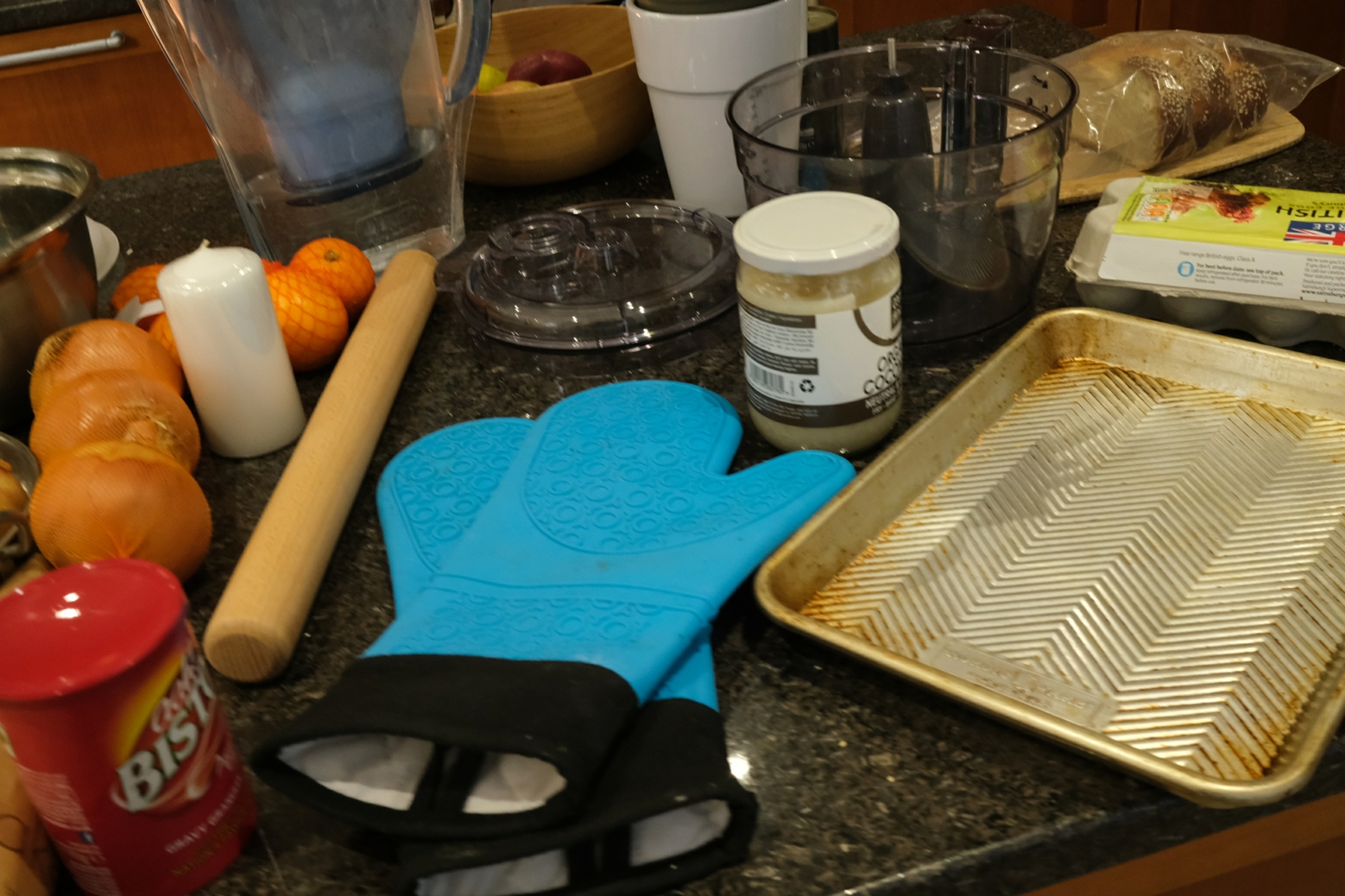}    \includegraphics[width=0.16\textwidth]{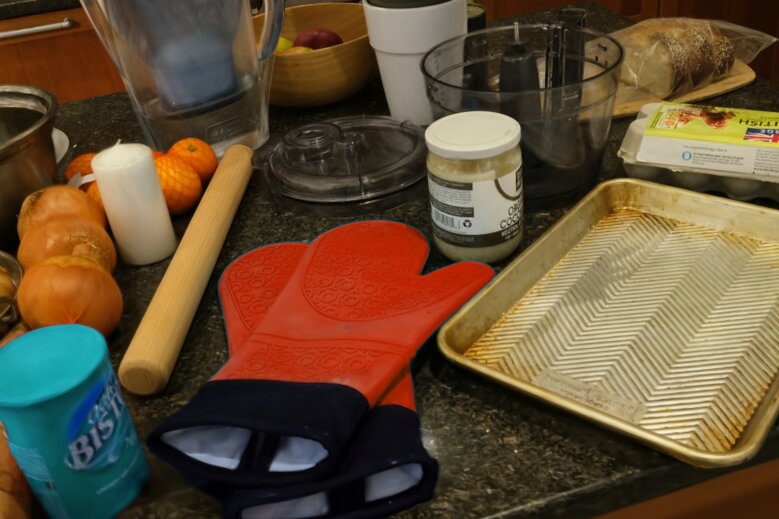} 
    \includegraphics[width=0.16\textwidth]{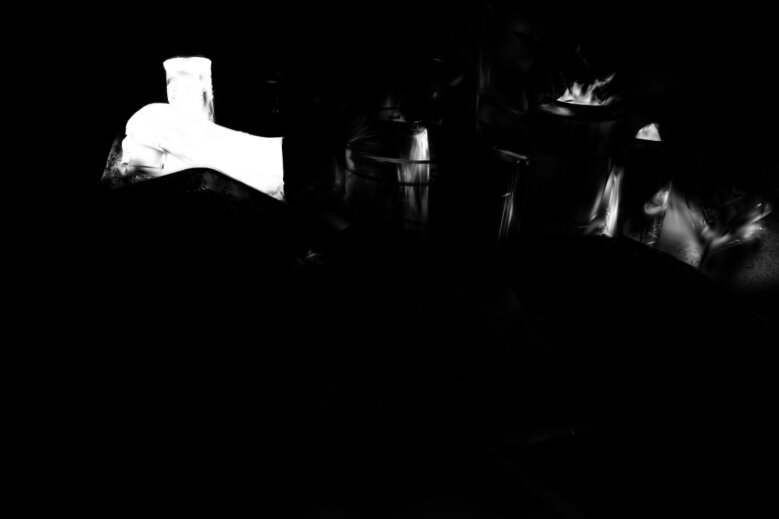} 
    \includegraphics[width=0.16\textwidth]{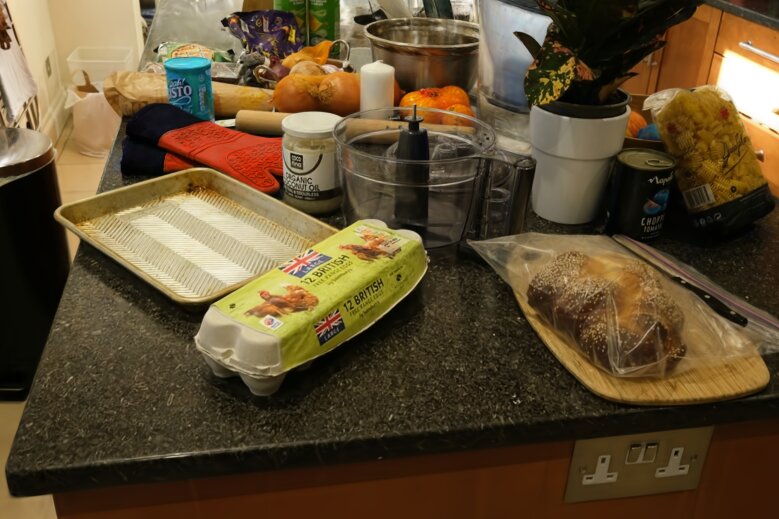}  
    \includegraphics[width=0.16\textwidth]{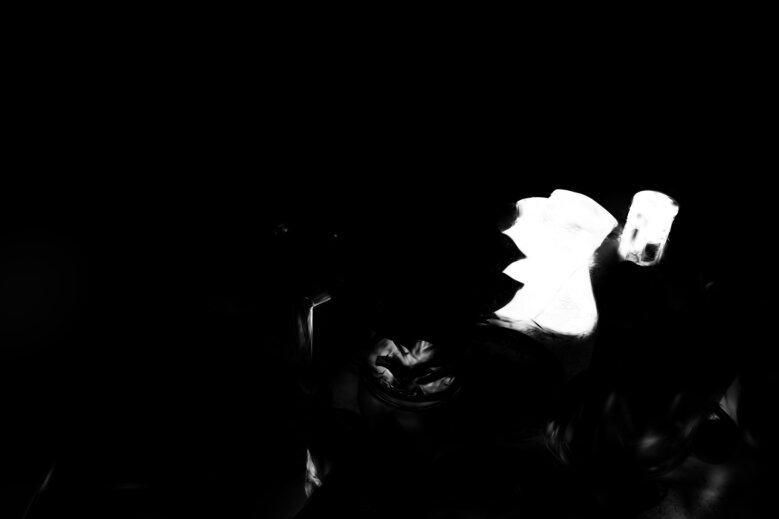} 
    \includegraphics[width=0.16\textwidth]{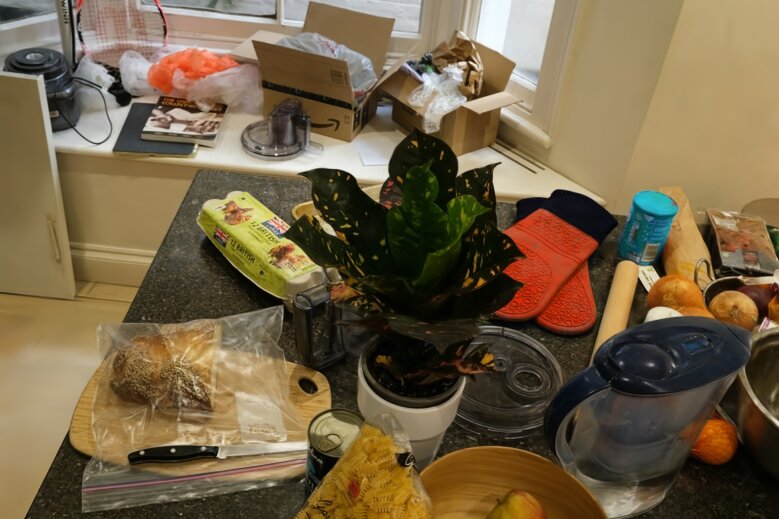} \\

    \includegraphics[width=0.16\textwidth]{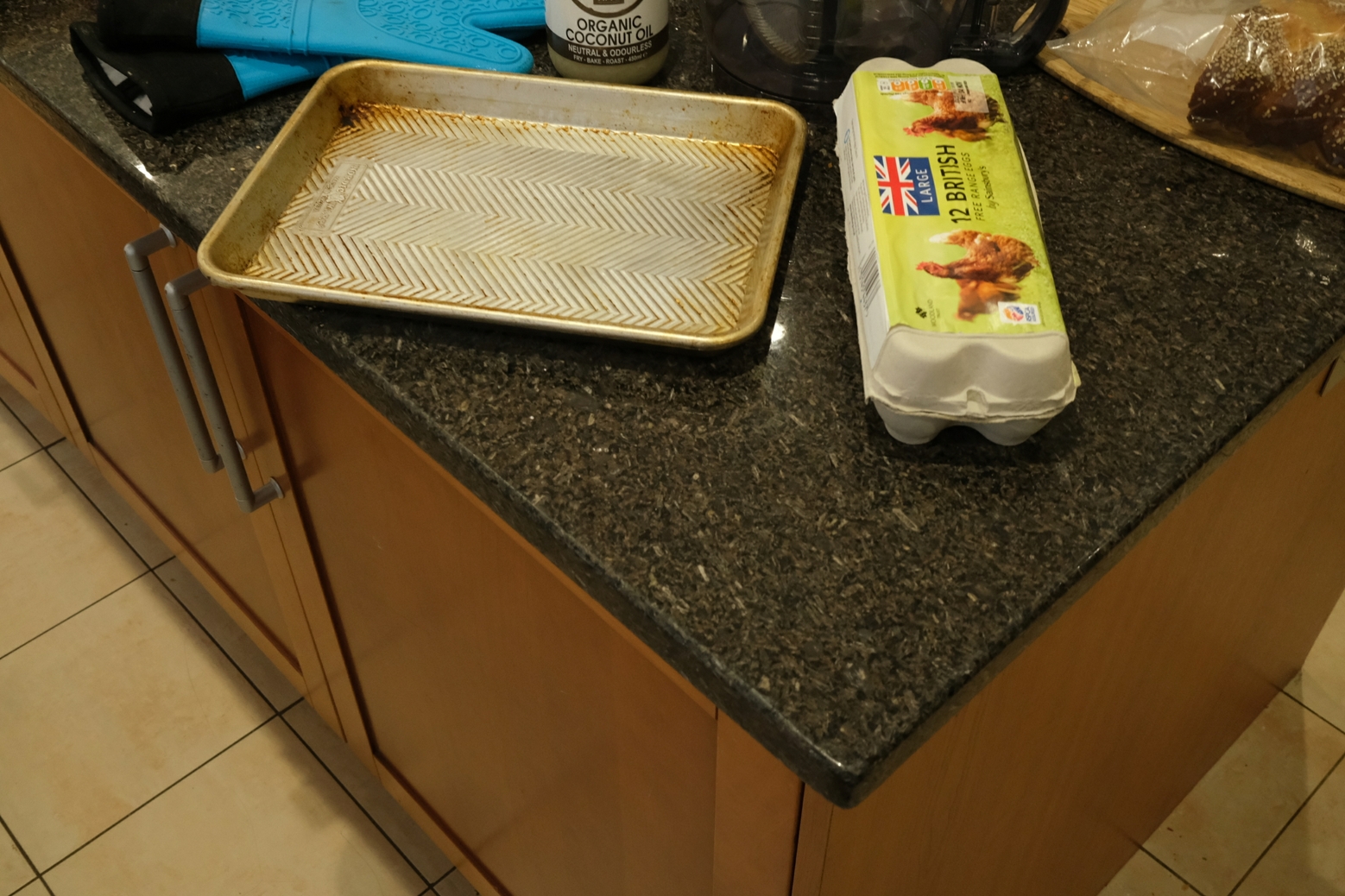}    \includegraphics[width=0.16\textwidth]{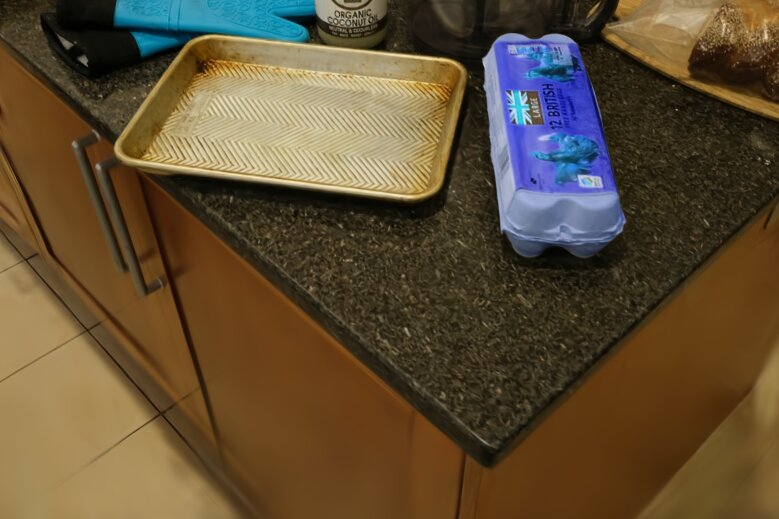} 
    \includegraphics[width=0.16\textwidth]{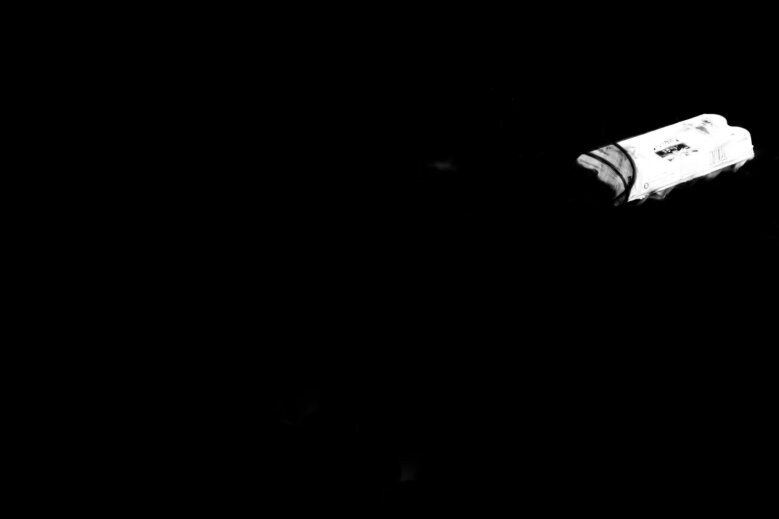} 
    \includegraphics[width=0.16\textwidth]{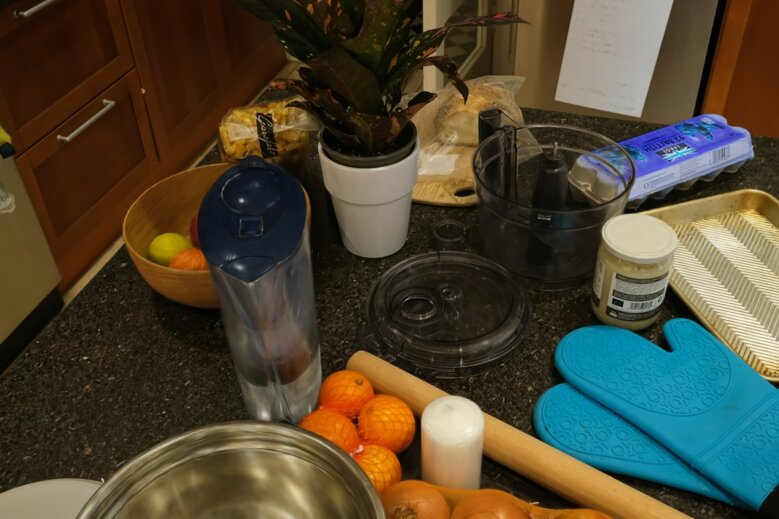}  
    \includegraphics[width=0.16\textwidth]{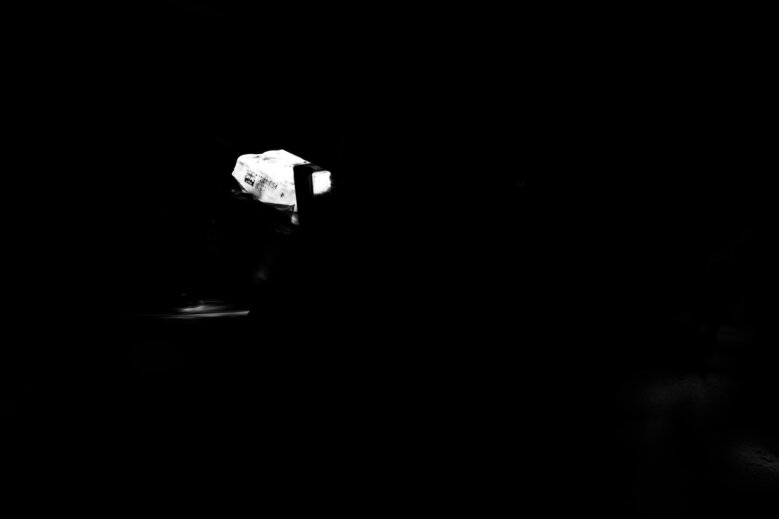} 
    \includegraphics[width=0.16\textwidth]{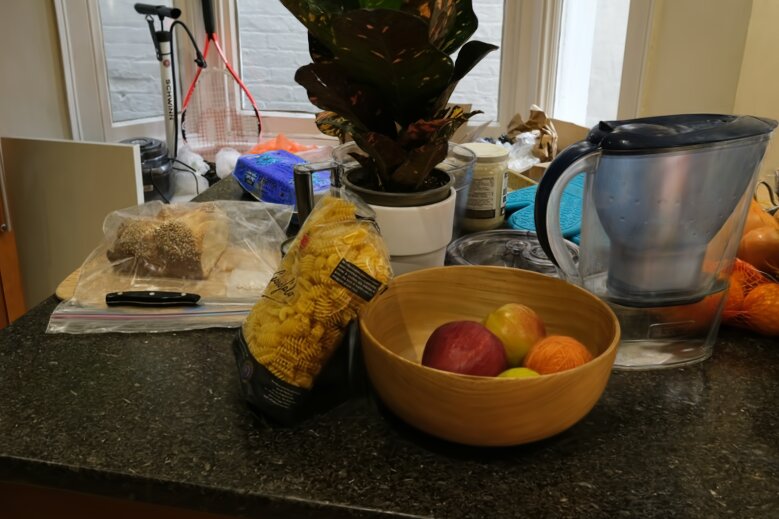} \\

    \includegraphics[width=0.16\textwidth]{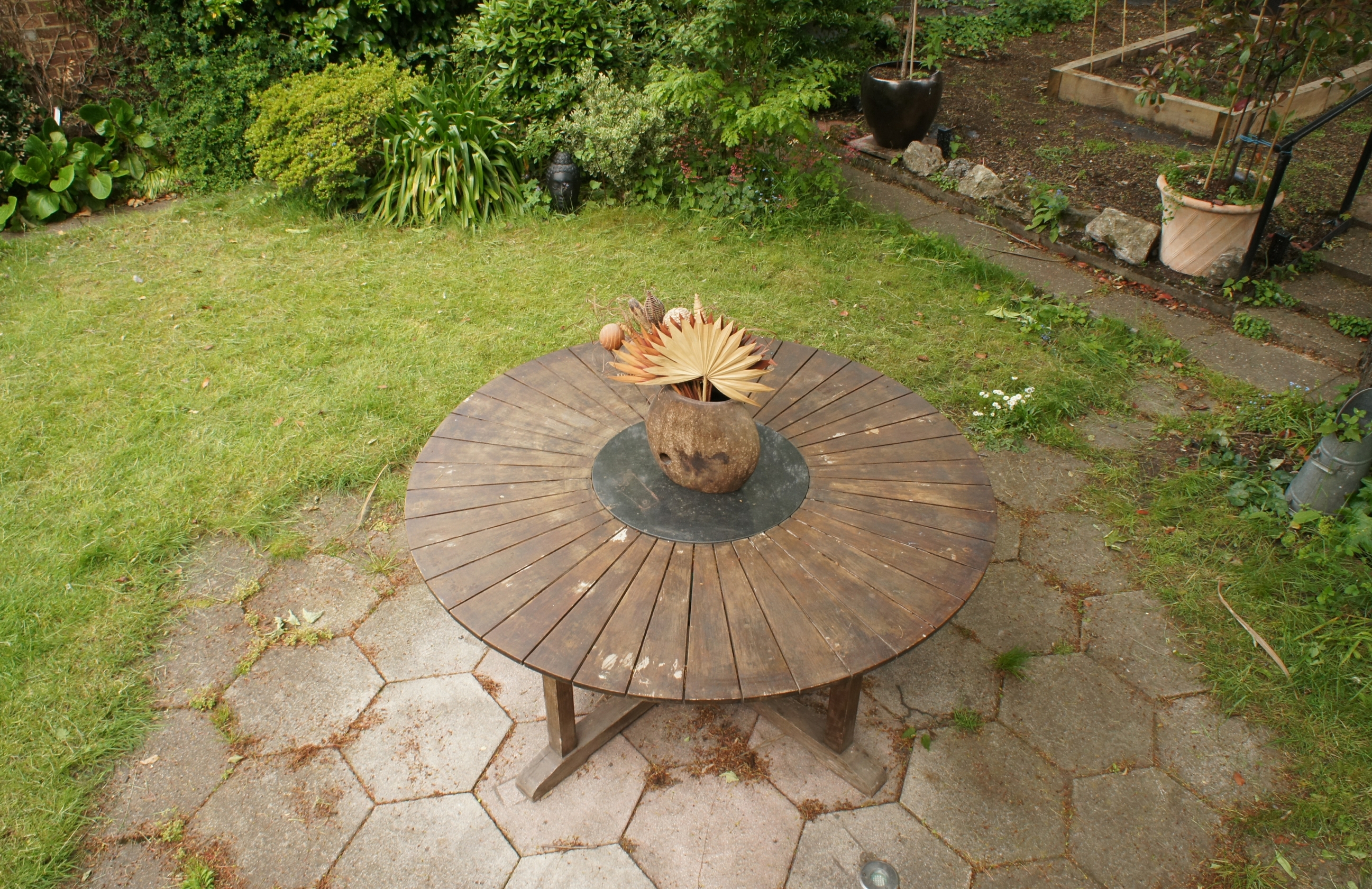}    \includegraphics[width=0.16\textwidth]{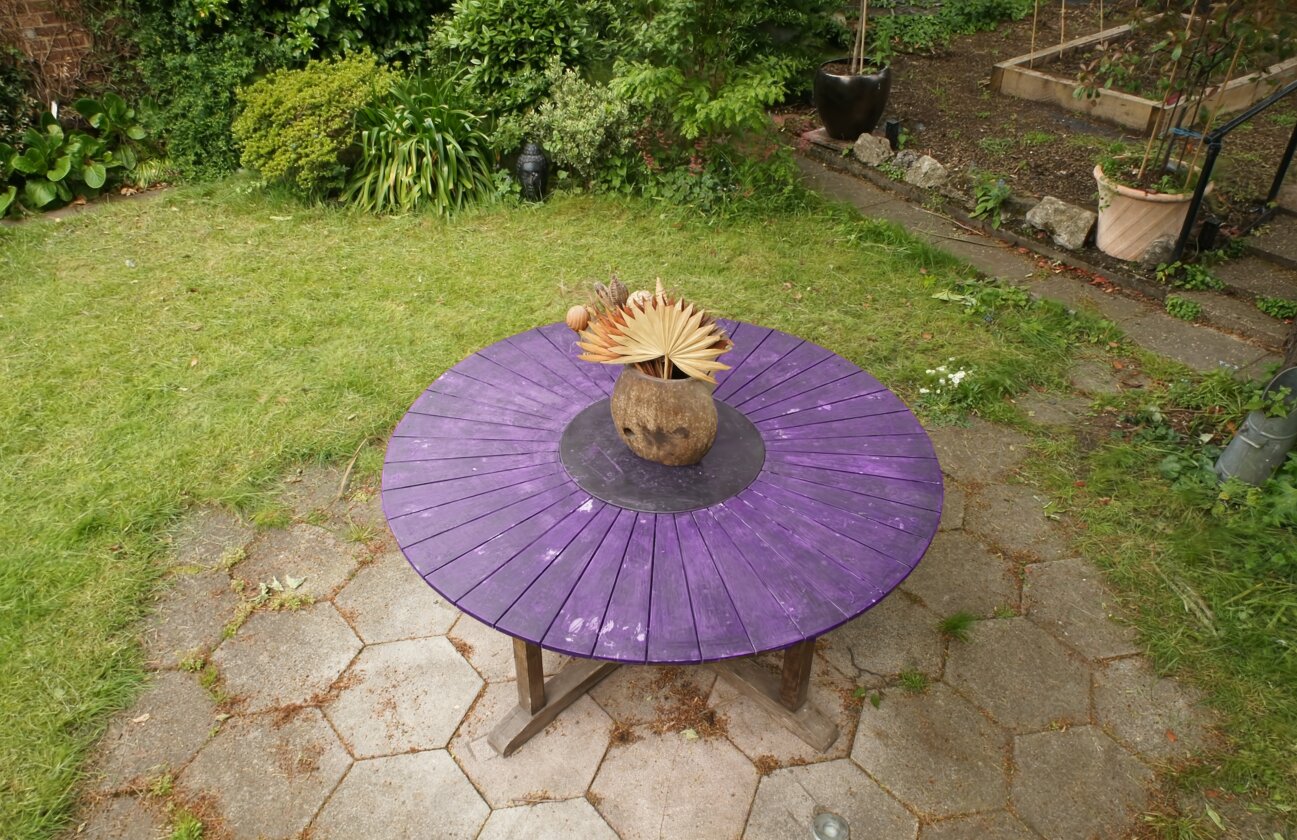} 
    \includegraphics[width=0.16\textwidth]{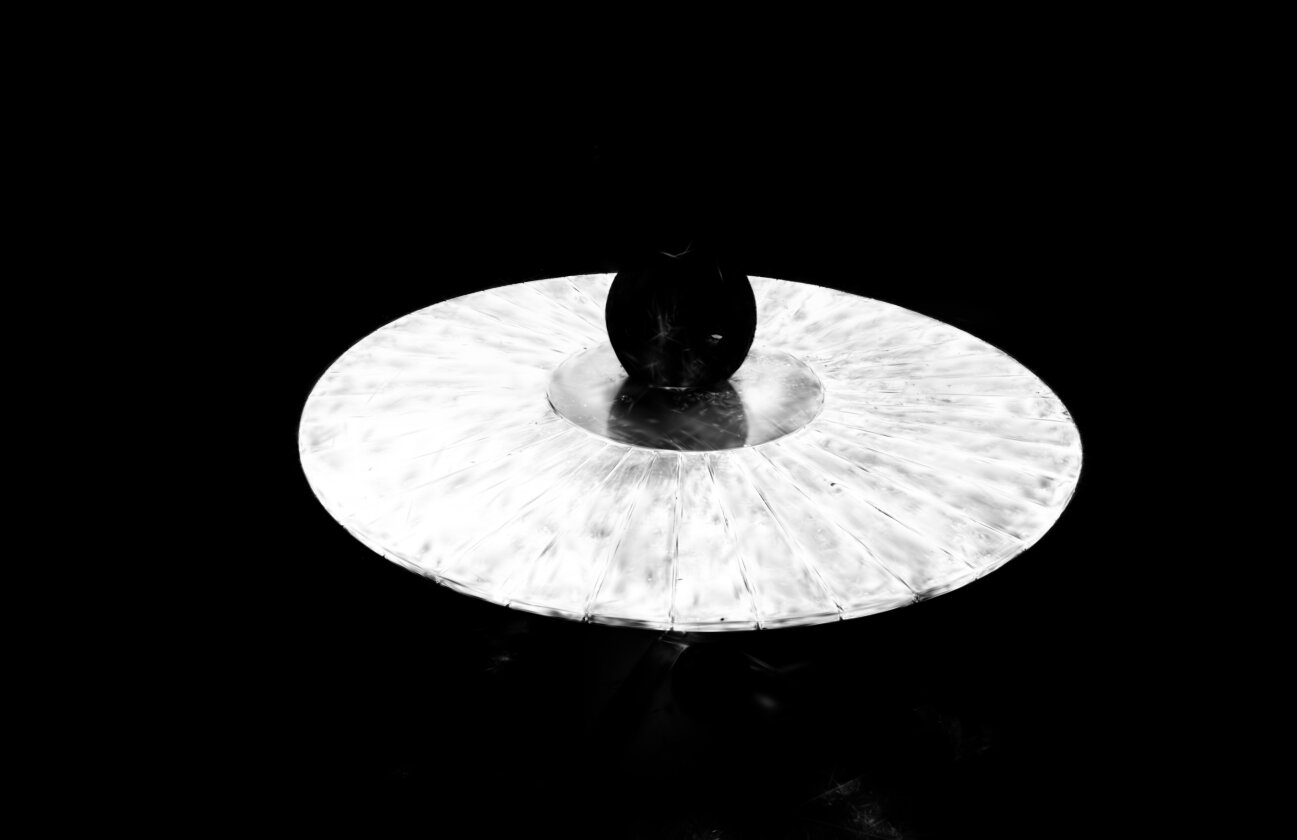} 
    \includegraphics[width=0.16\textwidth]{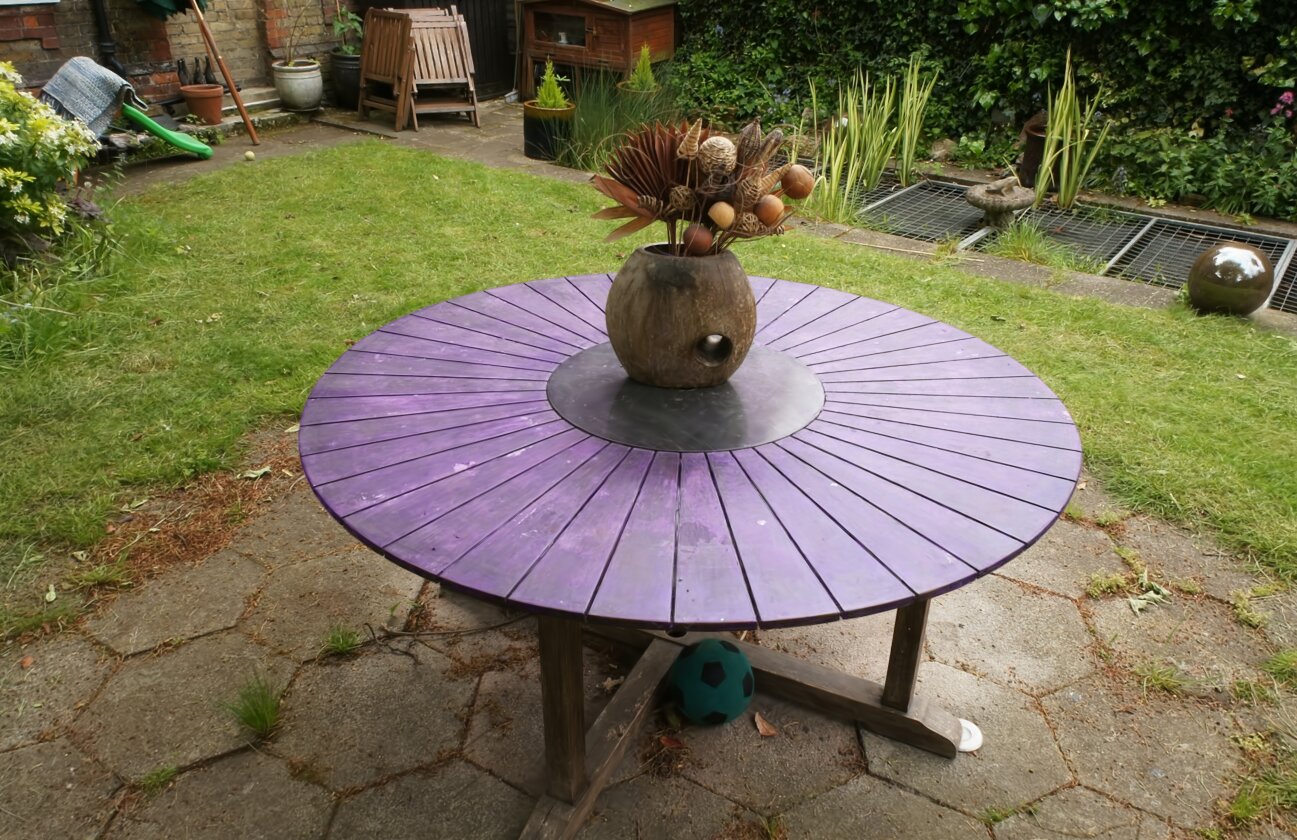}  
    \includegraphics[width=0.16\textwidth]{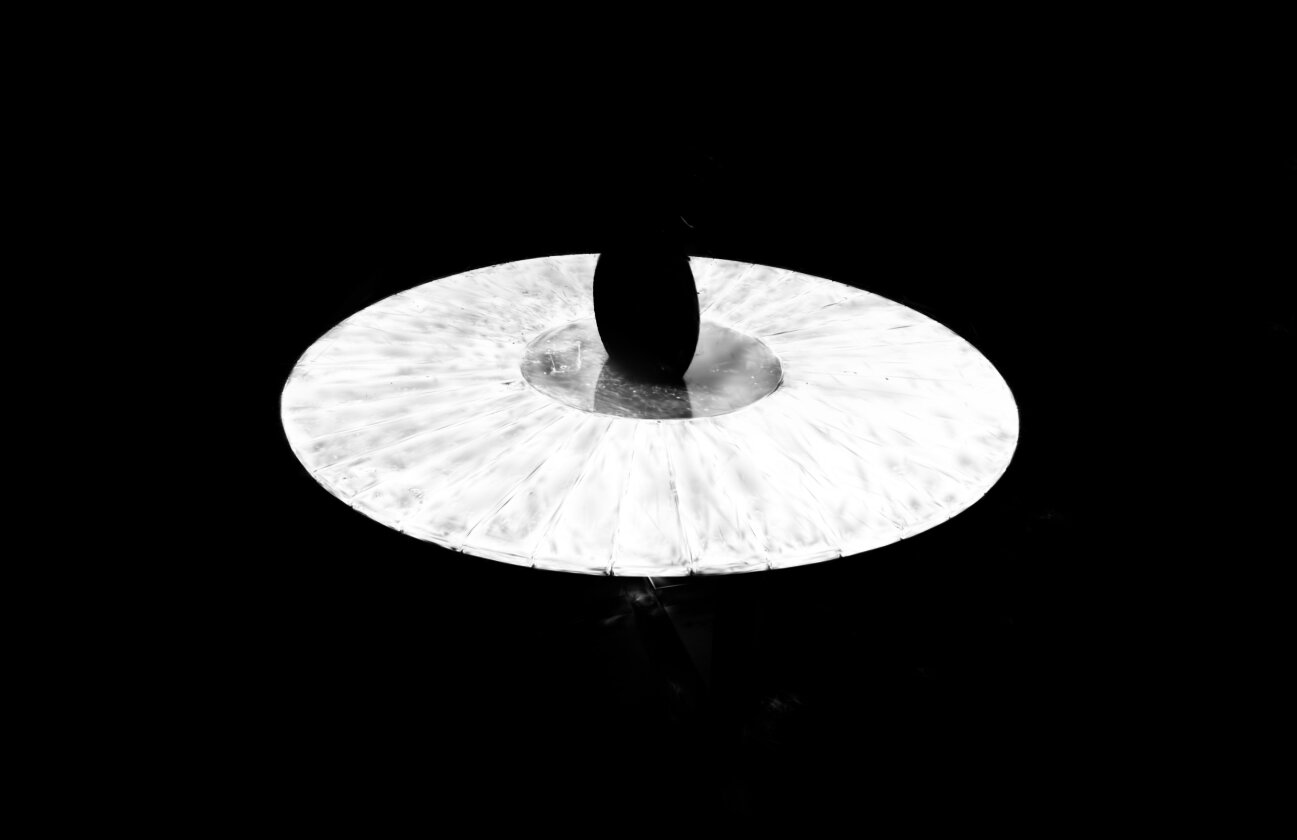} 
    \includegraphics[width=0.16\textwidth]{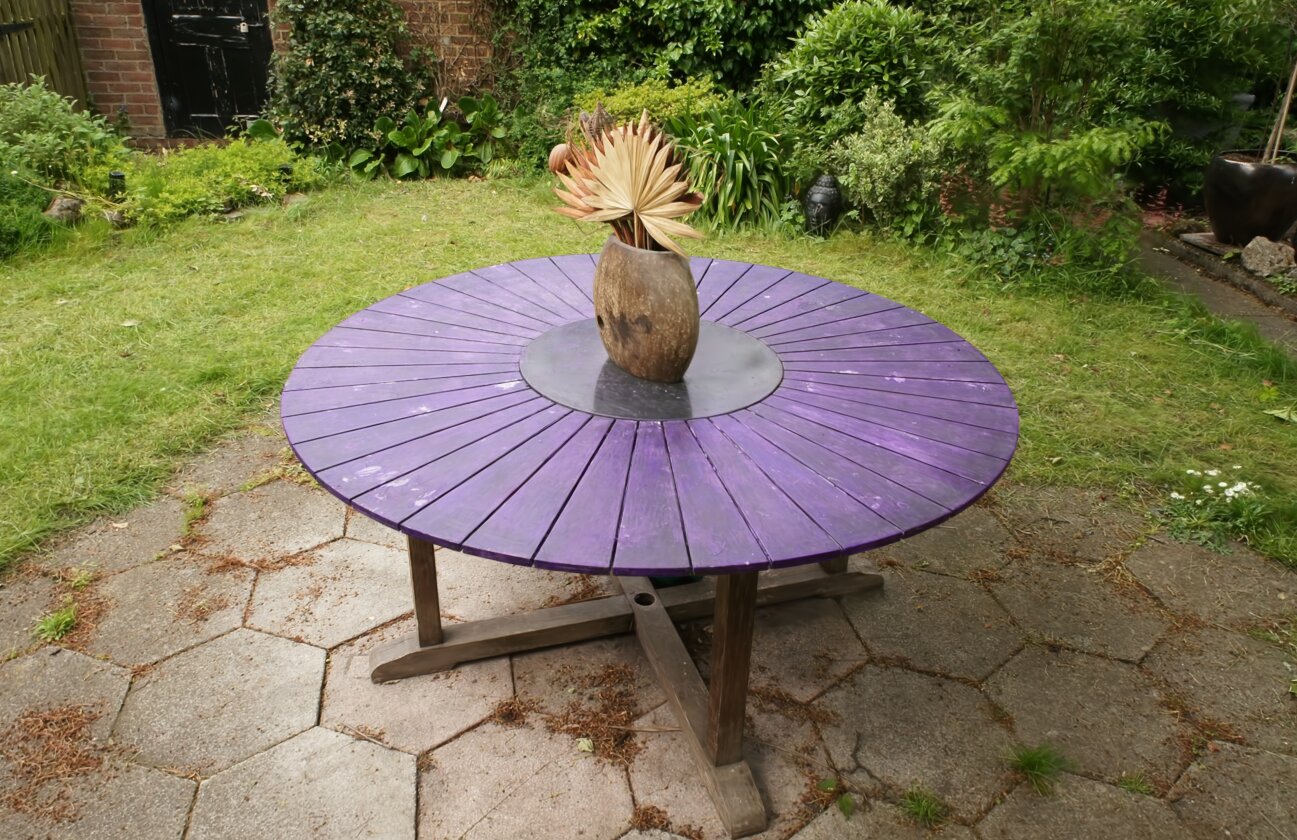} \\

    \includegraphics[width=0.16\textwidth]{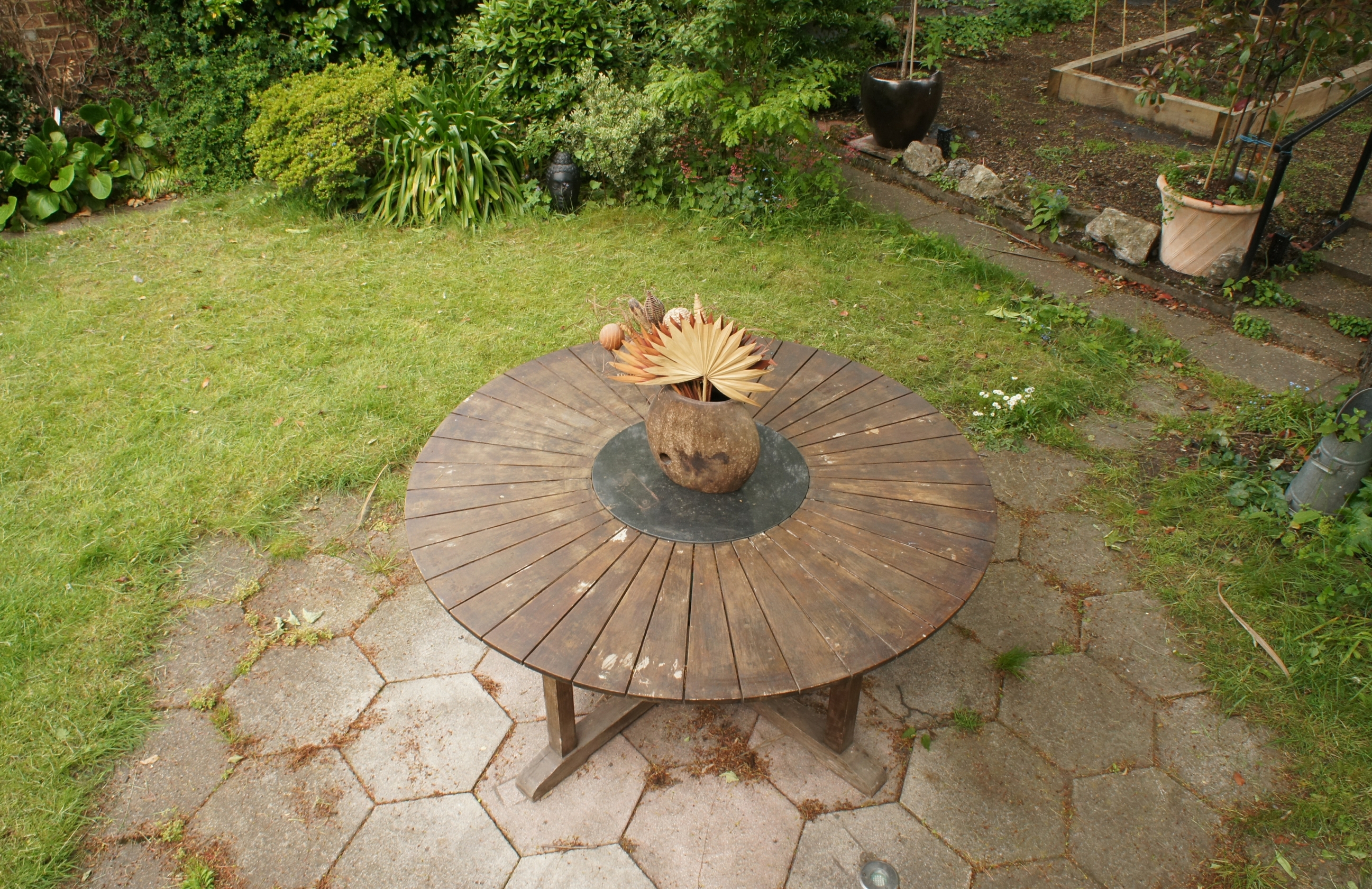}    \includegraphics[width=0.16\textwidth]{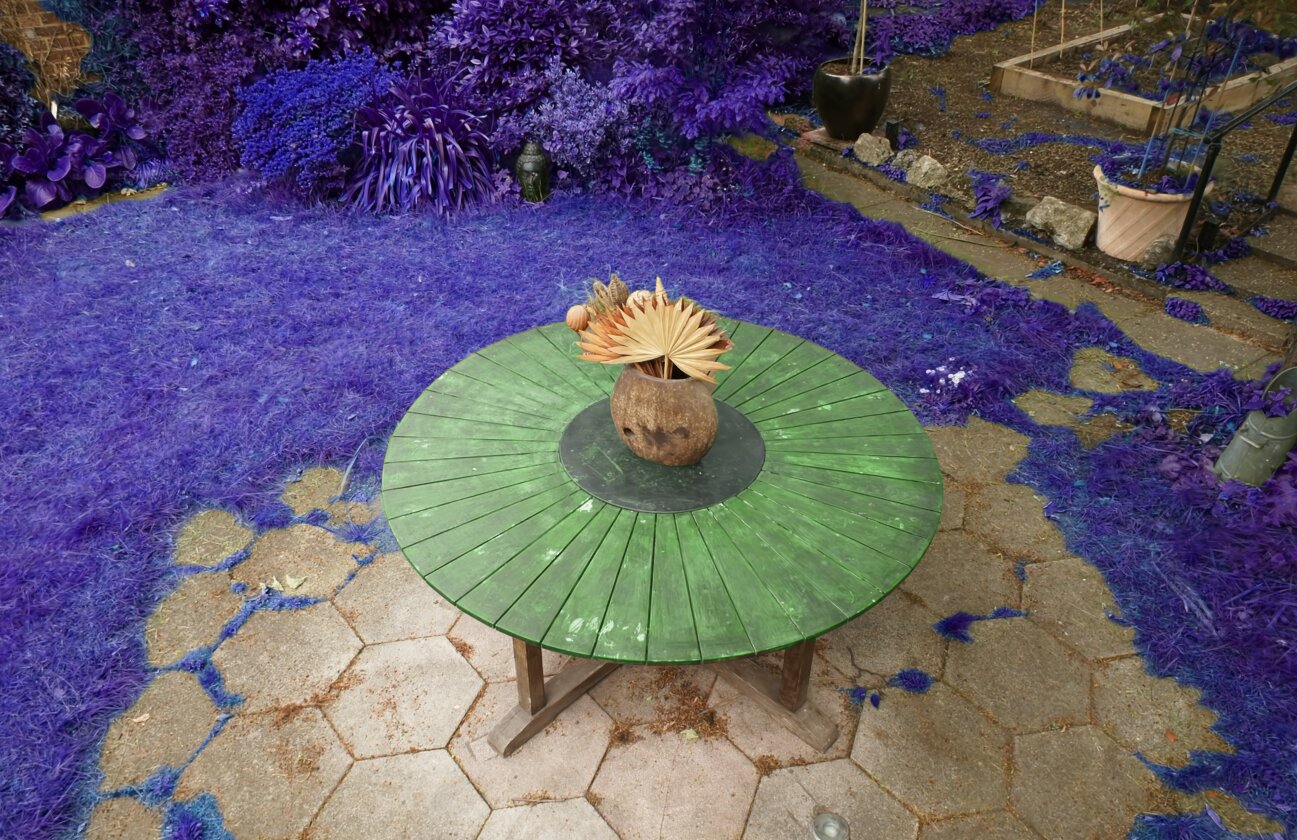}
    \includegraphics[width=0.16\textwidth]{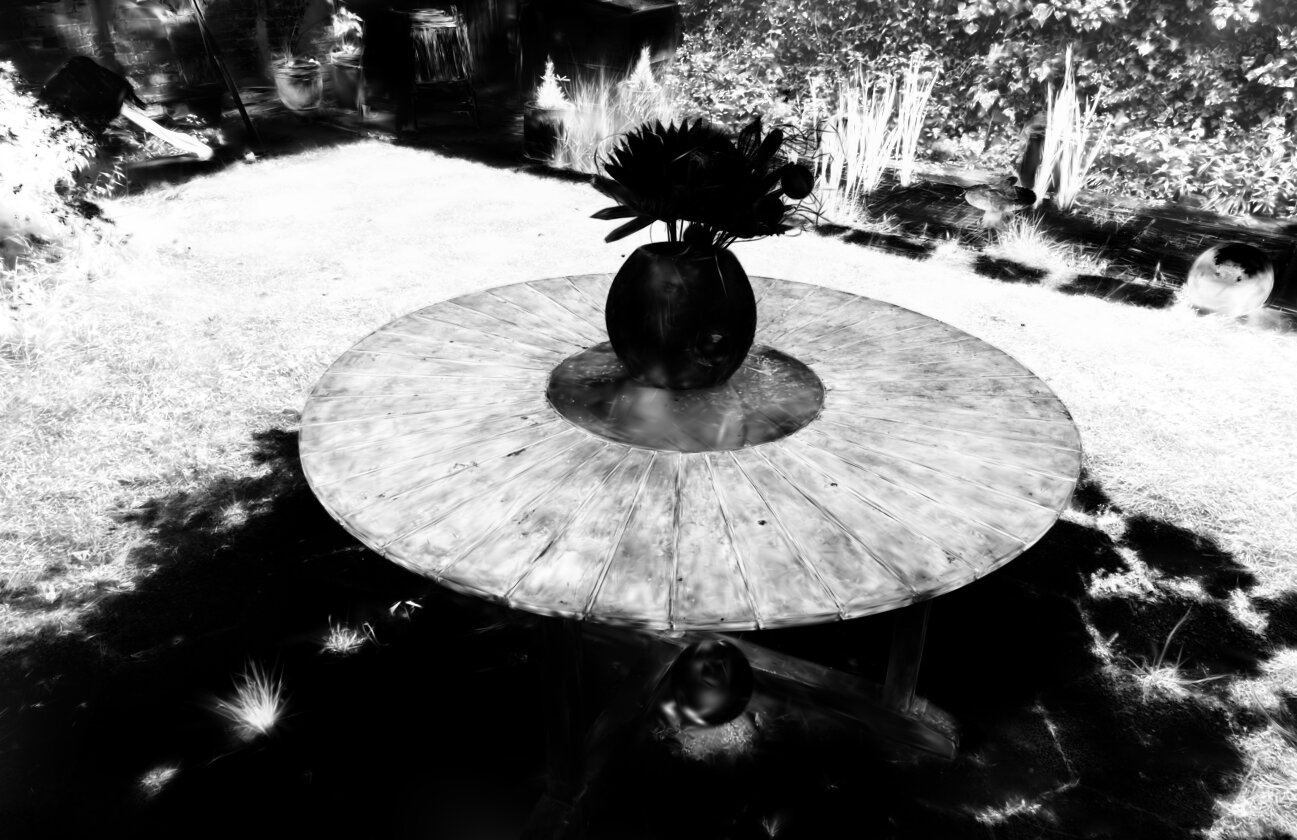}  
    \includegraphics[width=0.16\textwidth]{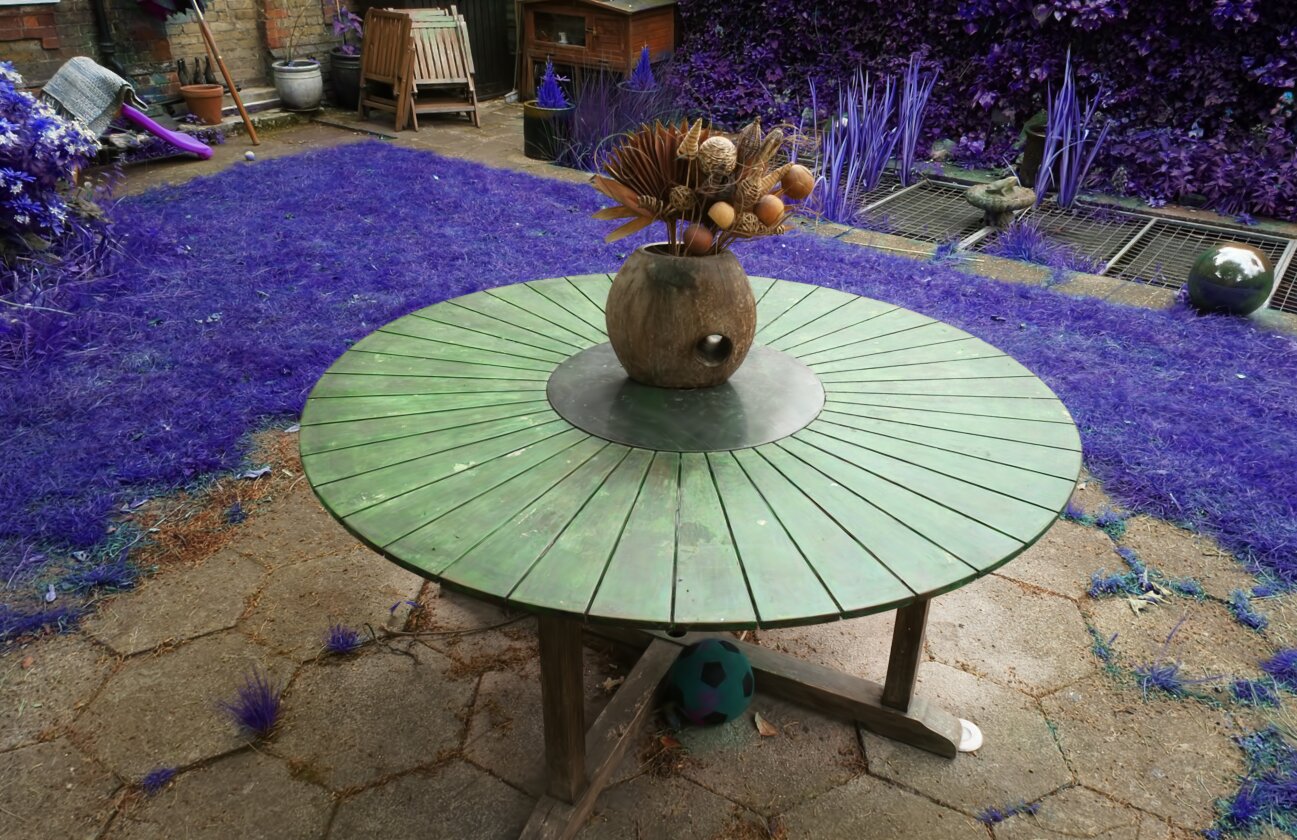} 
    \includegraphics[width=0.16\textwidth]{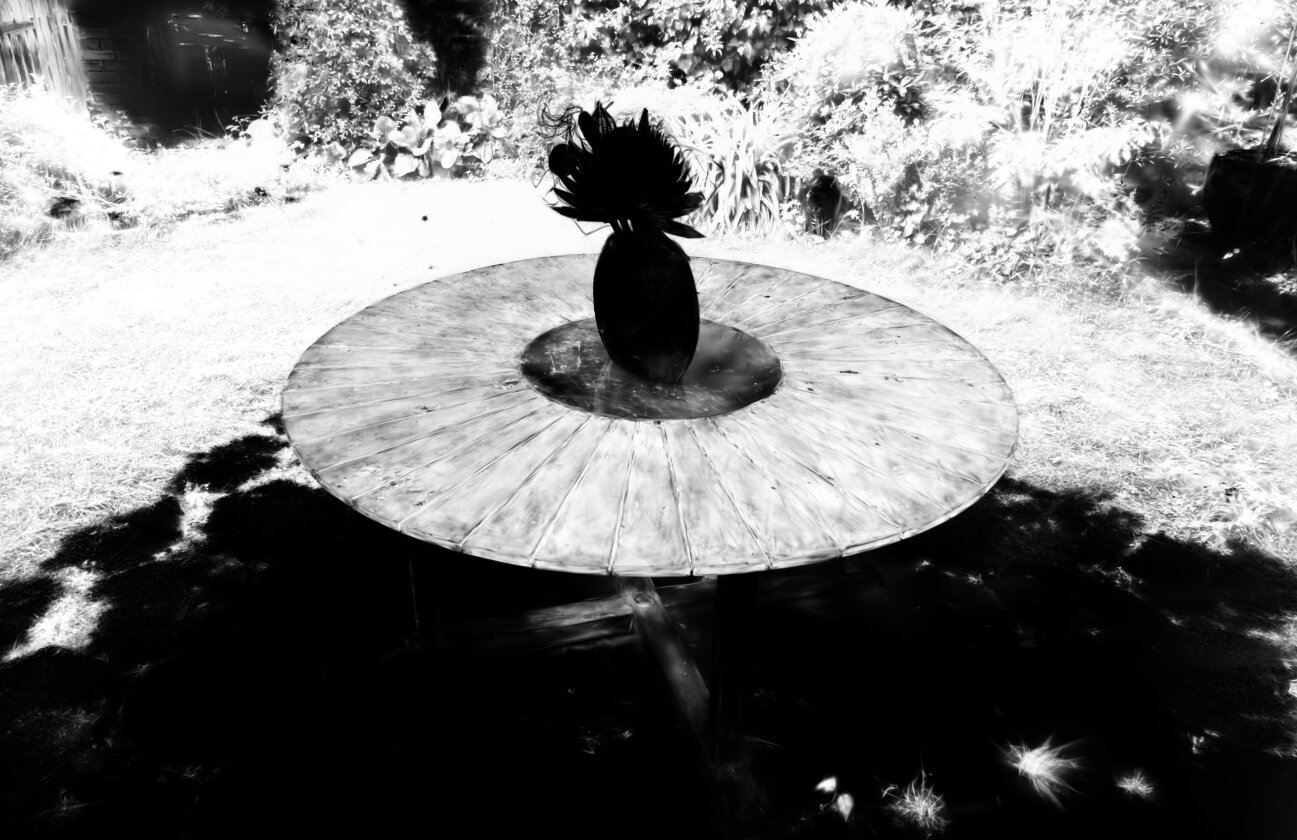} 
    \includegraphics[width=0.16\textwidth]{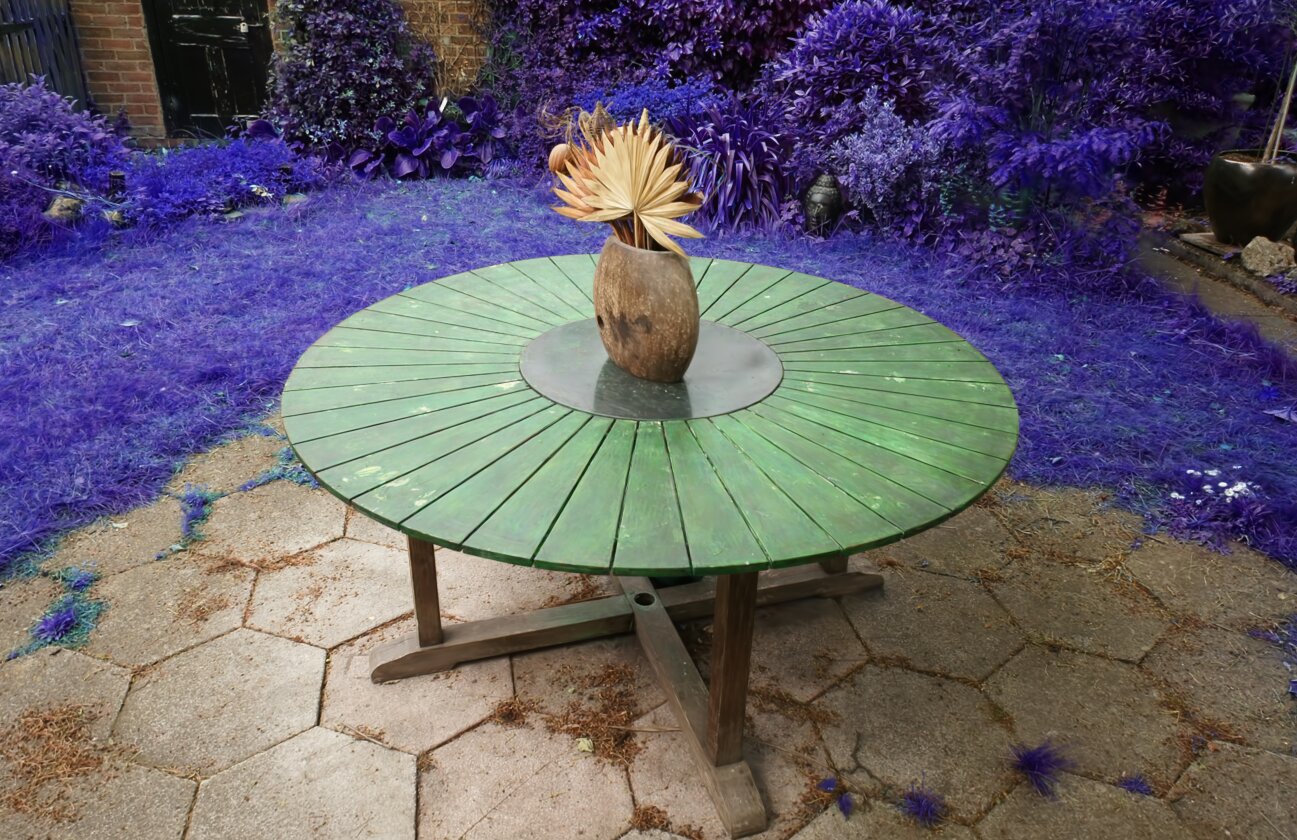} \\

    \includegraphics[width=0.16\textwidth]{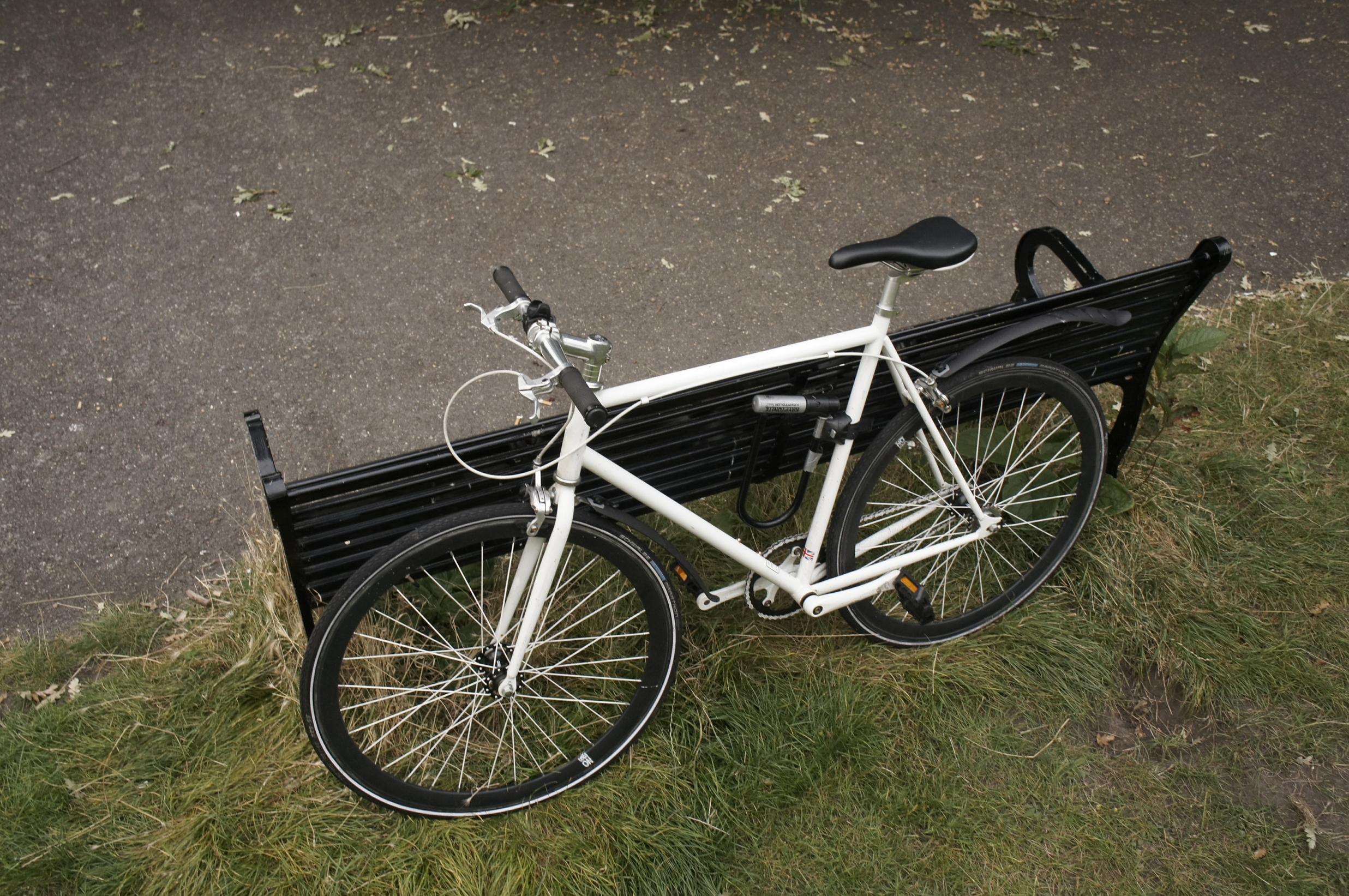}    \includegraphics[width=0.16\textwidth]{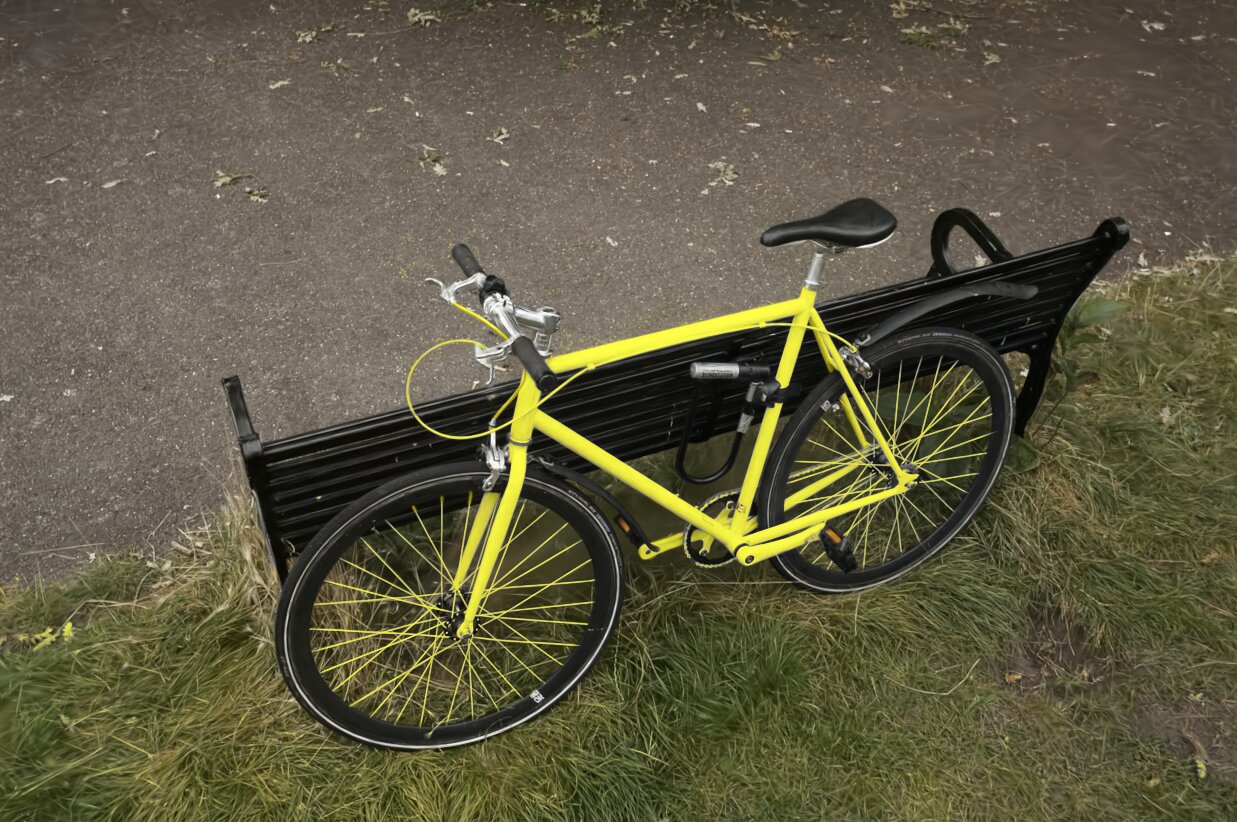} 
    \includegraphics[width=0.16\textwidth]{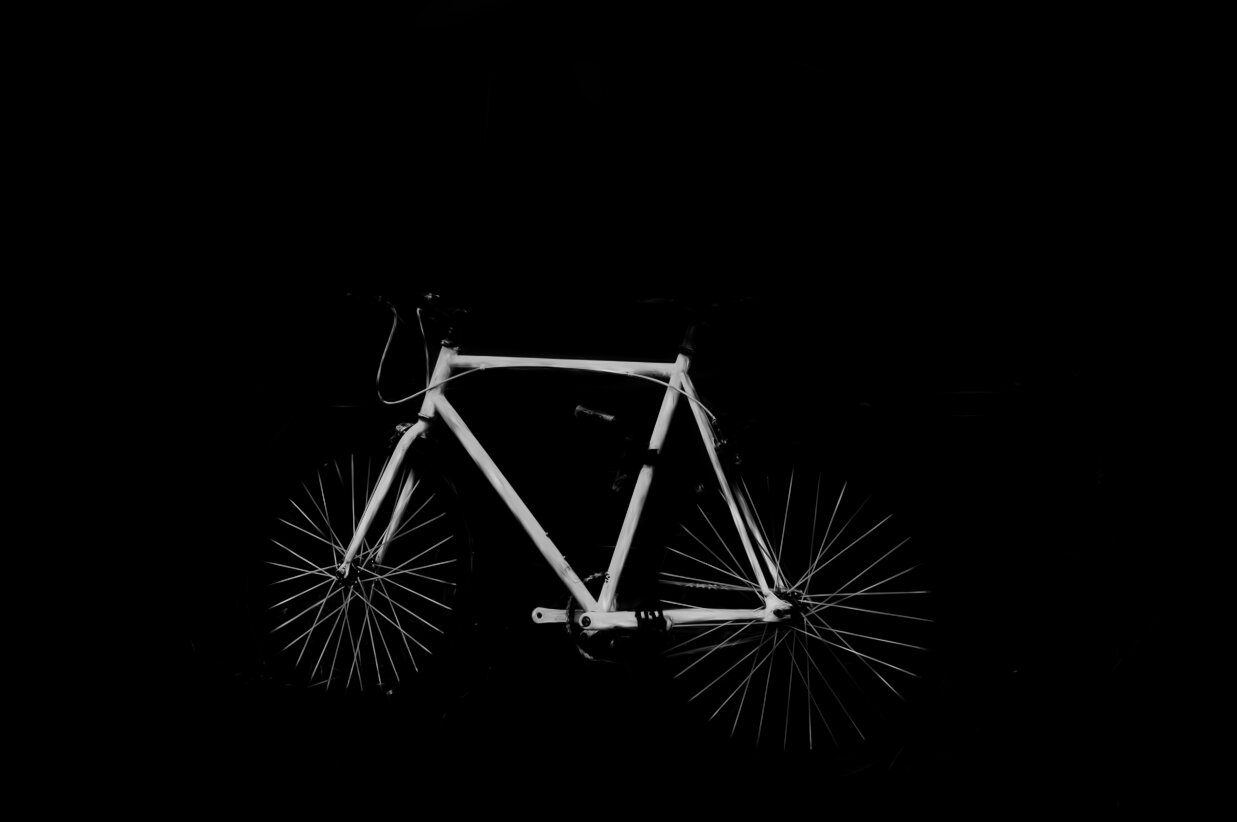} 
    \includegraphics[width=0.16\textwidth]{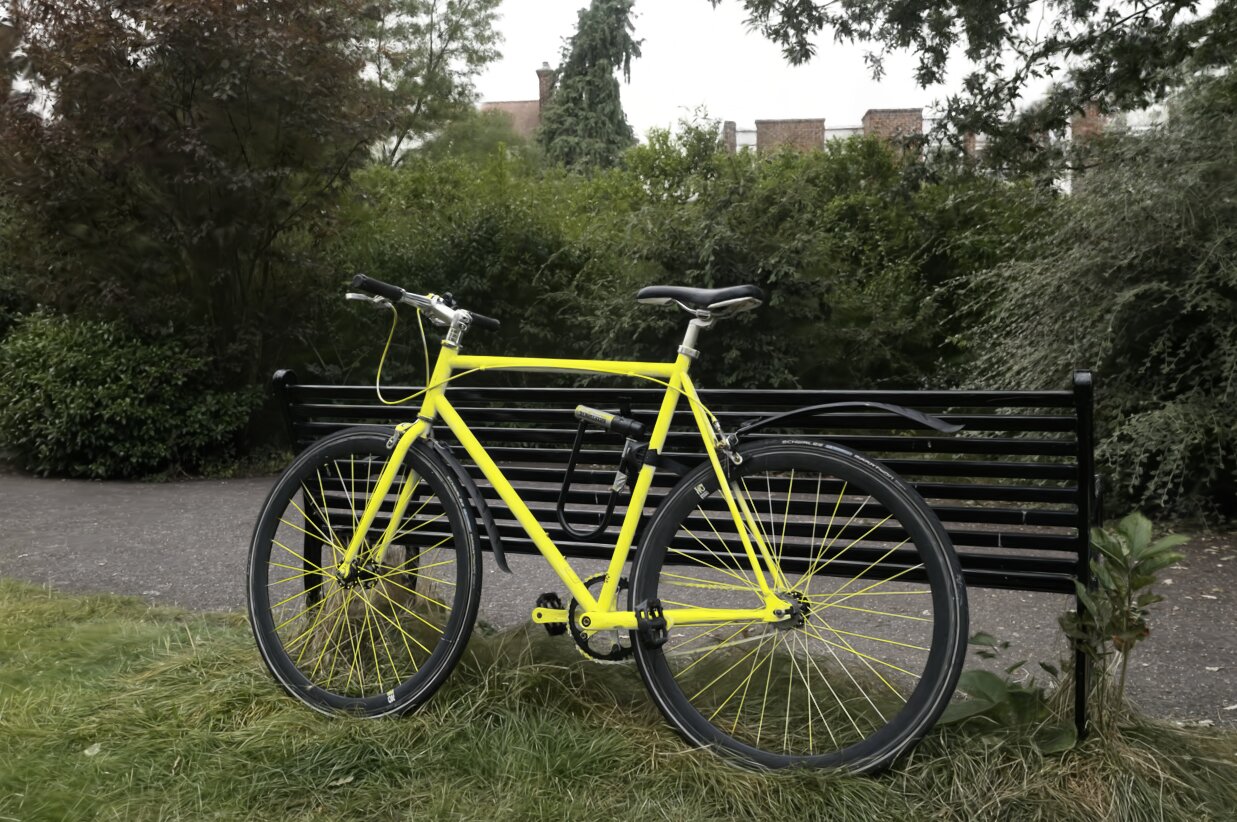} 
    \includegraphics[width=0.16\textwidth]{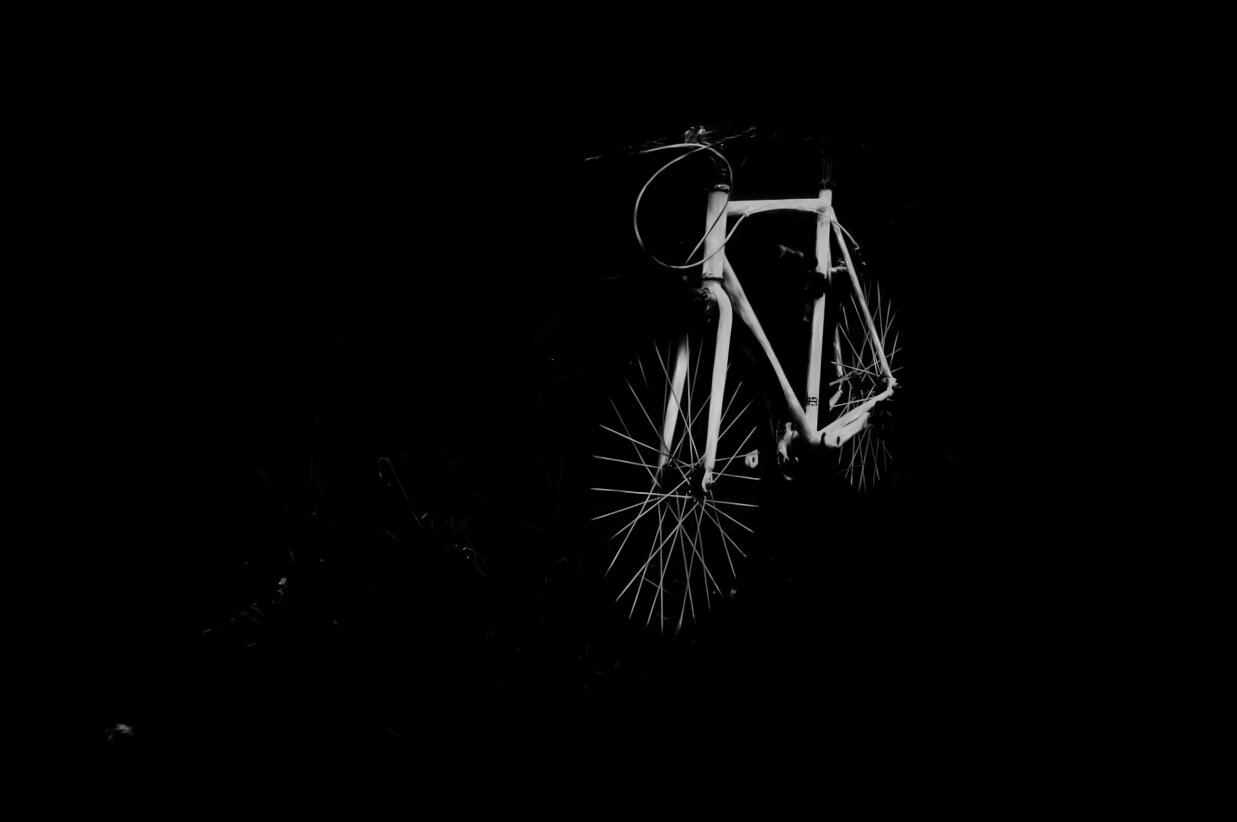}
    \includegraphics[width=0.16\textwidth]{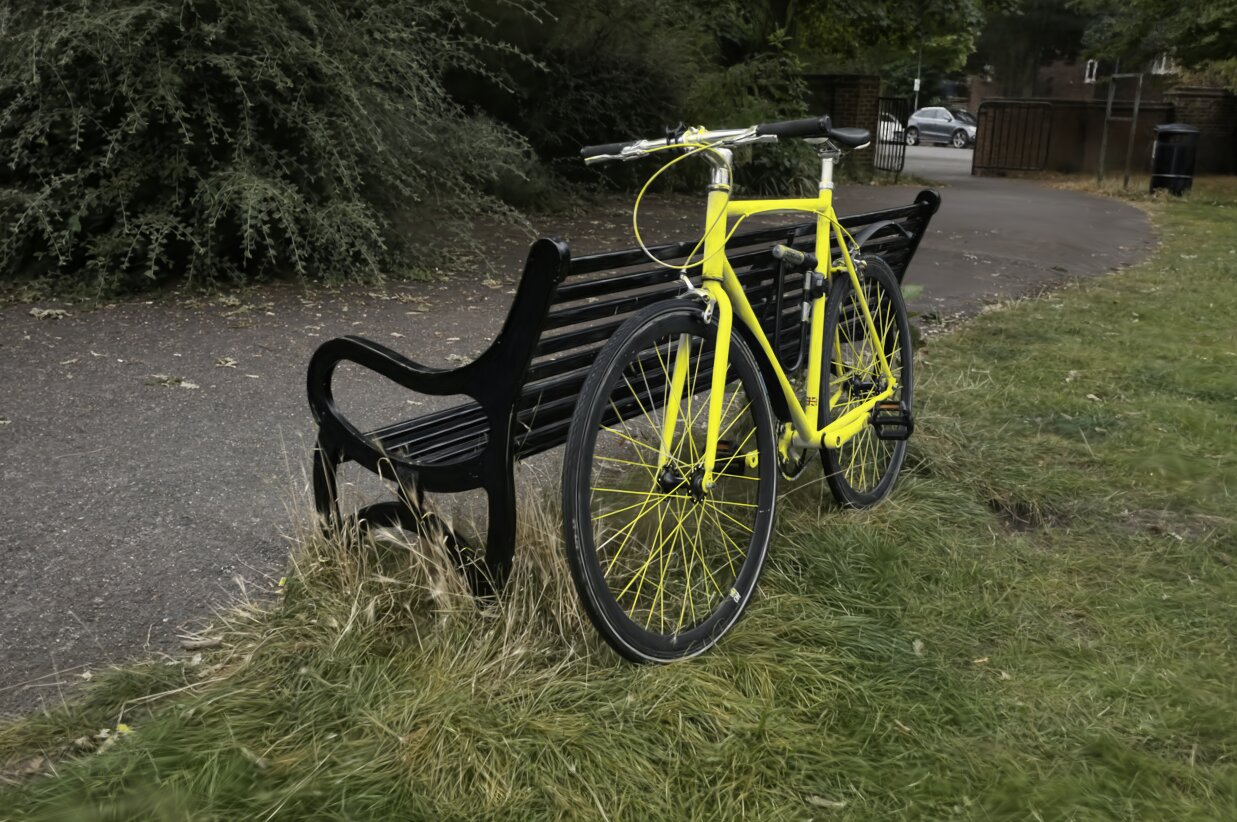} \\

    \includegraphics[width=0.16\textwidth]{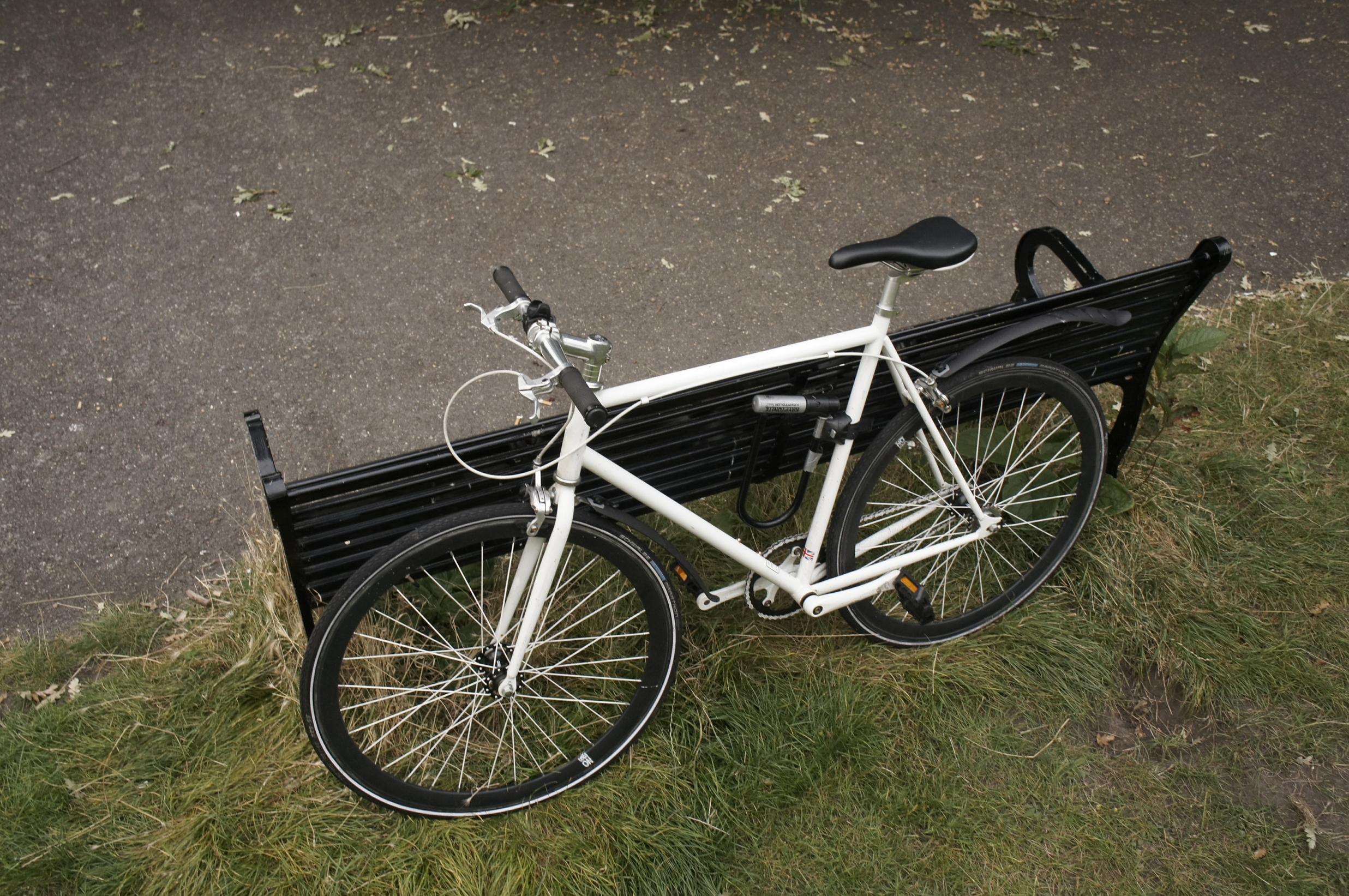}    \includegraphics[width=0.16\textwidth]{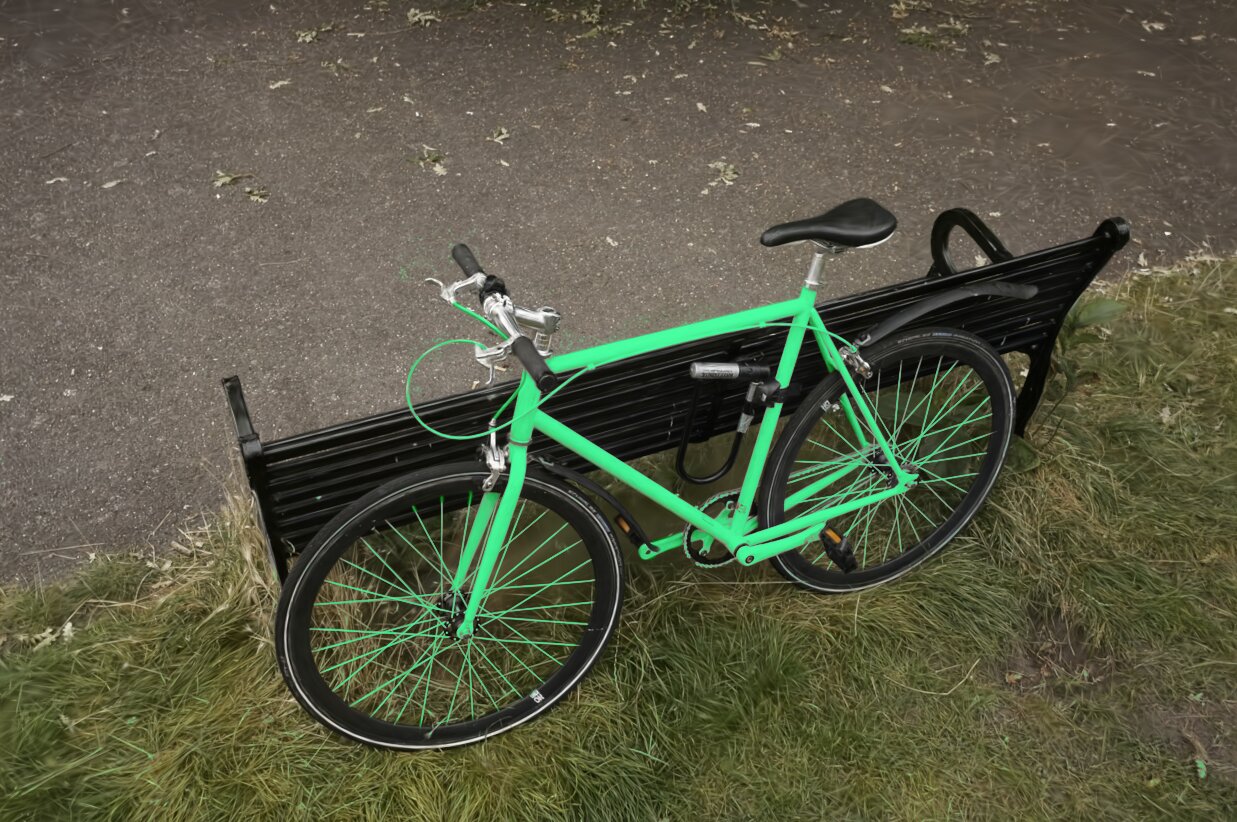} 
    \includegraphics[width=0.16\textwidth]{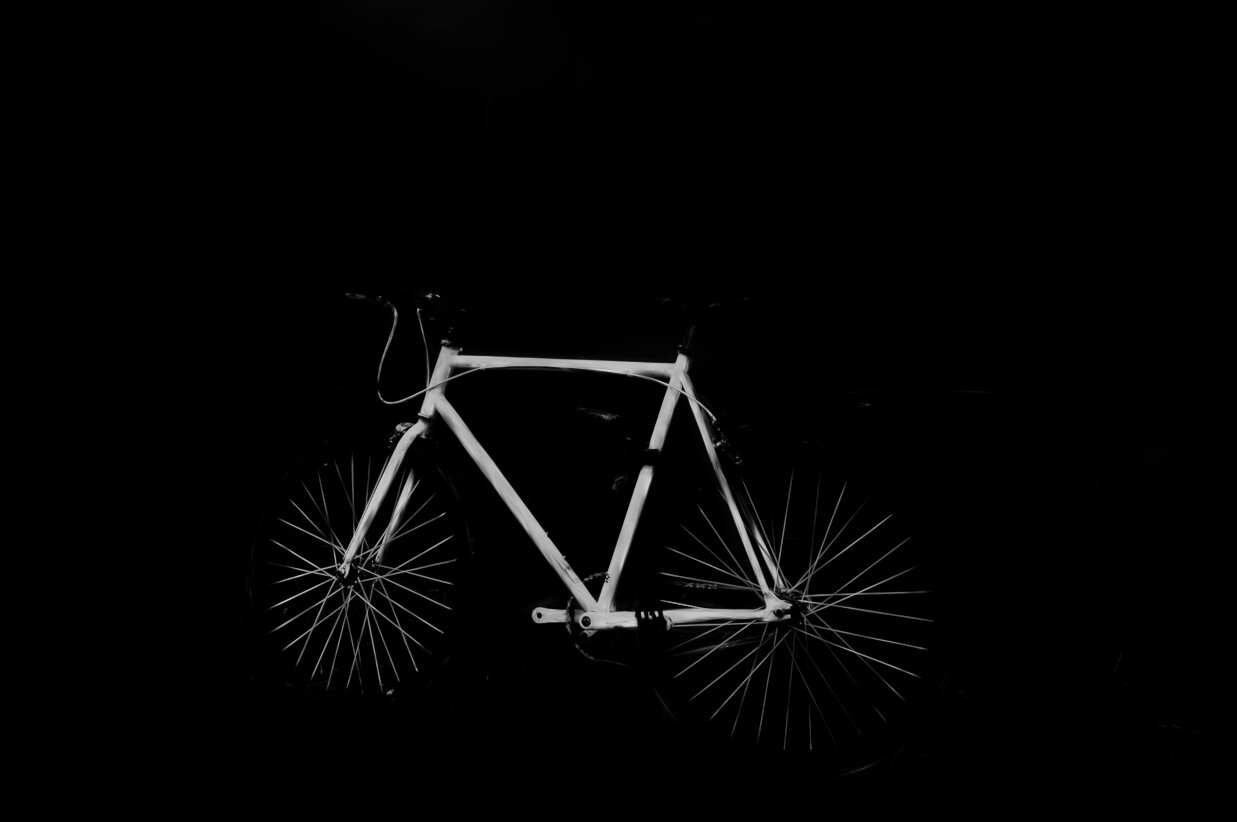} 
    \includegraphics[width=0.16\textwidth]{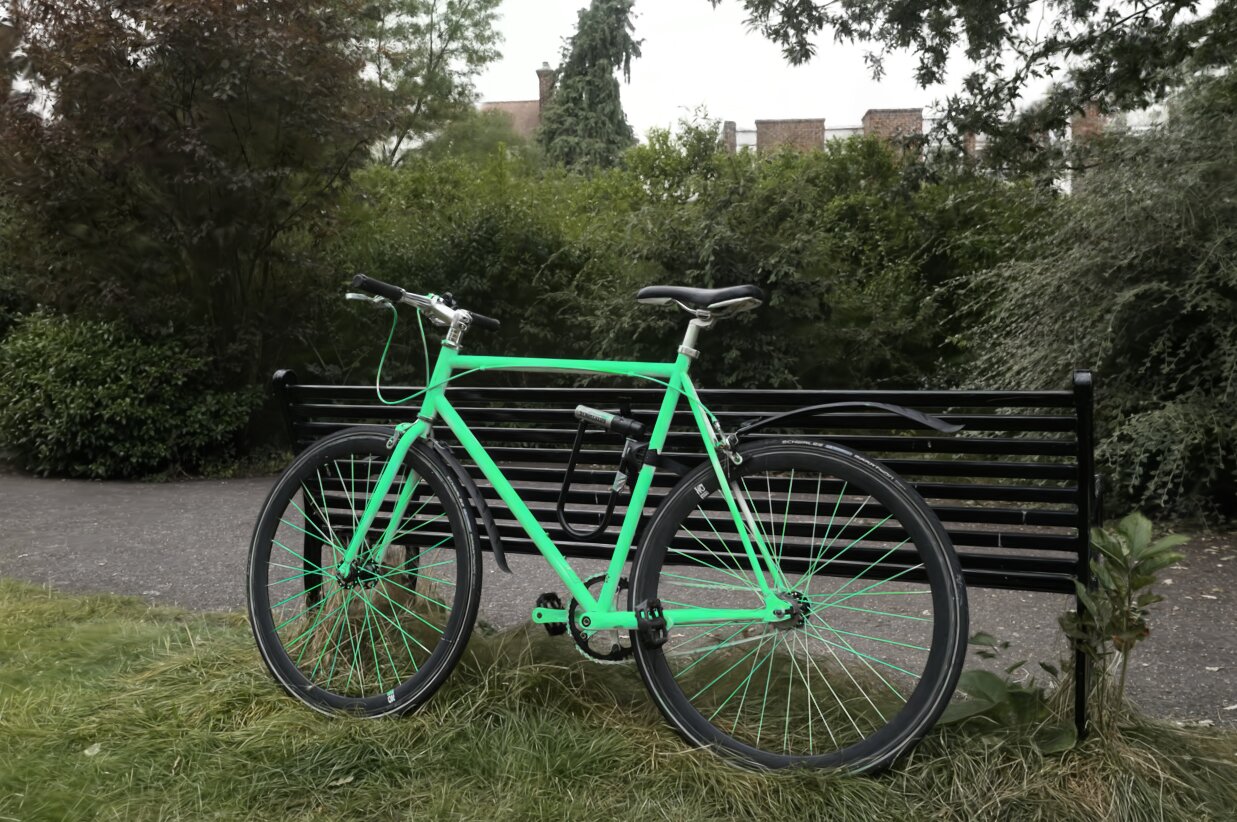} 
    \includegraphics[width=0.16\textwidth]{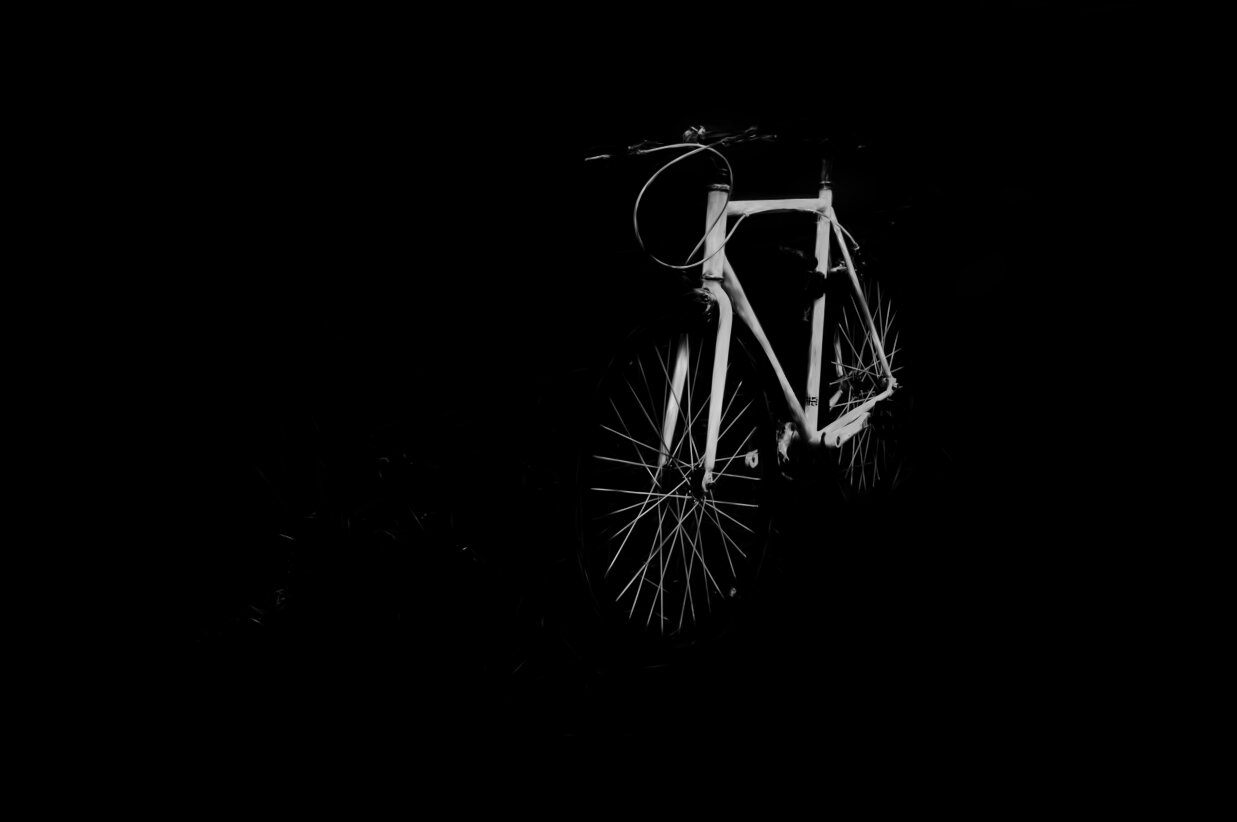} 
    \includegraphics[width=0.16\textwidth]{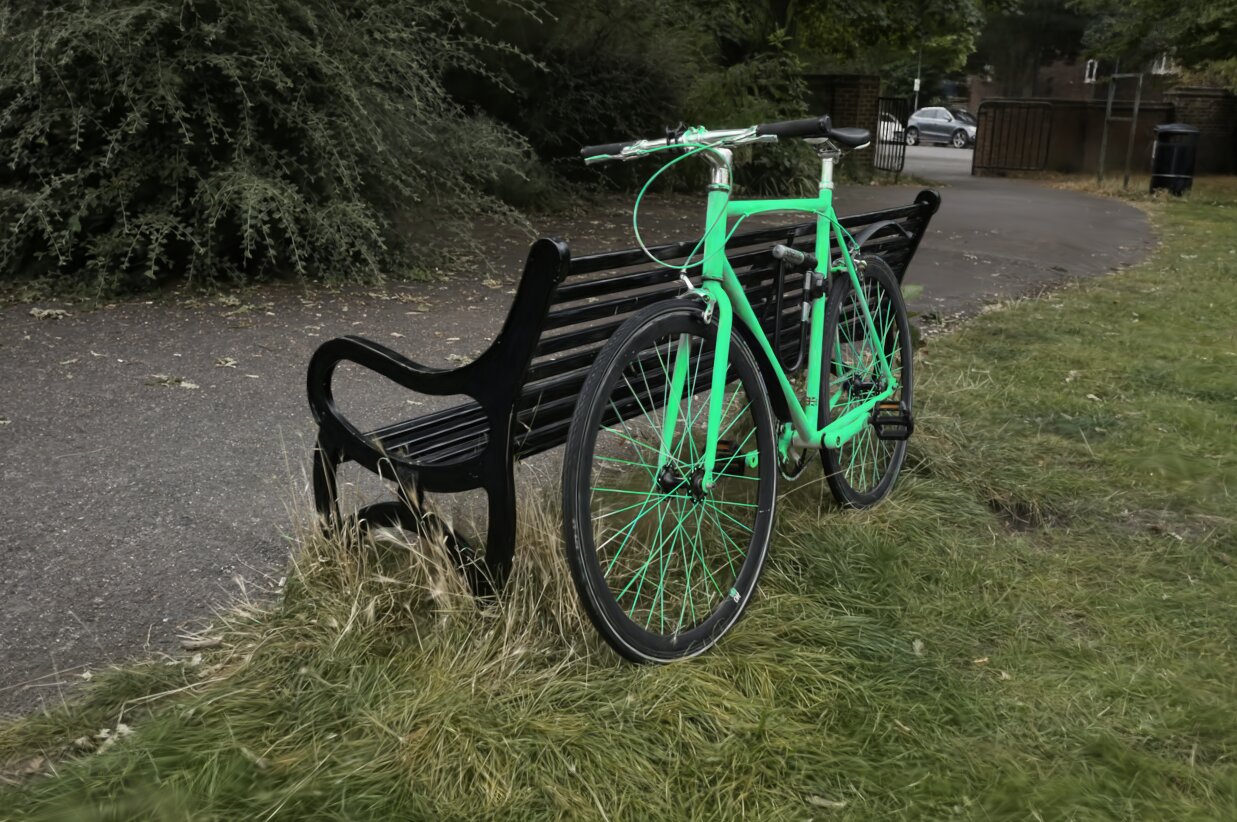} \\
    
\vspace{-1mm}

    \caption{{\bf Qualitative results.} Results on all scenes from the MipNeRF-360 dataset~\cite{barron2022mipnerf360}. This dataset shows complex real scenes with a complete 360 rotation on all scenes. We show several examples of single and multiple color editions. The first column is the original image. The second column is the edit performed by the user and used to train {\methodname}. We also show the soft segmentation predicted by {\methodname} for each recolored image.}
    \label{fig:qualitative_results}
    \centering
\end{figure*}

\noindent Fig.~\ref{fig:qualitative_results} showcases  qualitative results obtained from the MipNeRF-360 dataset~\cite{barron2022mipnerf360}. This dataset features complex real scenes with complete 360° coverage. The   edits produced by \methodname~ demonstrate consistency across viewpoints and effectively capture the intricate view-dependent effects present in the scenes. Additionally, the 3D segmentation masks for each recolored region, estimated by our method, are presented, further illustrating its capability to achieve 3D consistency.

\begin{figure}[t]
    \centering
    \includegraphics[width=0.98\linewidth]{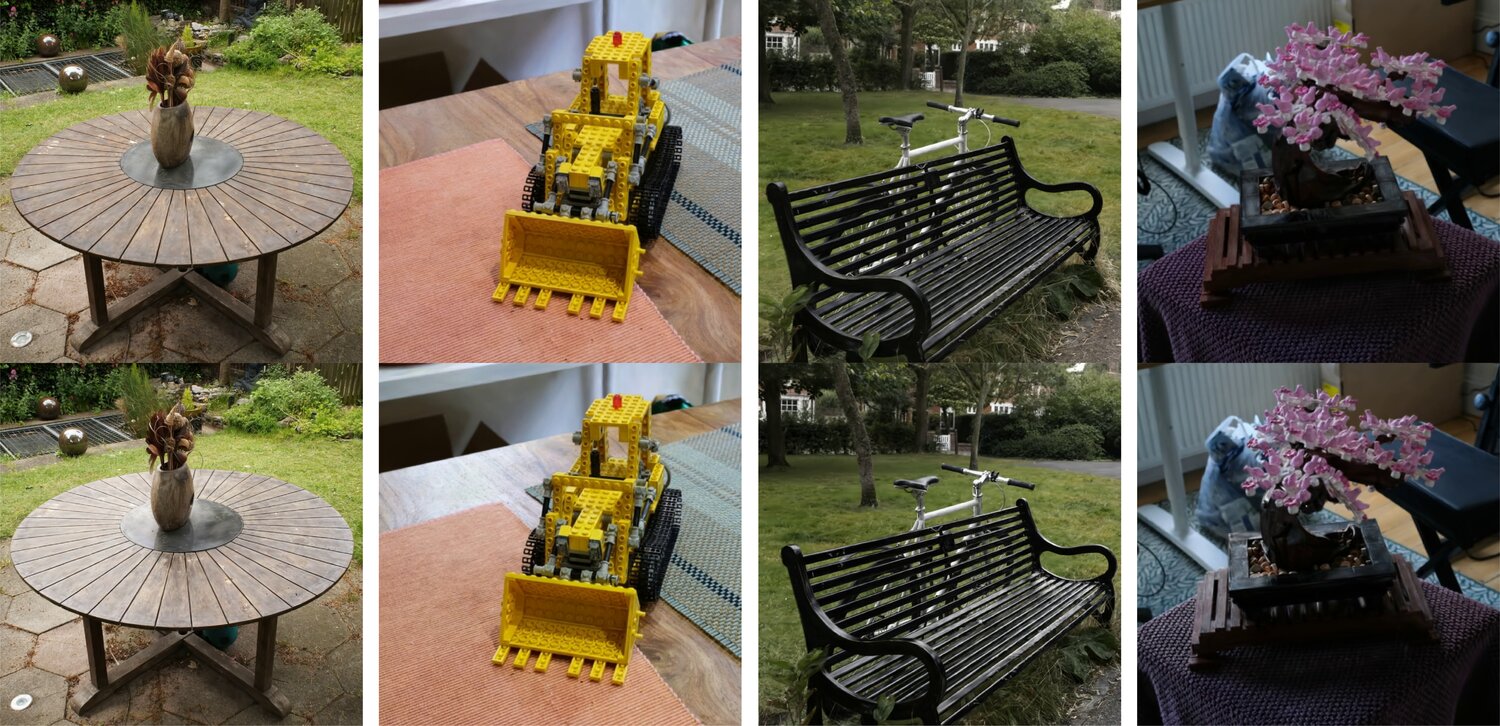} 
\vspace{-1mm}
    \caption{{\bf Illumination editing with decoupled output.} Top row: specular component reduced by applying a multiplicative factor of $0$. Bottom row: specular component increased by applying a   factor of $1.5$. }
    \label{fig:illumination_example}
    \centering

\end{figure}

%Fig.~\ref{fig:qualitative_results} shows extensive qualitative results on the MipNeRF-360 dataset~\cite{barron2022mipnerf360}, where complex real scenes with a complete 360 coverage are used to demonstrate how the edits produced by \methodname are consistent among very different viewpoints and view-dependent effects. Furthermore, the 3D segmentation mask for each of the recolored regions estimated by our method is shown, further demonstrating its ability to obtain 3D consistent results. 

%We show qualitative results of our recoloring approach in Fig.~\ref{fig:qualitative_results}. In this image, we have focused on showing results on the most challenging of the three datasets, the MipNeRF-360 dataset~\cite{barron2022mipnerf360}. This dataset allows us to do real scenes and showcase the robustness of our approach to viewpoint changes. We also show results that involve one or several color editions. We also show for each recolored image the segmentation mask that our approach estimates.

\vspace{1mm}
\noindent{\bf Comparison with Gaussian Editor.}  
As discussed in Sect.~\ref{sec:related}, no other method specifically addresses the challenge of performing appearance edition on Gaussian Splatting. However, some methods enable the modification of the final color of specific elements using text prompts. Fig.~\ref{fig:GEditor_comparison} presents a qualitative comparison of \methodname~and GaussianEditor~\cite{chen2023gaussianeditor}. GaussianEditor facilitates single-element recoloring by leveraging SAM~\cite{Kirillov2023SegmentA} to generate a mask of the desired object. Subsequently, guided by a text prompt and employing InstructPix2Pix~\cite{brooks2022instructpix2pix}, the scene edit is optimized until satisfactory results are achieved.
As shown, our method is more consistently more robust. Notably, GaussianEditor exhibits color bleeding into undesired areas of the scene, a limitation not observed in \methodname's results. Additionally, while GaussianEditor requires approximately 10 minutes to complete a given edit, our approach achieves the same task in just 2 seconds, significantly enhancing its interactivity and usability.

\vspace{1mm}
\noindent{\bf Specular editing.} In Fig.~\ref{fig:illumination_example}, we illustrate how decoupling the color output into diffuse and specular components enables fine adjustments to view-dependent effects. By manipulating the strength of the specular component $\Cspec$, we can make subtle changes in the material appearance. %It's important to note that while our approach allows for these modifications, it's not a replacement for comprehensive illumination modeling. 
Although the range of editions is limited, this experiment provides valuable insights into our model's understanding of scene appearance, a key aspect of its superiority over other recoloring methods.

%\noindent{\bf Specular editing.} In Fig.~\ref{fig:illumination_example} we provide an example of how by decoupling the color output in diffuse and specular, small changes in the view-dependent effects can be performed on the scene by increasing or decreasing the specular color estimation $\Cspec$. We show the results of this kind of edition right after the decoupled output has been learned, and before applying recoloring. We want to be clear that our work is no substitute for proper illumination modelization, we decouple the output because it helps to understand how to recolor the scene. However, we believe that this experiment gives insight into how much our model is able to understand the appearance of the given scene, which is a very important part of its superiority compared to other recoloring approaches.
%\noindent{\bf Specular editing.} In Fig.~\ref{fig:illumination_example} we provide an example of how by decoupling the color output in diffuse and specular, small changes in illumination can be performed on the scene by increasing or decreasing the specular color estimation $\Cspec$. We show the results on illumination changes right after the decoupled output has been learned, and before applying recoloring. We want to be clear that our work is no substitute for proper illumination modelisation, we decouple the output because it helps to understand how to recolor the scene. But such experiment gives insight on how much our model has understood from the scene.

\vspace{1mm}
\subsection{Limitations}
\begin{figure*}[t]
\centering

\begin{minipage}[b]{0.24\textwidth}
\centering
\small Ground Truth
\end{minipage}
\begin{minipage}[b]{0.24\textwidth}
\centering
\small \methodname{} Diffuse
\end{minipage}
\begin{minipage}[b]{0.24\textwidth}
\centering
\small \methodname{} Specular
\end{minipage}
\begin{minipage}[b]{0.24\textwidth}
\centering
\small \methodname{} Combined RGB
\end{minipage}

\vspace{1mm}

\begin{minipage}[b]{0.24\textwidth}
\centering
\includegraphics[width=\linewidth,height=0.22\textheight,keepaspectratio]{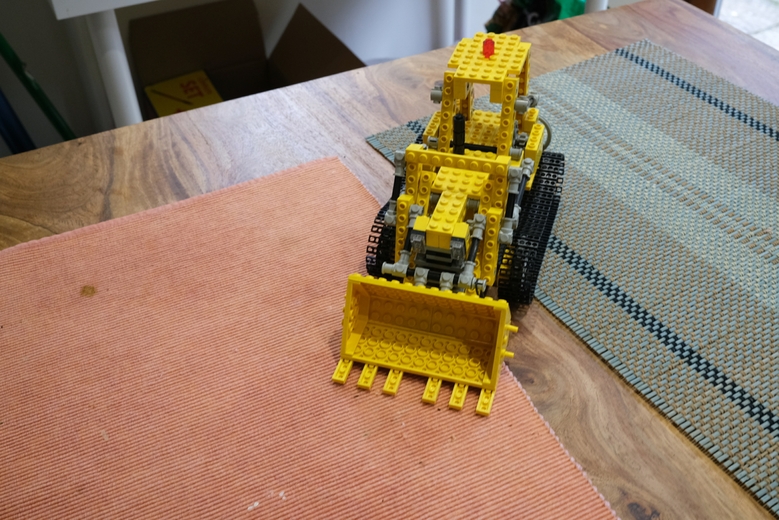}
\end{minipage}
\begin{minipage}[b]{0.24\textwidth}
\centering
\includegraphics[width=\linewidth,height=0.22\textheight,keepaspectratio]{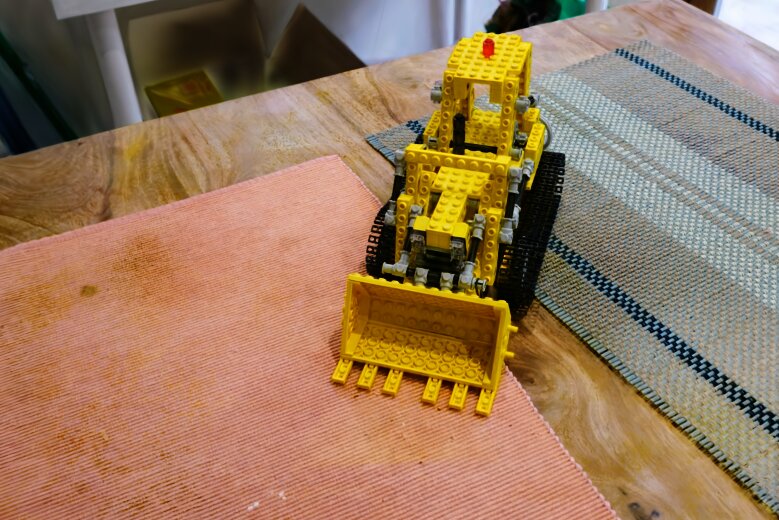}
\end{minipage}
\begin{minipage}[b]{0.24\textwidth}
\centering
\includegraphics[width=\linewidth,height=0.22\textheight,keepaspectratio]{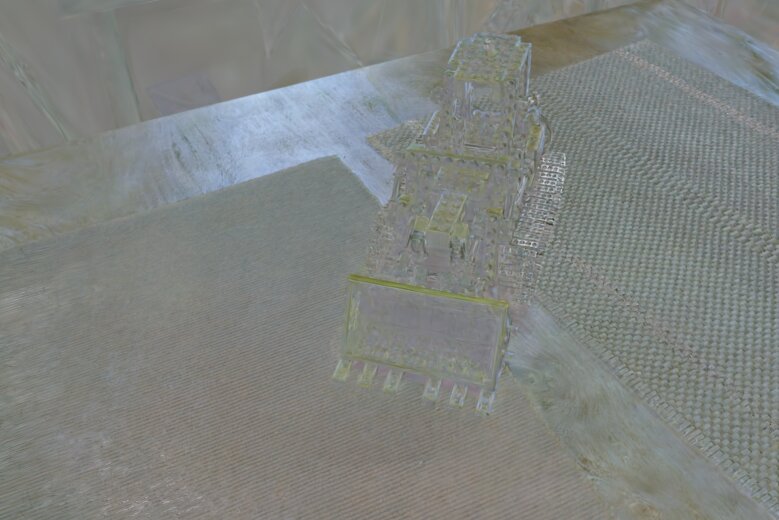}
\end{minipage}
\begin{minipage}[b]{0.24\textwidth}
\centering
\includegraphics[width=\linewidth,height=0.22\textheight,keepaspectratio]{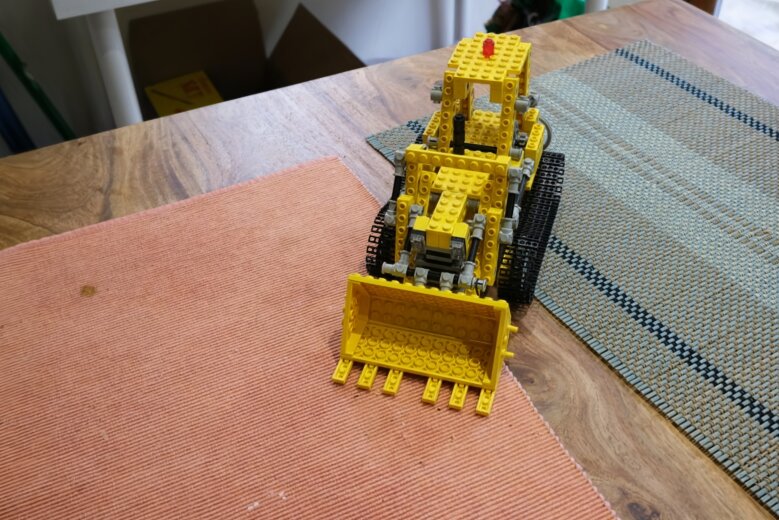}
\end{minipage}

\vspace{1mm} % <-- Added space between the two image groups

% Row of images (first scene)
\begin{minipage}[b]{0.24\textwidth}
\centering
\includegraphics[width=\linewidth,height=0.22\textheight,keepaspectratio]{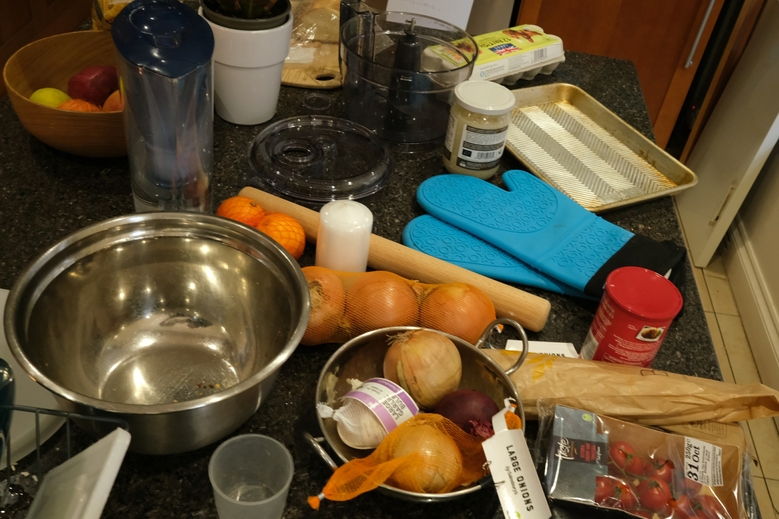}
\end{minipage}
\begin{minipage}[b]{0.24\textwidth}
\centering
\includegraphics[width=\linewidth,height=0.22\textheight,keepaspectratio]{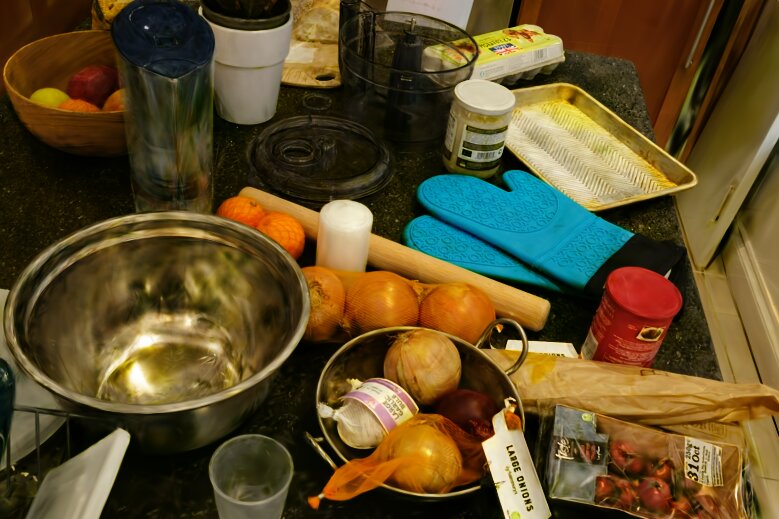}
\end{minipage}
\begin{minipage}[b]{0.24\textwidth}
\centering
\includegraphics[width=\linewidth,height=0.22\textheight,keepaspectratio]{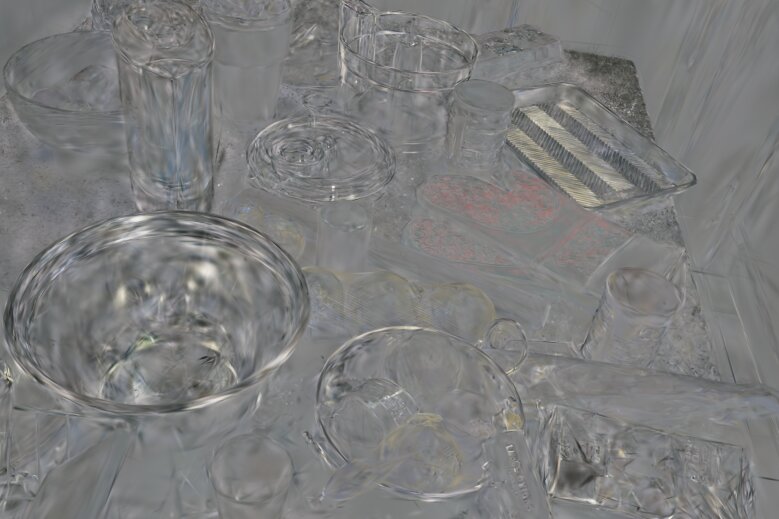}
\end{minipage}
\begin{minipage}[b]{0.24\textwidth}
\centering
\includegraphics[width=\linewidth,height=0.22\textheight,keepaspectratio]{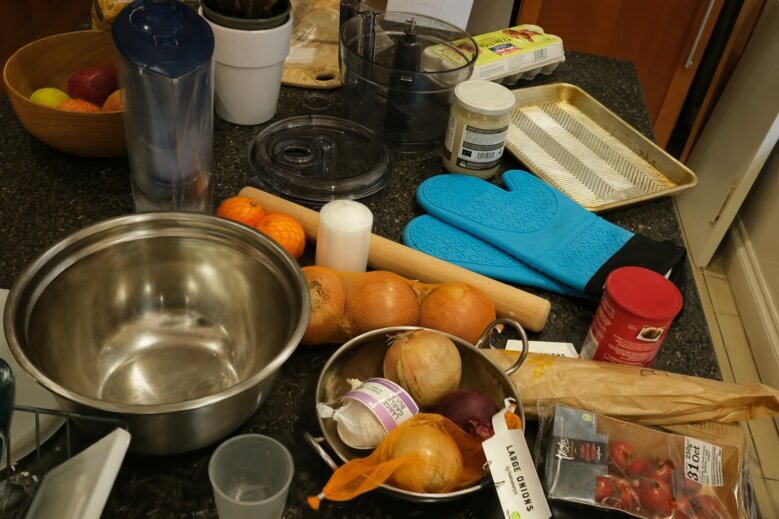}
\end{minipage}

\vspace{1mm}

% Row of titles
\begin{minipage}[b]{0.24\textwidth}
\centering
\small Ground Truth
\end{minipage}
\begin{minipage}[b]{0.24\textwidth}
\centering
\small Edit
\end{minipage}
\begin{minipage}[b]{0.24\textwidth}
\centering
\small \methodname{} Edited pose render
\end{minipage}
\begin{minipage}[b]{0.24\textwidth}
\centering
\small \methodname{} Novel View render
\end{minipage}

\vspace{1mm}

% Row of images
%kitchen
\begin{minipage}[b]{0.24\textwidth}
\centering
\includegraphics[width=\linewidth,height=0.22\textheight,keepaspectratio]{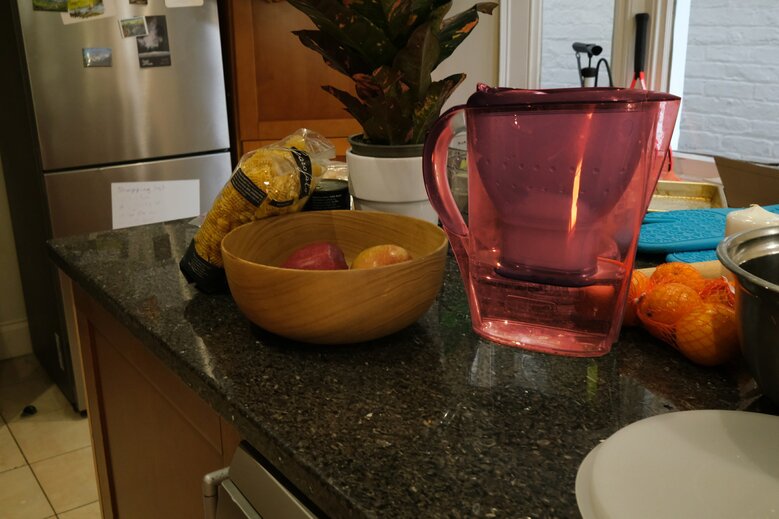}
\end{minipage}
\begin{minipage}[b]{0.24\textwidth}
\centering
\includegraphics[width=\linewidth,height=0.22\textheight,keepaspectratio]{figures/limitations/DSCF6047.jpg}
\end{minipage}
\begin{minipage}[b]{0.24\textwidth}
\centering
\includegraphics[width=\linewidth,height=0.22\textheight,keepaspectratio]{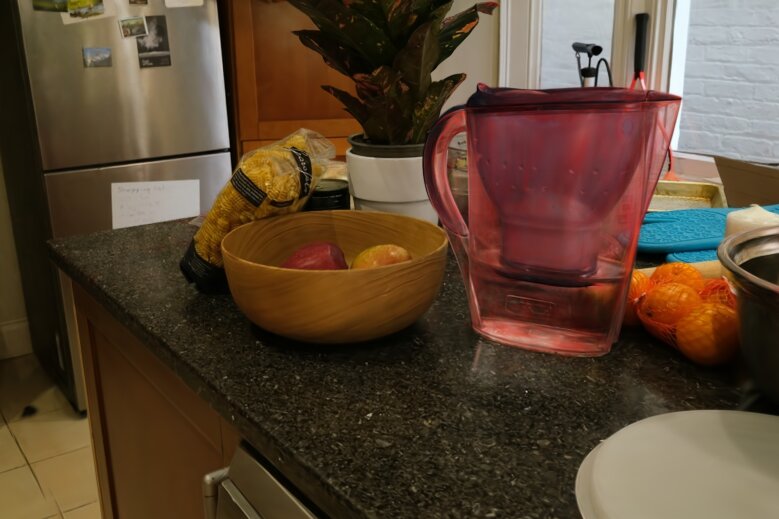}
\end{minipage}
\begin{minipage}[b]{0.24\textwidth}
\centering
\includegraphics[width=\linewidth,height=0.22\textheight,keepaspectratio]{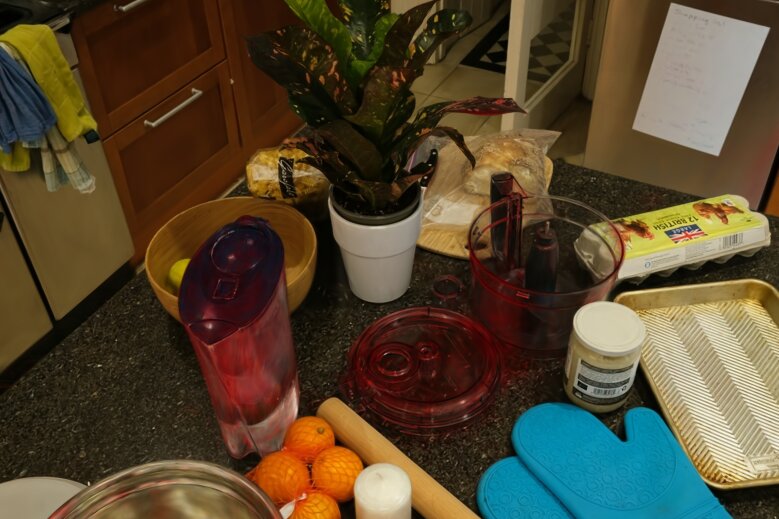}
\end{minipage}

\vspace{1mm} % <-- Added space between the two image groups

\begin{minipage}[b]{0.24\textwidth}
\centering
\includegraphics[width=\linewidth,height=0.22\textheight,keepaspectratio]{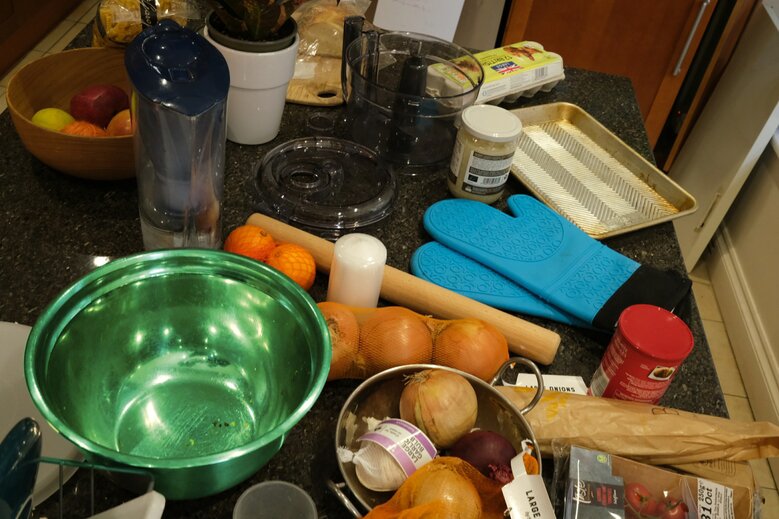}
\end{minipage}
\begin{minipage}[b]{0.24\textwidth}
\centering
\includegraphics[width=\linewidth,height=0.22\textheight,keepaspectratio]{figures/limitations/DSCF6083.jpg}
\end{minipage}
\begin{minipage}[b]{0.24\textwidth}
\centering
\includegraphics[width=\linewidth,height=0.22\textheight,keepaspectratio]{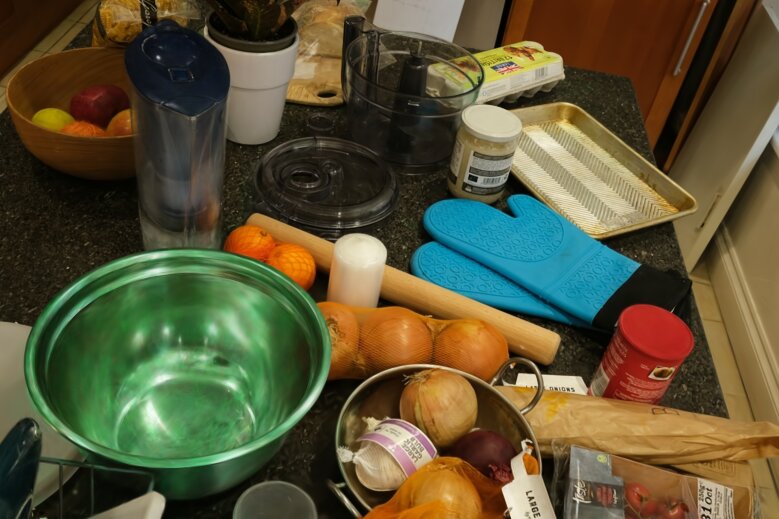}
\end{minipage}
\begin{minipage}[b]{0.24\textwidth}
\centering
\includegraphics[width=\linewidth,height=0.22\textheight,keepaspectratio]{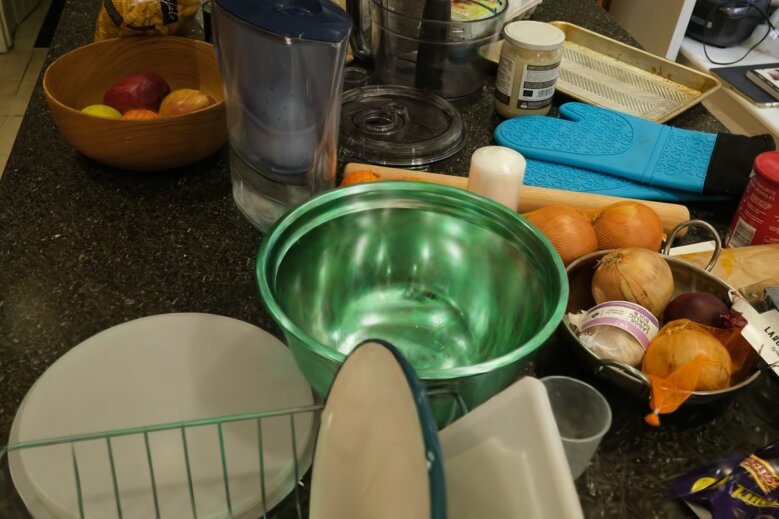}
\end{minipage}

\vspace{-1mm}

\caption{    \textcolor{review}{\textbf{Limitations of diffuse/specular modeling.} 
First row \textbf{Kitchen} scene, second row \textbf{Counter} scene. Comparison between Ground Truth, Diffuse, Specular and Combined RGB render. 
Our simple diffuse/specular model struggles with complex or anisotropic materials.
This can be seen in the Counter (bottom) scene, while the Kitchen (top) scene, especially the table, shows better results.
Still, both yield solid RGB reconstructions.
Third row - Fourth Row: Comparison of original, edited, and rendered views on specular failure cases. In the case of the water jug, although we are able to apply the recoloring, the result lacks uniformity: some areas of the jug appear more red than others, and the color bleeds onto objects visible through the jug’s transparent surface. For the reflective bowl, we manage to change its color as well, but the edits are less clean compared to those on non-reflective, opaque objects. Some regions of the bowl retain the original color, with only a slight green tint appearing in the edited output}}
\label{fig:combined_limits}
\end{figure*}

\begin{figure*}

    \centering
	\includegraphics[width=0.99\textwidth]{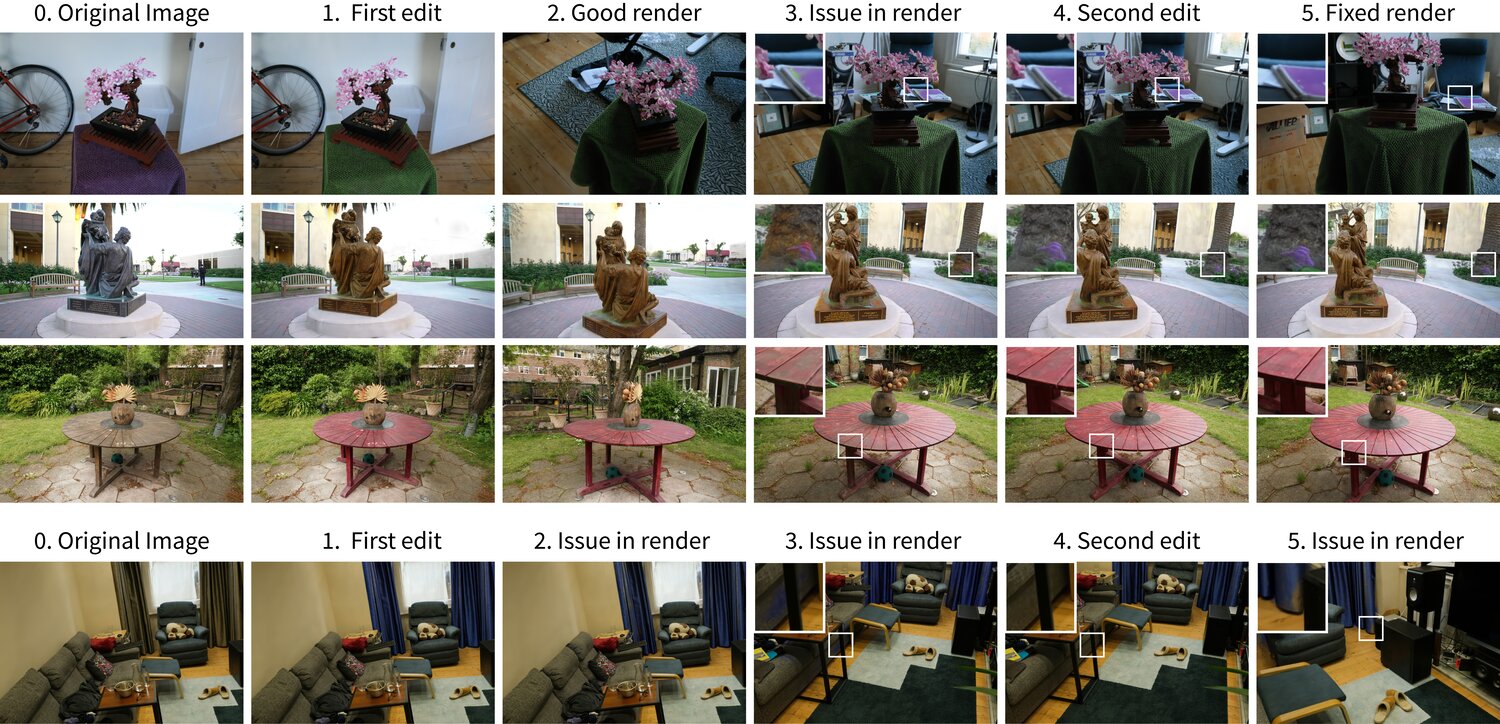} 

% Caption and label
    \caption{    {\bf Limitations in segmentation.} We show examples where issues in recoloring happened. These cases occur mainly with parts of the scene that were not present in the first edited image. In the first example, the notebook has been incorrectly recolored. In the second example, the palm tree has been incorrectly recolored. In the third example, the legs of the table have not been fully recolored to red. These cases were fixed by adding a second edited image. In most cases, such issues can be solved by adding an additional edited image, but in some cases, like the last example, certain edits cannot be achieved. In this particular case, the curtain is far from the camera, and not much viewpoint change is present in the dataset from which to learn the correct model.}

\label{fig:limitations_and_multiedits}
\end{figure*}

\noindent While our approach excels in handling a wide range of color editing tasks, it has several limitations. First, our current material model with diffuse/specular separation is very simple, thus, more complex materials with transparency, anisotropy, or with colored reflections are not captured by the proposed representation. 
\textcolor{review}{In Fig.~\ref{fig:combined_limits}, first and second rows, we show two examples of our decomposition. In the scene named as \textit{Kitchen} (first row, the one with the Lego), the result appears more coherent and smooth, successfully separating reflections from the table and the top of the tractor wheels.
While the \textit{Kitchen} scene handles reflections reasonably well, we observe that the decomposition breaks down in more complex cases, such as the mirror-like surfaces in the \textit{Counter} scene (second row).}
\textcolor{review}{As discussed previously, the goal of our decomposition is not physical accuracy but editability: specifically, enabling stable and multiview-consistent recoloring without degrading the final render.
This is precisely why we introduced the separation of MLPs in our design: to ensure that recoloring operations target only the components unaffected by view direction, maintaining consistency across novel views.}
\textcolor{review}{To show what our method would yield as a result in such failure cases, we have performed in Fig~\ref{fig:combined_limits} some examples where we try to edit a transparent object, the water filter jug, and an object with a mirror-like surface, the bowl. In both cases our approach edits parts of the scene that it should not modify, generating a bleeding out of the color edit into other objects.} Another limitation is that in certain cases a single edited view does not have enough information to segment the scene perfectly. This happens mainly in cases in which some parts of the scene, that were not present in the original edited image, are fairly similar to parts of the desired edition. Luckily, due to our method being interactive this can be easily fixed by adding a second edited image, as we show in the supplementary video. We show examples of wrong recoloring cases in Fig.~\ref{fig:limitations_and_multiedits}. In those cases we show how we tried to solve the issues by adding a second edited image.
%
% PENDING FOR V2 -- Figure \Elena{shows a few examples where our method blah}
%In those cases the classification between edited part and non edited part could be ill-posed and in some cases there might be bleeding out of the recoloring. Luckily, due to our method being interactive this can be easily fixed by adding a second edited image. We show in our video an example of such a case. This makes our method capable of recoloring any scene that can be modeled with gaussian splatting by just resolving the possible ambiguity of the first recolored image in real time.

\begin{comment}

\end{comment}

\section{Conclusions}  
\label{sec:conclusions}

\noindent We have presented a novel method that, by disentangling the color components and integrating multiview information during training, achieves high-quality in color editing for 3DGS. This represents a significant advancement, marking the first instance of a Gaussian splatting editing technique capable of generating recolored models of superior quality. Moreover, our method operates with remarkable speed ($\sim$2 seconds), rendering it highly suitable for interactive use by virtual production engineers or game designers seeking to efficiently edit assets.
As future work, we aim to explore more complex materials and appearances, such as extending to specular color editing.
Furthermore, investigating the minimum number of views required to ensure consistent representations while reducing memory consumption for complex edits is another promising direction.
%We have developed a novel method that, by decoupling the color components and introducing multiview information in the training is capable of producing high-quality color editions of Gaussian splatting scenes. The proposed approach is the first Gaussian splatting edition method that is capable of producing recolored models of very high quality. It not only learns high-quality recolors but also does it so fast ($\sim2$ sec) that it opens the possibility to be used as an interactive tool by virtual production engineers or game designers to edit assets.

\vspace{1mm}

%Our approach is not free of limitations. As shown in the video, 
%Another limitation, is that in certain cases the edited view does not have enough information to segment the scene perfectly. This happens mainly in cases in which some parts of the scene, that were not present in the original edited image, are fairly similar to parts of the desired edition. In those cases the classification between edited part and non edited part could be ill-posed and in some cases there might be bleeding out of the recoloring. Luckily, due to our method being interactive this can be easily fixed by adding a second edited image. We show in our video an example of such a case. This makes our method capable of recoloring any scene that can be modeled with gaussian splatting by just resolving the possible ambiguity of the first recolored image in real time.

\begin{comment}

\end{comment}
    
\clearpage
\begin{IEEEbiography}[{\includegraphics[width=1in,height=1.25in,clip,keepaspectratio]{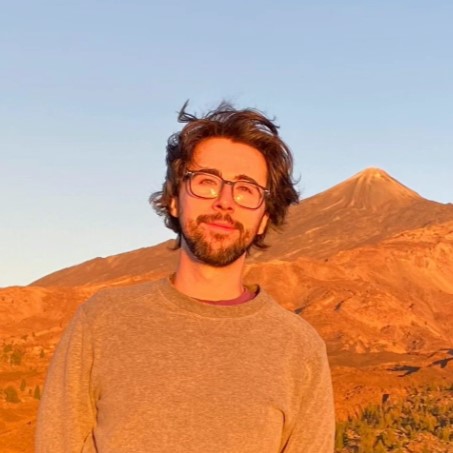}}]{Alessio Mazzucchelli} received the B.S. degree and the M.S. degree in computer engineering at Politecnico di Milano. He is currently pursuing the Ph.D. degree at Universitat Politècnica de Catalunya while developing its research at Arquimea Research Center. His research is mainly focused on editing implicit and explicit 3d representation.
\end{IEEEbiography}

\begin{IEEEbiography}[{\includegraphics[width=1in,height=1.25in,clip,keepaspectratio]{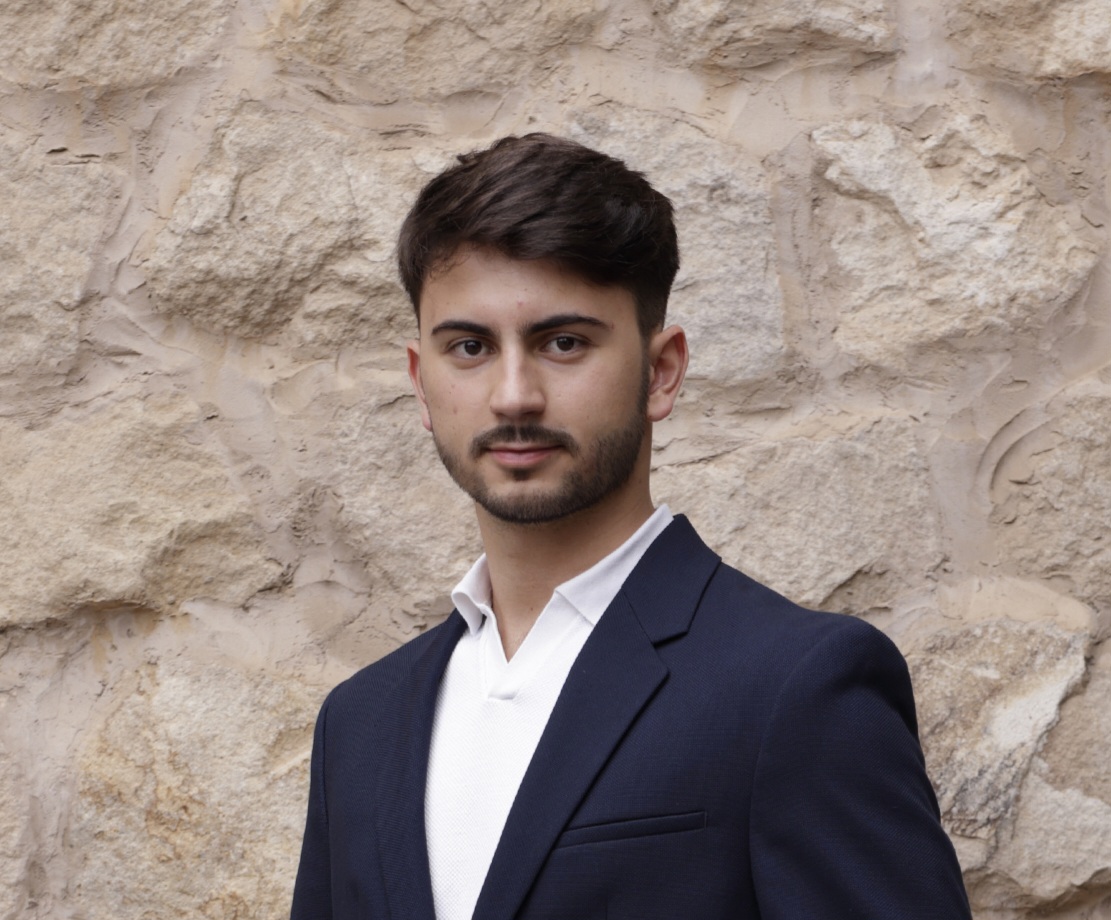}}]{Ivan Ojeda-Martin} received the B.S. degree in computer engineering at Universidad de Las Palmas de Gran Canaria. He is currently developing research at Arquimea Research Center. 
\end{IEEEbiography}

\vspace{-10mm}

\begin{IEEEbiography}[{\includegraphics[width=1in,height=1.25in,clip,keepaspectratio]{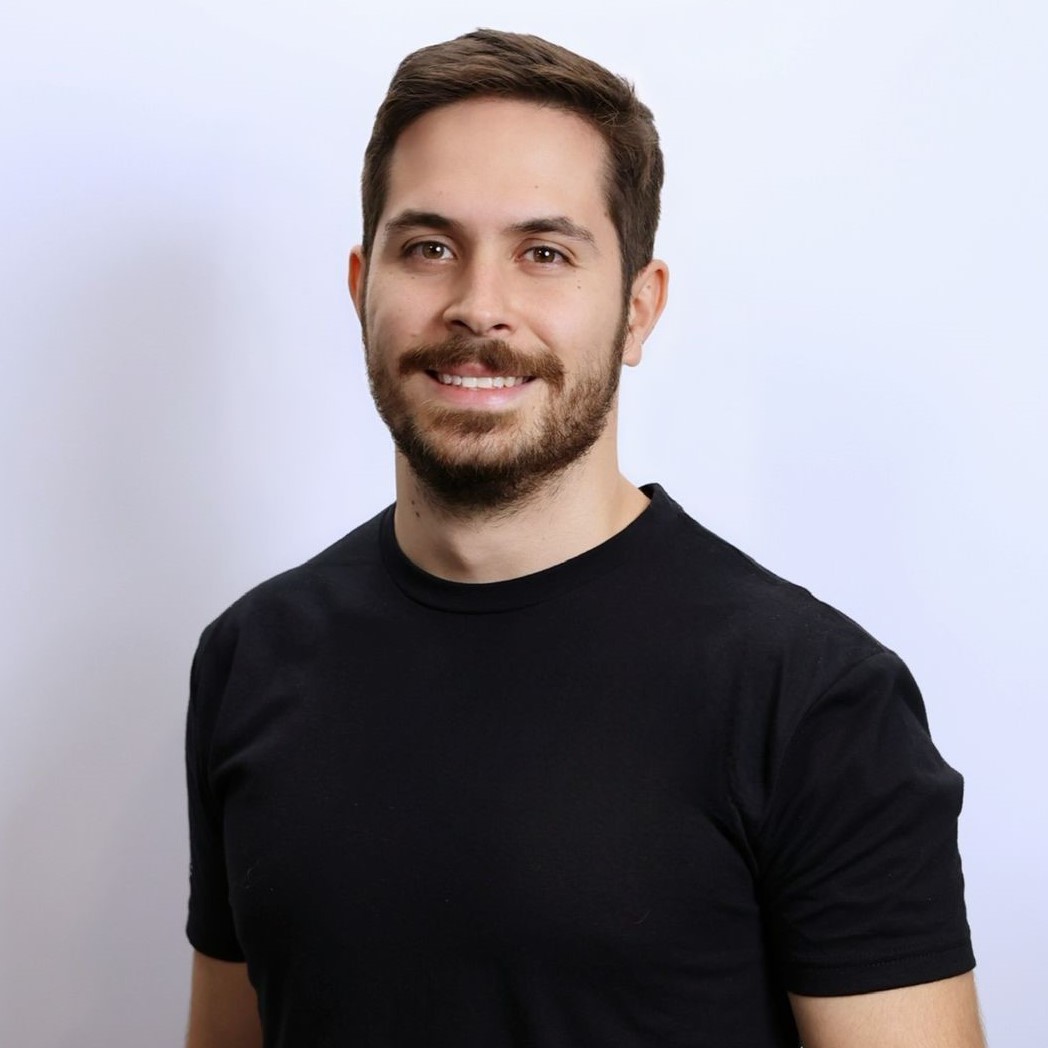}}]{Fernando Rivas-Manzaneque} received the B.S. degree and the M.S. degree in electrical engineering at Carlos III University of Madrid. He is currently pursuing the Ph.D. degree with Technical University of Madrid while developing its research at Volinga AI. His research is mainly focused on neural fields and his research interests include neural rendering, scene reconstruction, and image processing. 
\end{IEEEbiography}
\vspace{-10mm}
\begin{IEEEbiography}[{\includegraphics[width=1in,height=1.25in,clip,keepaspectratio]{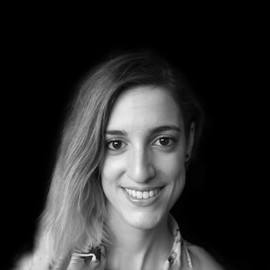}}]{Elena Garces} is a Senior Research Scientist at Adobe Paris since June 2024. Before that, she was Juan de la Cierva Fellow at the MSLab of the URJC and a Director at SEDDI (Madrid) where she led a technology area with expertise in render, optical capture, and AI. During 2016-2018, she was a Postdoctoral researcher at Technicolor R\&D (France). She completed her Ph.D. in 2016 in the Graphics and Imaging Lab of the University of Zaragoza (Spain). During her Ph.D. she had the opportunity to intern twice at Adobe Research (San Jose and Seattle, US). She proudly received the 2017 Premio de Investigación de la Sociedad Científica Informática de España y Fundación BBVA. Press coverage: FBBVA, Innova Spain, El País, El Mundo El Español. Her research interests span the fields of computer graphics, computer vision, and applied machine learning. In particular, I currently work on digitization (optics\&physics) of fabric materials, machine learning-based simulations of garments, and 3D scene reconstruction.
\end{IEEEbiography}
\vspace{-10mm}

\begin{IEEEbiography}[{\includegraphics[width=1in,height=1.25in,clip,keepaspectratio] {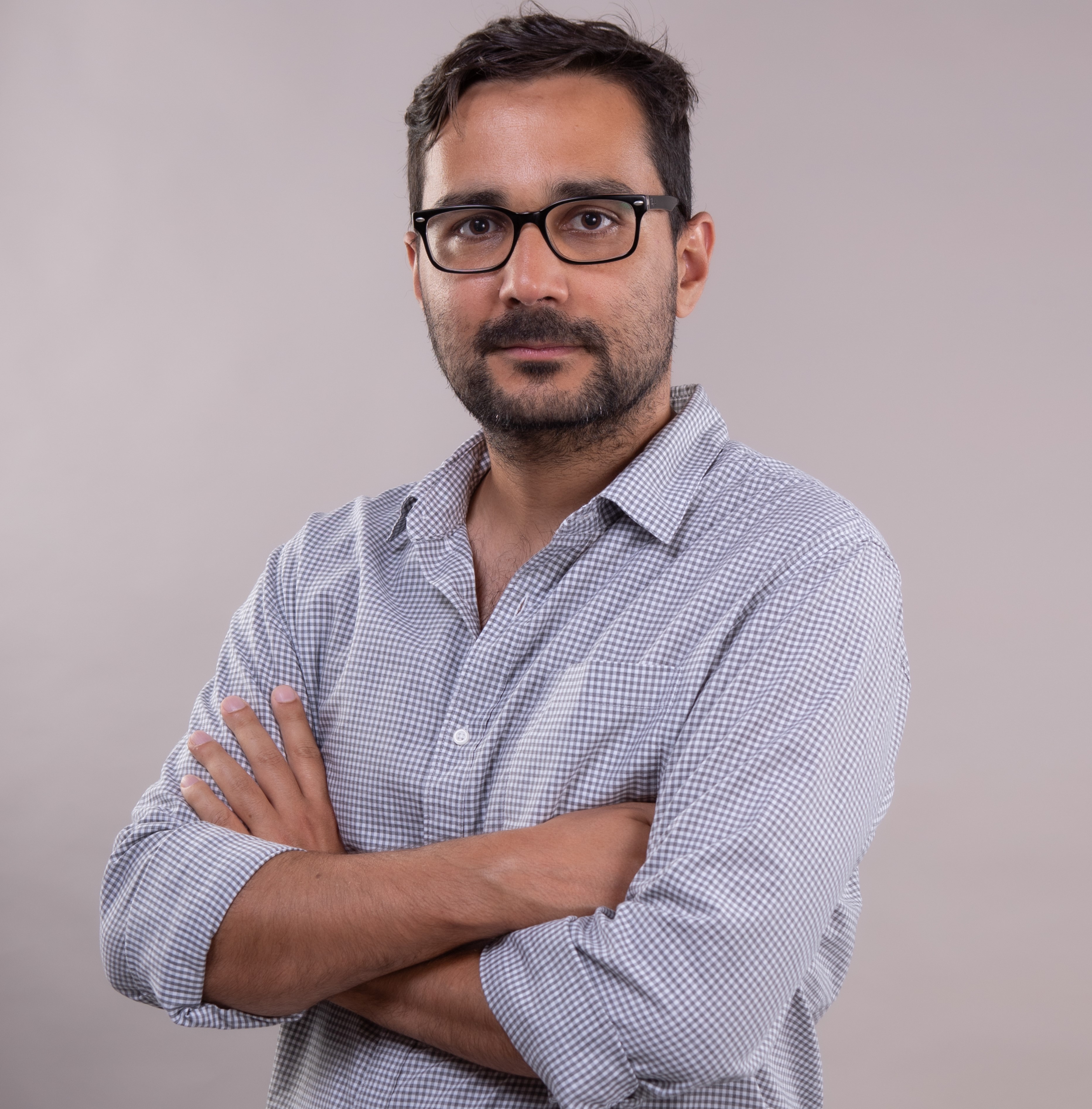}}] {Adrián Penate-Sanchez} (PhD) received his PhD from the CSIC-UPC robotics lab (Institut de Robòtica i Informàtica Industrial) in Barcelona under the supervision of Juan Andrade Cetto and Francesc Moreno Noguer. In 2014, he joined the Computer Vision group at Toshiba’s Cambridge Research Laboratory for a 5 month internship. From Nov 2015 till February of 2018 he was a Postdoctoral Research Associate at University College London (UCL) working in the fields of 3D computer vision and machine learning. In March of 2018 he joined the University of Oxford as a Postdoctoral Research Associate working to find solutions to robot navigation in unstructured environments like forests. In 2020 he joined the University of Las Palmas de Gran Canaria as a Distinguished Researcher and Lecturer under a Beatriz Galindo grant.
\end{IEEEbiography}

\vspace{-10mm}

\begin{IEEEbiography}[{\includegraphics[width=1in,height=1.25in,clip,keepaspectratio] {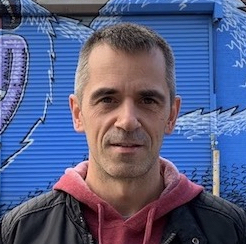}}] {Francesc Moreno-Noguer} is a Research Scientist of the Spanish National Research Council at the Institut de Robotica i Informatica Industrial. His research interests are in Computer Vision and Machine Learning, with topics including human shape and motion estimation, 3D reconstruction of rigid and nonrigid objects and camera calibration. He received the Polytechnic University of Catalonia’s Doctoral Dissertation Extraordinary Award, several best paper awards (e.g. ECCV 2018 Honorable mention, ICCV 2017 workshop in Fashion, Intl. Conf. on Machine Vision applications 2016), outstanding reviewer awards at ECCV 2012 and CVPR 2014, 2021, and Google and Amazon Faculty Research Awards in 2017 and 2019, respectively. He has (co)authored over 150 publications in refereed journals and conferences (including 10 IEEE Transactions on PAMI, 5 Intl. Journal of Computer Vision, 28 CVPR, 11 ECCV and 9 ICCV)
\end{IEEEbiography}

\bibliographystyle{IEEEtran}
\bibliography{bibliography}

\end{document}